\documentclass[a4paper,fleqn]{cas-sc}
\usepackage[numbers]{natbib}
\usepackage{enumitem}
\usepackage{hyperref}
\usepackage{float}
\usepackage{svg}
\usepackage{graphicx}
\usepackage{multirow}
\usepackage{subcaption}
\usepackage{caption}
\usepackage{longtable}

\usepackage{tabularx}
\usepackage{booktabs}
\usepackage{array}

\usepackage{adjustbox}

\newcommand{\cascaptionoftable}[2]{%
  \refstepcounter{table}%
  \label{#2}%
  \par\noindent
  {\parbox{\linewidth}{%
    \rightskip=0pt
    \sffamily\small
    \textbf{\color{scolor}Table~\thetable}\par
    #1\par\vskip4pt
  }}%
}

\makeatletter
\newsavebox{\casfigcaptionbox}
\newcommand{\cascaptionoffigure}[3][\linewidth]{%
  \refstepcounter{figure}%
  \label{#3}%
  \addcontentsline{lof}{figure}{\protect\numberline{\thefigure}{#2}}%
  \vskip6pt%
  \sbox{\casfigcaptionbox}{%
    \sffamily\small\textbf{\color{scolor}Figure \thefigure:}~#2%
  }%
  \ifdim\wd\casfigcaptionbox<#1\relax
    \parbox{#1}{%
      \centering
      \sffamily\small
      \textbf{\color{scolor}Figure \thefigure:}~#2\par
    }%
  \else
    \parbox{#1}{%
      \raggedright
      \sffamily\small
      \textbf{\color{scolor}Figure \thefigure:}~#2\par
    }%
  \fi
  \vskip6pt%
}
\makeatother



\newcommand{\breakers}{\texttt{breakers}}
\newcommand{\builders}{\texttt{builders}}
\newcommand{\bibir}{\texttt{BiBiR}}

\begin{document}
\let\WriteBookmarks\relax
\def\floatpagepagefraction{1}
\def\textpagefraction{.001}

\title [mode = title]{\texttt{Build it}, \texttt{Break it}, \texttt{Repeat}: Benchmarking and improving LLM-manipulated disinformation detection in social media posts}

\author[]{Kevin Thomas}
\fnmark[1]
\author[]{Milosz Kasprzyk}
\fnmark[1]
\author[]{Reuel C Igbokwe Onuigbo}
\author[]{Elliott Pert}
\author[]{Cameron Tovey}
\author[]{João A. Leite}
\author[]{Olesya Razuvayevskaya}
\author[]{Carolina Scarton}[type=editor,
                        auid=000,bioid=1,
                        orcid=0000-0002-0103-4072]
\cormark[1]
\ead{c.scarton@sheffield.ac.uk}
\ead[url]{https://sheffield.ac.uk/cs/people/academic/carolina-scarton}

\affiliation[]{organization={University of Sheffield},
    addressline={Department of Computer Science, Regent Court, 211 Portobello Street, Sheffield, S1 4DP},
    country={United Kingdom}}

\cortext[cor1]{Corresponding author}
\fntext[fn1]{Authors with joint first-author contribution.}

\shorttitle{\texttt{Build it}, \texttt{Break it}, \texttt{Repeat}: Detecting LLM-manipulated social media posts}
\shortauthors{Thomas and Kasprzyk et~al.}

\begin{abstract}
Detecting machine-generated disinformation on social media is increasingly difficult as large language models (LLMs) make it easier to generate and rewrite misleading content at scale. Static benchmark evaluations, measuring detector performance on fixed held-out datasets, do not capture how detectors behave when posts are deliberately transformed to evade classification. This paper adapts the \texttt{Build it, Break it, Fix it} framework into \texttt{Build it, Break it, Repeat} (\bibir): iterative sessions designed to stress-test detectors' robustness under iterative adversarial conditions, evaluating whether models remain reliable when disinformation posts are systematically transformed to evade classification. Across five iterations, the findings show that the best adversarial \breakers' transformations came from a combination of back-translation and LLM persona-based rewriting, with the best performing technique achieving a 95\% label flip rate ($LFR$), whilst still preserving the meaning of the original posts. The best \builders' model was a triplet contrastive model with a dynamic anchor switching (DASS) architecture, which achieved an average accuracy of  72.68\%, outperforming the strong baseline (a fine-tuned \texttt{e5-small-LoRA}) by 15 percentage points on the most robust set of \breakers' adversarial attacks. The results demonstrate that an iterative framework best exposes detector weaknesses and pushes robustness improvements; however, it may still require semantic preservation analysis to distinguish valid adversarial evasion from transformations that changed the original disinformation claims' meaning.

\end{abstract}

\begin{highlights}
\item Iterative adversarial framework to assess machine-generated text detection. 
\item \breakers~developed (chained) adversarial techniques to \textit{break} machine-generated text detection models. Adversarial techniques included: character-level and lexical perturbations, stylometric camouflage and prompt-based evasion.
\item \builders~developed robust models to minimise attack evasion. Models started with a strong baseline (a fine-tuned \texttt{e5-small-LoRA}) and evolved to contrastive learning approaches. 
\item The \breakers-\builders~process was repeated in various iterations, where \breakers~increased the complexity of attacks and \builders~tested incrementally robustness in models.  
\item Findings show that (1) chained adversarial techniques achieved the highest evasion scores; (2) manual evaluation and semantic preservation analysis established prompt-based evasion and best adversarial techniques; (3) triplet networks with dynamic anchor switching was the best detector; and (4) iterative evaluations showed more reliable results than static benchmarks.
\end{highlights}


\maketitle

%
%
\section{Introduction}
\label{sec:Introduction}

Recent reports indicate a sharp rise in websites publishing AI-generated news articles and political content at scale and in multiple languages\footnote{\url{https://www.newsguardtech.com/special-reports/ai-tracking-center/}}, with part of this content being potentially categorised as Large Language Model (LLM)-generated\footnote{The terms \emph{LLM-generated}, \emph{Machine-generated} and \emph{AI-generated} are used interchangeably in this paper.} disinformation\footnote{\textit{Disinformation} refers to false information that is deliberately created/shared with malicious intent \cite{TURCILOandOBRENOVIC}. The term disinformation is used throughout this paper to refer to machine-generated misleading content.}. One of the main risks of LLM-generated disinformation is that generative models reduce the production barriers that constrained large-scale influence operations. While traditional campaigns, such as state-backed propaganda coordinated by the Russian Internet Research Agency, previously depended on coordinated human labour, linguistic expertise and strategic planning \citep{ref14a}, LLMs eliminate these constraints by allowing misleading content to be produced at low cost and with limited supervision \citep{barman2024darkside, ref18, zhou2023synthetic}. This enables less technically-skilled actors to generate believable content in different styles and domains, whilst larger malicious organisations can use the same system to increase their own speed and volume of existing operations. 
Another risk is the adaptive personalisation capabilities of LLMs, as false claims can be rewritten across many languages, personas and target audiences, which makes disinformation content easier to personalise \citep{buchanan2021truth, matz2024personalized, zugecova2025evaluation, JoaoPaper}. These modifications can be particularly harmful if used on social media, where false information have been shown to spread faster and more broadly than true information \citep{ref13}. In particular, emotionally charged content, provoking fear, anger or outrage, is more likely to facilitate engagement and dissemination \citep{ref25, brady2017emotion}.
These factors threaten to overwhelm existing verification processes, making it increasingly difficult for 
general audience, professional fact-checkers and journalists, as well as  automated systems to distinguish between genuine and synthetic content \citep{ref4}. 

\begin{figure}
    \centering
    \includegraphics[width=0.8\linewidth]{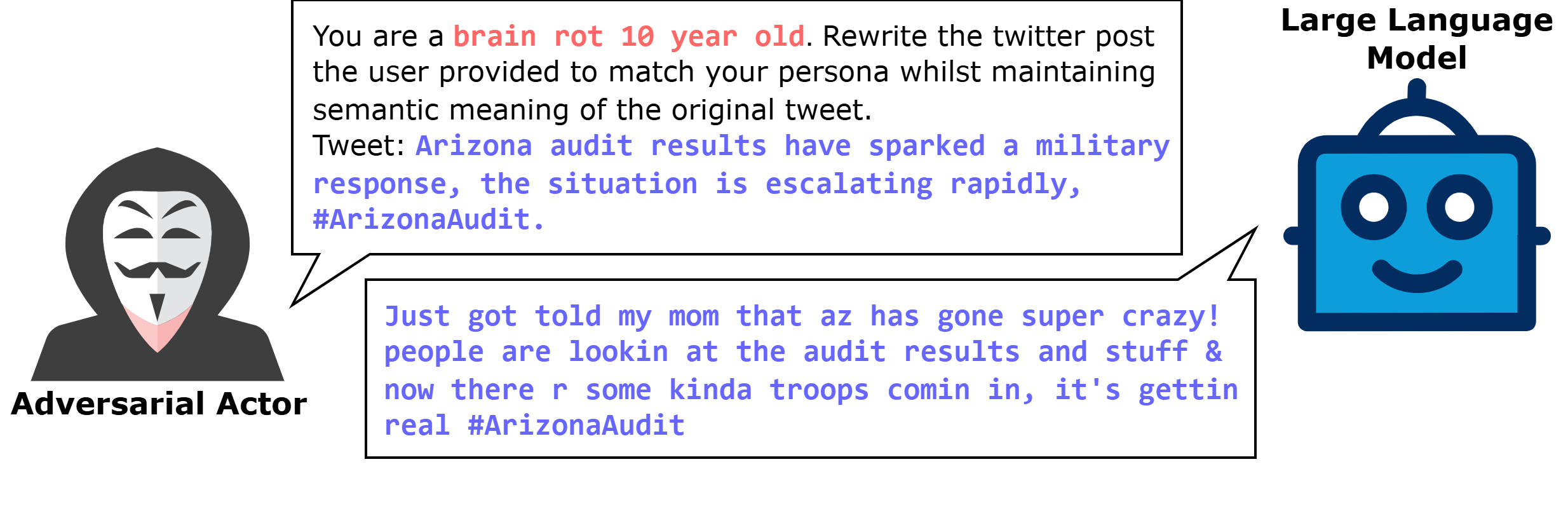}
    \caption{An example of an adversarial actor prompting an LLM to rewrite the a disinformation tweet into the style of the persona \textit{brain-rot 10-year old}.}
    \label{fig:adversarial_prompt_llm_example}
\end{figure}


These challenges motivate the development and evaluation of automated systems to ensure reliable detection of LLM-generated content, especially when the content is deliberately modified to avoid detection \citep{ref54, ref21}. Existing detectors often perform well on benchmark datasets, effectively separating machine-generated and human-written texts, via zero-shot and feature-based detection settings \citep{gltr, ref57, ref54}. However, real-world disinformation is unlikely to remain static as adversarial actors can rewrite, paraphrase, translate, reframe or stylistically alter generated content while preserving the underlying disinformation claim (see example in \hyperref[fig:adversarial_prompt_llm_example]{Fig.~\ref*{fig:adversarial_prompt_llm_example}}) \citep{ref28, ref6, ref19, ref31}. Techniques such as paraphrasing and other surface-level transformations have been shown to reduce the reliability of AI-generated text detectors,   including classifier-based, watermarking and zero-shot approaches \citep{ref28, ref20, ref21}. This creates a detection problem, as detectors are not only distinguishing human-written text from machine-generated one, but are doing so against content that has been intentionally transformed to appear more human-like or less machine-identifiable \citep{ref17, ref31, ref6}. 
Existing research on LLM-generated general text or disinformation has often focused on longer-form content, such as news articles, propaganda-style narratives or essays \citep{ref17, ref24, ref44, ref54, JoaoPaper}. Significantly less attention has been devoted to detecting LLM-generated disinformation in short-form social media posts \citep{ref22, ref23, ref24}, despite the urgent need to 
combat LLM-generated disinformation in social media. 

To address the above-mentioned gaps, this paper adapts the \texttt{Build it}, \texttt{Break it}, \texttt{Fix it} framework\footnote{\url{https://builditbreakit.org/}} \citep{10.1145/2976749.2978382} into a \texttt{Build it}, \texttt{Break it}, \texttt{Repeat} (\bibir) framework, which suits the machine-generated detection's inherently adversarial nature, with \builders~attempting to detect LLM-generated content and \breakers~attempting to modify machine-generated content so that it preserves the disinformation claims, but becomes harder to identify as being machine-generated. Iterative versions of this framework have previously been used for offensive and safe language detection in social media \citep{BIBIeg1}, fact-checking model benchmarking \citep{BIBIeg2} and LLM jailbreaking evaluation \citep{BIBIeg3}. 

The overarching objective of this paper is, therefore, to evaluate and improve detector robustness under iterative adversarial conditions, in the context of short social media posts, where detection systems are improved by \builders~and then stress tested by \breakers\footnote{This work received ethics approval by the University of Sheffield Research Ethics Committee.}. \texttt{Builders} measure model performance based on accuracy ($ACC$) and $F1$-score with the aim of improving the classification of human-authored vs. LLM-generated disinformation posts. \texttt{Breakers}' effectiveness is measured using detector performance degradation via label flip rate ($LFR$) and semantic preservation checks to ensure $LFR$ is not an effect of the original disinformation claims' meaning being significantly modified. The paper also discusses the advantages of \bibir~evaluation (repeated \builders-\breakers~rounds where transformed posts are used to test and refine detector robustness) over static evaluation (a one-off test on held-out data), assessing whether \bibir~can reveal weaknesses in performance that may be missed by single benchmark evaluation. 
To address its key objective, the paper answers the following research questions:

    
    

    \begin{itemize}[label={}]\item \textbf{RQ1} Which adversarial transformation techniques cause the largest reduction in detector performance, measured through accuracy and label flip rate?

    \item \textbf{RQ2} Which adversarial techniques are best at preserving the meaning of the original text, whilst still evading detection?
    
    \item \textbf{RQ3} Between baseline transformer and contrastive model architectures, which is more effective at distinguishing machine-generated and human-written social media disinformation?
    
    \item \textbf{RQ4} What are the benefits of an iterative \bibir~evaluation over static evaluation in exposing detector vulnerabilities?
\end{itemize}

In answering these research questions, the paper's main contributions can be summarised as follows:

\begin{enumerate}
    \item A focus on \textbf{disinformation narratives / stories on short social media posts} from Twitter (now X), which (1) sheds light on how LLMs deal with disinformation narratives and short social media texts; (2) assesses the performance of machine-generated text detection for short social media texts.
    \item An iterative \bibir~framework adapted for \textbf{evaluating whether machine-generated disinformation detectors remain robust when social media posts are adversarially transformed}, using LLMs and/or rule-based edits. 
    \item \textbf{Two new datasets:} (1) \textbf{bld\_data:} A tailored dataset for fine-tuning machine-generated text detectors to the social media domain with 2,560 examples for baseline training and 3,840 for contrastive learning models with dynamic anchor switching (DASS). (2) \textbf{brk\_data:} A large-scale balanced dataset containing 1,08M LLM-generated/altered disinformation posts, constructed from an evaluation set with 125 machine-generated original posts (MGO) and 125 human-generated original posts (HGO). 
    \item A \textbf{benchmark of multiple adversarial generation and transformation techniques}, measuring their effect on detector robustness using accuracy, $LFR$ and semantic-preservation criteria.
    \item A \textbf{comparison between static and iterative adversarial evaluation}, showing how \bibir~rounds can provide a realistic assessment of detector robustness under adaptive adversarial conditions. 
    \item Available code and data at \url{https://github.com/GateNLP/bibir_machine_generated_detection}.
\end{enumerate}


The paper is structured as follows. \hyperref[sec:related-work]{Section~\ref*{sec:related-work}} introduces the related work and provides context on prior studies of LLM-generated disinformation. The \bibir~overall framework is discussed in \hyperref[sec:proposed-framework]{Section~\ref*{sec:proposed-framework}}, with the specific \breakers~and \builders~approaches being discussed in Sections \hyperref[sec:breakers_approach]{~\ref*{sec:breakers_approach}} and \hyperref[sec:builders_approach]{~\ref*{sec:builders_approach}}, respectively. Then, \hyperref[sec:results]{Section~\ref*{sec:results}} discusses the results in relation to the research questions. This leads to \hyperref[sec:discussion]{Section~\ref*{sec:discussion}} which concludes our findings and discusses their broader implications in relation to prior work. 

%
%
\section{Related Work}
\label{sec:related-work}

\subsection{LLM-Generated Text Detection Evasion (\breakers)}


\paragraph{Red teaming} is a systematic approach for stress-testing AI systems by applying adversarial inputs or attack strategies to expose failure modes and vulnerabilities \citep{redteamingdef, ref46}. In this paper, \breakers~focus on \textit{\textbf{jailbreaking}}, i.e. the re-framing of a restricted disinformation request so that it appears as ``benign'' rewriting \citep{BIBIeg3, ref64}. LLMs may refuse direct requests to generate misleading political content due to guardrails, but tend to comply when the same underlying claim is embedded within an indirect/role-based prompt \citep{ref39, ref64}. This vulnerability can be used in conjunction with other techniques like character-level perturbation, lexical perturbation, stylometric camouflage and prompt-based evasion to reduce detector reliability \citep{ref6, ref28, ref31}.

\paragraph{Character-level perturbation} involves using spacing edits (e.g. double space \& newline insertion) and punctuation noise to mimic typos made by humans. These type of actions disrupt tokenisation-based detectors and can significantly reduce detector accuracy \citep{ref6, ref19, ref26}. Additionally, evidence shows meaning preserving surface edits reliably break text classifiers even when human judgment of the sentence semantics remains intact \citep{ref59, ref60, ref61}. \citeauthor{ref27}~\cite{ref27} show that under synonym-substitution perturbation, standard classifiers fall from their usual accuracy of 80-90\% to single digit or low teen accuracies, despite the underlying meaning being the same. 
In the context of disinformation, these modifications allow the LLM-generated message to resemble authentic human text by mimicking natural noise, which facilitates detector evasion.

\paragraph{Lexical perturbation.}
Techniques like paraphrasing and back-translation are another type of attack used to mask LLM-generated text. Such attacks preserve the semantic context of text while altering its surface level structure \citep{ref28, ref29, ref19, ref20}. Paraphrasing is effective as it rewrites the syntactic and lexical choices whilst maintaining the original meaning of the text. For instance, \citeauthor{ref28}~\cite{ref28} show that paraphrasing can reduce the effectiveness of detectors such as DetectGPT \citep{ref57}, GPTZero \citep{ref62}, watermarking-based methods and OpenAI's text classifiers\footnote{Discontinued tool; used for early evaluation of AI-text detection: \url{https://openai.com/blog/new-ai-classifier-for-indicating-ai-written-text}}. \citeauthor{ref29}~\cite{ref29} use the OPUS MT models \citep{opus} to back-translate using French as the intermediate language between English translations, achieving an average accuracy drop of 11.54\% across the 8 LLMs evaluated. This finding indicates that back-translation is an effective obfuscation technique because it disrupts the text's underlying linguistic structure. 

\paragraph{Stylometric Camouflage.}
Stylometric irregularity complicates authorship detection. For example, \citeauthor{ref30}~\cite{ref30} show that obfuscation and mimicry attacks reduce state of the art authorship verification accuracy by 25 to 40\%. \citeauthor{ref31}~\cite{ref31} demonstrate that injecting human noise (i.e. informal markers, emoji variation, punctuation drift) lowers AI text detector accuracy from above 90\% to below 50\% once stylistic camouflage is applied. These techniques combined show that small stylistic edits in adversarial tweets can reduce detector performance and allow the tweets to blend in as genuine human authored ones. 

\paragraph{Prompt Based Evasion}
involves exploiting prompt engineering to alter tone and persona types. This includes tones that are political, emotionally manipulative or mimic the linguistic profile of a specific demographic \citep{matz2024personalized, zugecova2025evaluation, JoaoPaper, ref77, ref39}. This type of prompt produces adversarial outputs that are more natural, rather than outputting a default model-shaped text. Benchmarking work by \citeauthor{ref32}~\cite{ref32} shows that simple prompt manipulation can cause up to a 33\% decrease in core LLM task performance. \citeauthor{ref19}~\cite{ref19} show that including deliberate style shifts can achieve attack success rates exceeding 70-85\% when the model is told to write in alternative personas or rhetorical registers. Similarly, \citeauthor{ref31}~\cite{ref31} demonstrate that the humanisation of prompts reduced detector accuracy by over 50\% without the need to manually edit the output text. These findings show that prompting alone can be sufficient to hinder detector behaviour, making them less reliable at distinguishing LLM-generated text from human-written text.


\paragraph{Summary.} Prior work shows that surface edits, stylometric camouflage and prompt-based rewriting can each reduce the reliability of AI-generated text detectors \citep{ref6, ref19, ref20, ref28, ref31}. However, existing studies tend to evaluate attacks within a single family, such as paraphrasing or surface perturbation, rather than examining how multi-stage combinations affect detector robustness \citep{ref28, ref30, ref31}. In this paper, \breakers~innovate by combining and iterating existing attack types and measuring whether chained transformations reduce detector performance more than individual attacks. Another significant contribution is the assessment of human evaluation and automatic metrics to measure semantic preservation, fluency and disinformation grounding.

\subsection{LLM-Generated Text Detection 
(\builders)} \label{rw:binary_classifiers} 

The task of machine-generated text detection has traditionally been approached as a supervised classification problem where a neural classifier is fine-tuned on the labelled data \citep{wu2026advancing, zeng2026human, gu2024aispace, siino2024badrock, voznyuk2024deeppavlov}. Such methods often achieve state-of-the-art performance on in-domain data, but are known to suffer from poor generalisation \citep{tang2024science, pu2023deepfake}. The rise of LLMs has created a need for detection methods that generalise across different families of generator models and do not rely on newly annotated datasets, which would otherwise need to be continually collected as new LLMs are released. This has led to the emergence of zero-shot machine-generated text detection methods \citep{ref57, ma2026zero, chen2025repreguard}. Such methods typically employ a surrogate LLM to extract internal statistics given the input text, with a lightweight classification model often trained on those features to determine an appropriate classification threshold. 

The prevalent majority of existing benchmarking datasets \citep{ref34,dugan2024raid, wu2024detectrl, yu2025evobench, li2024mage} and data challenges \citep{wang2024semeval, marchitan2024team} position the task of machine generated text detection as a binary classification problem, distinguishing between human-written and machine-generated text. However, the increasing popularity of text humanisation tools and sophisticated adversarial attacks has motivated the development of a more fine-grained task definition. For example, \citeauthor{abassy2024llm} \cite{abassy2024llm} propose a detection framework that extends beyond the conventional binary setting by distinguishing not only between human-written and machine-generated text, but also between machine-generated text that has been humanised and human-written text that has been refined using LLMs. 
Our work occupies an intermediate position between these two formulations. While we retain the binary detection setting, we explicitly investigate the robustness of LLM-generated text detection methods in the presence of adversarial modifications, with the goal of identifying their vulnerabilities and improving their resilience to such attacks.

\subsubsection{Supervised Methods}\label{rw:supervised}

The majority of state-of-the-art LLM-generated text detection approaches represent fine-tuned transformer-based models or an ensemble of such models \citep{mobin2025luxveri, kandula2025bbn, agrahari2025random, edikala2025leidos, marchitan2024team, wei2024team, xiong2024ncl}. Adversarial training with proximal policy optimization (PPO) is an efficient supervised approach, particularly when the input text undergoes different adversarial attacks \citep{hu2023radar}. In this setting, the model includes a paraphraser and a detector, where the task of the paraphraser is to generate efficient adversarial examples by taking the prediction
of the detector and using it as a reward to update using PPO.  
Another successful approach is training transformer models using 
\textbf{\textit{contrastive learning (CL)}}, since it was shown to achieve high accuracy in detecting machine-generated texts \citep{chen2024team, kandula2025bbn}. Rather than directly predicting labels, these models are trained to bring positive pairs closer together in an embedding space, while pushing negative pairs further apart \citep{SimCLR, SimCSE}. In the context of machine-generated detection, machine-machine pairs are pulled together and minimised, whilst machine-human pairs are maximised. The goal of training is, therefore, to optimise the distance between these pairs, which is done using a specialised loss function. If a positive pair that is semantically equivalent produces embeddings that are far apart, the loss function updates the weights to move the embeddings closer together in future iterations. Contrastive loss functions, such as InfoNCE \citep{InfoNCE} or NT-Xent \citep{SimCLR}, carry out this optimisation process to create an embedding space that clusters semantically similar texts while unrelated texts remain distant in the space. This model architecture is particularly efficient for LLM-generated text detection in the era of adversarial attacks because its ability to cluster semantically similar texts allows the model to look past attacks that change wording and syntax and to focus on the semantic meaning, unlike traditional classifiers that rely on fixed boundaries.

\paragraph{Siamese neural networks.} \citep{chicco2021siamese} is a specific case of a contrastive learning model, which learns relational representations between two input samples simultaneously while using two identical architectures with same weights. Models with siemese architectures, such as Sentence-BERT \citep{SBERT}, have demonstrated strong performance in semantic textual similarity and text comparison tasks, as they preserve semantic meaning and capture contextual relationships. This is particularly relevant for the LLM-generated text detection task because paraphrasing attacks have been shown to reduce the accuracy of detection systems while preserving semantic meaning \citep{chen2024team}. 

\paragraph{Triplet networks} further extend the contrastive pairing architecture of the siamese neural network by evaluating three samples simultaneously: an anchor, a positive sample from the same class, and a negative sample from the opposing class \citep{Schroff_2015}. 
By using the loss function, the anchor is pulled closer to the positive than to the negative by at least a fixed margin, producing a more discriminative embedding space than pairwise comparison alone. In the context of LLM text detection, contrastive triplet networks show great potential in the long-form domain while remaining unexplored in the short-form texts. For example, The WhosAI \citep{La_Cava_2024} triplet network was designed to detect and attribute AI-generated text across multiple generators simultaneously. Although it achieved an F1 score of 0.999 on the Turing Test in the news articles domain, its efficiency on short text inputs remains unexplored.

\paragraph{Dynamic anchor switching strategy (DASS).}Even when triplet networks are trained with carefully selected hard negative examples, paraphrased machine-generated texts with semantic claim preservation remains a challenging attack as it can drift close enough to the human cluster, thus collapsing the inter-class margin. As a defence strategy, GravText \citep{GravText-Triplet} proposes a detection framework that augments triplet contrastive learning with DASS. In a DASS triplet, the three inputs consist of an original LLM sample, its paraphrased variant, and a human sample on the same topic. What distinguishes DASS as an effective defence strategy against paraphrasing is that the triplet loss equation's anchor alternates between the original and the paraphrase across training steps. This bidirectional signal forces the model to learn that the original and the paraphrased variants of the same text belong in a tight machine-generated cluster, while the human text remains beyond a grounded fixed margin. GravText demonstrated the effectiveness of this strategy on static long-form essays and formal question-answering datasets, however, similarly to the triplet network evaluation, short-form texts has not yet been investigated.

\subsubsection{Zero-shot Methods}
The capabilities of LLMs in generating highly coherent and believable text has motivated a need for more universal approaches that do not rely on human-annotated training data. Consequently, a growing body of work has explored zero-shot approaches that leverage surrogate LLMs to extract statistical signals from the input text for distinguishing between human- and machine-generated content. 

\paragraph{DetectGPT, Fast-DetectGPT and Binoculars.} The pioneering method in this direction is \textit{DetectGPT} \citep{ref57}, which uses a surrogate LLM to extract entropy, log-perplexity, and log-rank under difference of a given text. The authors demonstrate that these features are able to distinguish between human and machine-generated text under multiple textual perturbations. Building on this idea,  F\textit{ast-DetectGPT} replaces the computationally expensive perturbation step of DetectGPT with the sampling of the alternative next tokens \citep{bao2024fast}. It then compares their conditional probabilities to that of a given next token. As an extension to DetecGPT, \textit{Binoculars} approach \citep{hans2024spotting} explores log-perplexity of the surrogate LLM as a normalised alternative of perplexity and demonstrates its effectiveness for zero-shot detection.

\paragraph{RepReGuard} method \citep{chen2025repreguard} lies at the intersection between supervised and zero-shot approaches. It extracts representation-based features from a surrogate LLM and trains a lightweight machine learning classifier on a small labelled dataset to calibrate the decision threshold. Specifically, the method exploits fine-grained activation patterns of the surrogate model and applies Principal Component Analysis (PCA) to identify the most discriminative representations. By requiring only a limited amount of labelled data for threshold calibration, RepReGuard combines the generalisation capabilities of zero-shot methods with the adaptability of supervised approaches.

\paragraph{Temperature sensitivity (TS).} Finally, \citeauthor{ma2026zero} \citep{ma2026zero} propose TS, a normalized feature that quantifies the sensitivity of a surrogate LLM's output distribution to changes in the sampling temperature. Specifically, the method measures the normalised difference between the observed and expected logits over the model's vocabulary under low and high temperature settings. The underlying intuition is that LLM-generated text is typically produced by selecting highly probable next tokens, making the difference with the expected logits more sensitive to temperature variations than that for human-written text, which is not generated under this assumption. Experimental results demonstrate that TS substantially outperforms previous zero-shot detection methods across a range of benchmarks.

Although zero-shot methods generalise well across domains and families of generative models while avoiding the need for newly annotated training data, supervised approaches continue to achieve substantially higher performance in in-domain detection settings. Therefore, this work focuses on developing supervised detection methods that remain robust in the presence of adversarially modified inputs.

\subsubsection{Limitations of existing approaches} Existing detectors for AI-generated text have mainly been evaluated on static benchmarks consisting of long-form, formal text, where  adversarial data augmentation is either absent or applied in a single pre-defined round \citep{dugan2024raid}. These constraints make the evaluation results not reflective of the real-world, where evasion techniques and generation models constantly evolve, making detection particularly challenging for social media texts due to their short-form nature. 

It remains unexplored how an architectural ladder of detection approaches performs against chained (i.e. applied sequentially on top of each other) evolving attacks on short-form social media misinformation in an iterative evaluation framework. Specifically, it is unknown whether a progression of detection strategies from simple classifiers and data augmentation to contrastive models with hard negative mining and paraphrase-aware anchor switching (DASS) offers a genuine robustness strategy against adaptive and advanced adversarial techniques in an already difficult and noisy domain. Without addressing this gap, current research in short-form text domains is limited to detector development guided by static conditions that do not reflect the evolving threats in real-world settings.  







\section{Proposed Framework}
\label{sec:proposed-framework}

The results of this paper were achieved as part of a master-level course were a group of five students is required to develop a research project. In the \bibir~framework, the students were divide in \builders~(three students) and \breakers~(two students). \hyperref[fig:framework_img]{Fig.~\ref*{fig:framework_img}} outlines the proposed adversarial \bibir~framework. The framework is organised into five stages (described below in more detail): \textbf{(I) Dataset Construction}, \textbf{(II) \texttt{Breakers}' Transformation}, \textbf{(III) Semantic Preservation Analysis}, \textbf{(IV) \texttt{Builders}' Evaluation}, and \textbf{(V) Challenge Week}.

\begin{center}
    \includegraphics[width=0.8\linewidth]{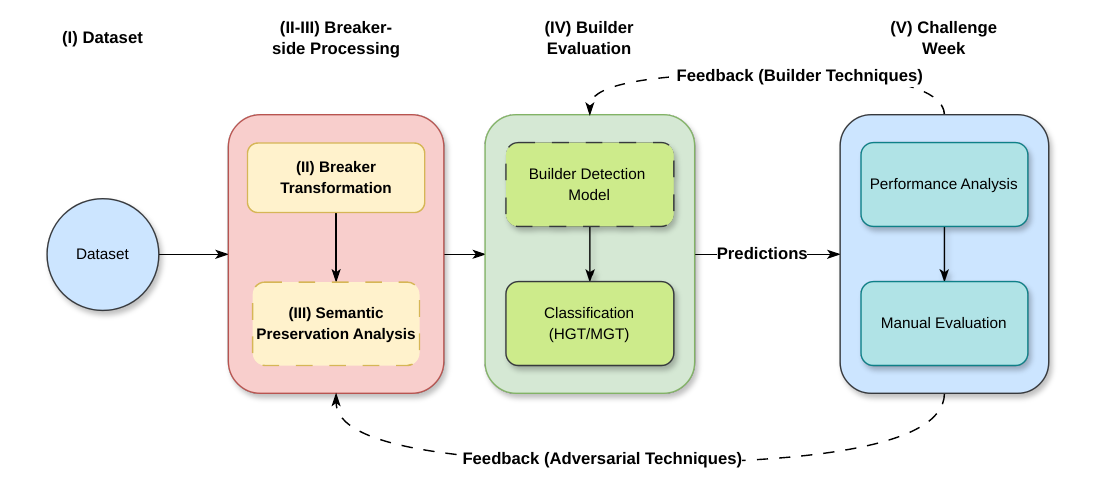}
    \cascaptionoffigure
    {Overview of the proposed adversarial evaluation framework, showing dataset construction, \breakers-side transformation and semantic filtering, \builders' evaluation, Challenge Week analysis, and feedback loops for refining \builders' and \breakers' techniques.}
    {fig:framework_img}
\end{center}


\paragraph{(I) Datasets.} \hyperref[tab:dataset_model_summary]{Table~\ref*{tab:dataset_model_summary}} summarises the data sources and methods used by \builders~and \breakers. The \breakers' dataset (bkr\_data) is composed of two static ground-truth datasets, HGO and MGO, both with the same size to guarantee a balanced evaluation. 
For the HGO dataset, human-authored misinformation tweets were randomly sampled from the rumour subset of PHEME \citep{zubiaga2016pheme}. These samples were then classified using the \builders' baseline classification model before any adversarial transformation. Only the examples classified as human-generated (125 in total) were retained in the HGO baseline set.
The MGO dataset was generated using zero-shot prompting with LLaMA-3.1 \citep{LLaMA3.1}. A total of 50 disinformation claims were used as generation inputs, drawn from TruthSeeker \citep{2023truthseeker} and the false narratives in AI-TRAITS \citep{JoaoPaper}, with five candidate X-style posts generated per seed disinformation claim, producing 250 candidate MGO samples. These candidates were then passed through the baseline \builders' classifier model, and 125 posts classified as machine-generated were selected at random for the static MGO set. 
Filtering the bkr\_data dataset to include only examples initially classified correctly by the \builders' baseline classifier supports a clearer interpretation of $LFR$, since label flips are measured from initially correct predictions rather than from examples already misclassified before transformation. Nevertheless, this introduces selection bias because initially misclassified examples are excluded, including HGO false positives and MGO false negatives. This trade-off was acceptable because our aim was to measure whether \breakers' transformations can cause initially correctly detector predictions to fail. Using both HGO and MGO examples allows the framework to evaluate both directions of detector error: machine-generated posts being misclassified as human-written, and human-written posts being incorrectly flagged as machine-generated. This is important in the context of short-form disinformation as LLM-generated claims can be difficult for both human readers and automated detectors to distinguish from human-authored claims \citep{ref64}. It is also useful to keep the detector grounded in a real-world scenario, i.e. a detector that correctly identifies all MGOs, but at the same time misclassifies all HGOs, is not a useful detector.

To avoid sample-level data leakage, \builders' data (bld\_data) was independent from the posts used by \breakers' (bkr\_data). This ensured that \builders' models were not evaluated on the same examples used during training or validation. For human-authored data, PHEME \cite{zubiaga2016pheme} and Constraint 2021: {COVID-19} datasets \citep{constraint2021dataset} were chosen (1,296 posts were sampled from these datasets). They both have human-verified misinformation and have been used in works such as \citep{usedPheme1, usedPheme2, usedConstraint2}, which selected them for their authenticity in misinformation labelling. For generating the machine-generated counterpart of their dataset, \builders' used tweet injection: the human-authored tweets selected from PHEME and Constraint were embedded into a persona-engineered prompt and passed to an LLM (more details in \hyperref[sec:builder_MGO]{Section~\ref*{sec:builder_MGO}}).

\paragraph{(II) \texttt{Breakers}' Transformation.}
\texttt{Breakers} transform HGO and MGO into adversarial variants using a combination of rule-based perturbations and LLM-based edits. Rule-based perturbations are applied programmatically to the text, while LLM-based transformations are implemented through prompt-controlled rewriting (see \hyperref[sec:transformation]{Section~\ref*{sec:transformation}}). These transformations aim to exploit the reliance on surface-level linguistic features used by detectors \citep{ref19}. 

\renewcommand{\arraystretch}{1.4}
\begin{center}
\cascaptionoftable
{Summary of the datasets used by \builders~(bld\_data) and \breakers~(bkr\_data). The table shows the source data, sample counts, purpose of each subset, and the overlap-control step used to avoid data leakage between \builders' training and \breakers' evaluation.}
{tab:dataset_model_summary}
{\sffamily\small
\begin{tabular}{@{}p{0.02\linewidth} p{0.13\linewidth} p{0.26\linewidth} p{0.1\linewidth} p{0.38\linewidth}@{}}
\noalign{\hrule height 0.75pt}

&\textbf{Component} 
& \textbf{Source / Model} 
& \textbf{Size} 
& \textbf{Purpose / Overlap Control} \\ 
\hline

\multirow{4}{*}{\rotatebox[origin=c]{90}{\parbox[c]{5cm}{\centering \textbf{bkr\_data}}}}&
HGO dataset
& PHEME
& 125 posts
& Human-authored misinformation baseline. Samples were kept disjoint from \builders' data. All samples are classified as \textit{human} by the \builders' baseline classifier. \\ 
\cline{2-5} 

& 
MGO generation
& TruthSeeker, AI-TRAITS
& 50 claims / narratives
& Inputs used to generate machine-authored X-style disinformation posts.\\ 

&
MGO candidate samples
& \texttt{Llama-3.1-8B-Instruct}
& 250 posts
& Five candidate posts generated per input claim or narrative using zero-shot prompting with Llama-3.1.\\ 

&
MGO dataset
& \texttt{e5-small-lora}
& 125 posts
& MGO candidates classified as machine-generated by the \builders' baseline detector and retained for adversarial evaluation.\\

\hline

\multirow{2}{*}{\rotatebox[origin=c]{90}{\parbox[c]{3cm}{\centering \textbf{bld\_data}}}}&
human-authored text
& PHEME, Constraint
& 1,296 posts
& Human misinformation samples used for training and validation. The set has 639 posts from PHEME and 657 post from Constraint samples and is completely independent from the \breakers' set. \\ 
\cline{2-5}

& \multirow{4}{=}{machine-generated text}
& \texttt{grok-4-1-fast-non-reasoning}
& \multirow{4}{=}{1,296 posts}
& \multirow{4}{=}{The human misinformation samples are passed through an LLM embedded into persona-engineered prompts.} \\

& 
&  \texttt{Llama-3.3-70B-Instruct} 
& 
& \\

&
& \texttt{gpt-4o} 
&
& \\

&
& \texttt{DeepSeek-R1}
&
&\\

\noalign{\hrule height 0.75pt}
\end{tabular}
}
\end{center}
\renewcommand{\arraystretch}{1}

\paragraph{(III) Semantic Preservation Analysis.}
Transformed samples are passed through a semantic preservation pipeline that computes automatic similarity and consistency metrics between the original post and its transformed variant. The purpose of this stage is to quantify whether adversarial transformations preserve the original meaning and disinformation claim. The pipeline applies Natural Language Inference (NLI), token length ratio, E5 cosine similarity and BLEURT to characterise aspects of preservation, such as contradiction, semantic similarity, structural drift and generation quality. The pipeline is used as an automatic analysis layer rather than an exclusion mechanism and is described in more detail in \hyperref[sec:semantic_preservation]{Section~\ref*{sec:semantic_preservation}}. 

\paragraph{(IV) \texttt{Builders}' Evaluation.}


The transformed adversarial samples are passed to the \builders' detector, which classifies each post as human-generated or machine-generated (see \hyperref[sec:builders_approach]{Section~\ref*{sec:builders_approach}}). At this stage, the techniques employed by the \breakers~is unknown to the \builders. \texttt{Builders}' performance is evaluated using accuracy ($ACC$) and $F1$-score:\footnote{True Positive (TP) refers to an MGO correctly identified as machine-generated, while True Negative (TN) are the HGO correctly identified as human-generated. False Positive (FP) refers to an HGO incorrectly identified as machine-generated. False Negative (FN) refers to a machine-generated post incorrectly identified as human-generated. $N$ is the total number of samples.}

\begin{equation}
ACC =
\frac{\text{TP} + \text{TN}}
{N}
\label{eq:f1_score}
\end{equation}

\begin{equation}
F_1 =
\frac{2 \times \text{TP}}
{\text{2TP} + \text{FP} + \text{FN}}
\label{eq:f1_score}
\end{equation}

\noindent{A successful \builders' model has a high $F1$ score and accuracy, meaning the model makes few FP and FN errors.}

\noindent{Since the bkr\_data contains HGO and MGO samples that were initially classified correctly by the \builders' detector, $LFR$ is calculated as the inverse of accuracy on the transformed adversarial samples: }
\begin{equation}
LFR = 1 - ACC
\label{eq:lfr}
\end{equation} 

\noindent{where \(ACC\) refers to the \builders' model accuracy on the transformed adversarial samples. A high $LFR$ highlights that a \breakers' technique frequently causes initially correct \builders' predictions to become incorrect. A successful \breakers' transformation is therefore interpreted as one that causes a detector label to flip while retaining the original disinformation claim, with preservation assessed through the semantic preservation pipeline and manual evaluation (see \hyperref[sec:manual_evaluation]{Section~\ref*{sec:manual_evaluation}}).}

\paragraph{(V) Challenge Week.}
Results from the \builders' evaluation are manually reviewed to identify the \breakers' techniques that produce the largest changes in detector behaviour, including high $LFR$ and substantial drops in classification accuracy. The results from the \textit{challenge week} are reviewed by both teams, and observed trends are used to refine subsequent \builders' and \breakers' strategies. 
The pipeline is then repeated (except stage I): starting from II and III, followed by IV and V. In total, five iterations were completed. \hyperref[tab:iterations_overview]{Table~\ref*{tab:iterations_overview}} presents an overview of the \breakers' and \builders' techniques and data usage per iteration.   

\renewcommand{\arraystretch}{1.4}
\begin{center}
\cascaptionoftable
{Overview of the \bibir~activity with five iterations.}
{tab:iterations_overview}
{\sffamily\small
\begin{tabular}{@{}l| l c || l c}
\noalign{\hrule height 0.75pt}

& \multicolumn{2}{c||}{\texttt{Breakers}} & \multicolumn{2}{c}{\texttt{Builders}}\\
\textbf{Iter.}
& \textbf{Adversarial techniques} 
& \textbf{\# generated posts} 
& \textbf{Models}
& \textbf{Data size (train:dev:test)}\\ 
\hline

1 & D2, B2, C3, A3         & 12,000  &  baseline (\texttt{e5-small-LoRa}) & 2,560 (1,920:384:256)\\
2 & D1, D2, B2, C3, A3     & 90,000  & baseline + back-translation & 3,200 (2,560:384:256) \\
3 & B1, D1, D2, B2, C3, A3 & 180,000 & Siamese(1:1) Siamese(TF-IDF) &  2,560 (1,920:384:256)\\
4 & B3, D1, D3, B2, C3, A3 & 180,000 & Triplet(TF-IDF) &  2,560 (1,920:384:256)\\
5 & B3, D1, D4, B2, C3, A3 & 180,000 & Triplet(DASS) & 3,840 (2,880:576:384) \\

\noalign{\hrule height 0.75pt}
\end{tabular}}
\end{center}
\renewcommand{\arraystretch}{1}

\section{\texttt{Breakers}' Approach}
\label{sec:breakers_approach}

\begin{center}
    \includegraphics[width=1\linewidth]{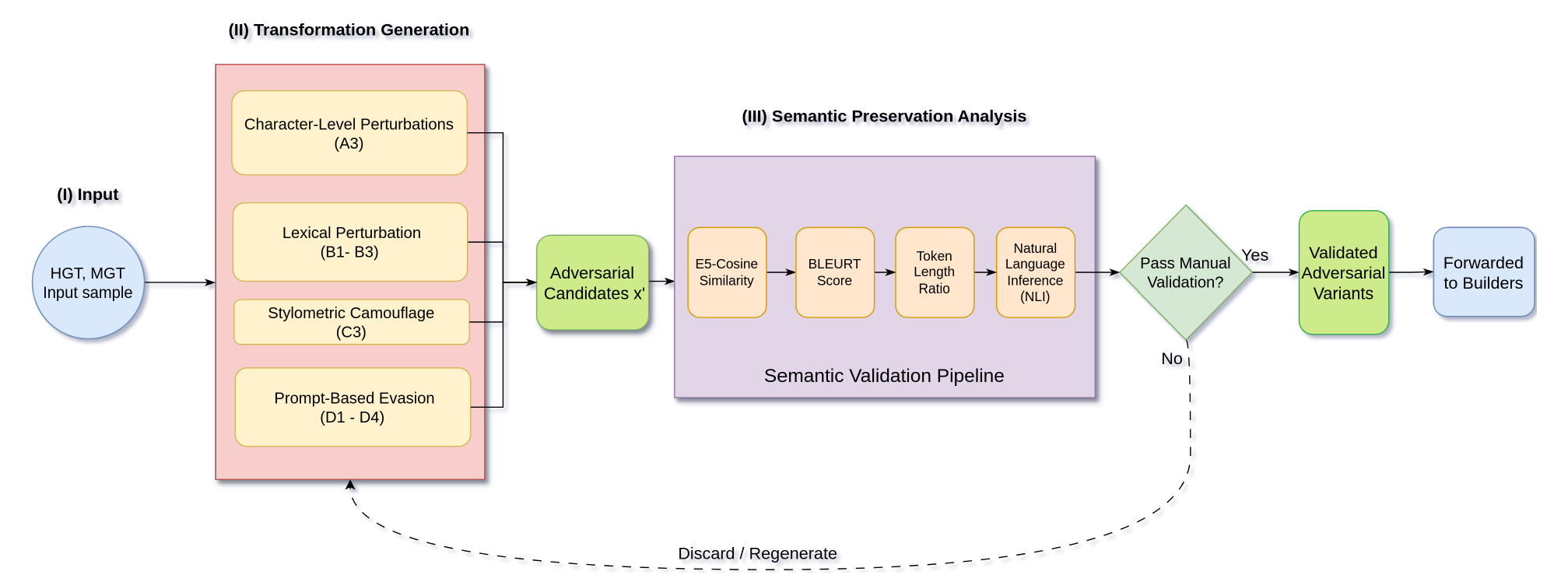}
    \cascaptionoffigure
    {Overview of the \breakers' methodology, showing how HGO and MGO input samples are transformed into adversarial variants, analysed using semantic preservation metrics, and passed forward for \builders' evaluation and performance analysis.}
    {fig:breaker_approach}
\end{center}

\renewcommand{\arraystretch}{1.4}
\begin{center}
\cascaptionoftable
{Implementation summary of \breakers' techniques, showing each attack family, technique code, transformation method, and model or tool used. 
}
{tab:technique_family}
{\sffamily\small
\begin{tabular}{@{}p{0.13\linewidth} l p{0.2\linewidth} p{0.55\linewidth}@{}}
\noalign{\hrule height 0.75pt}

\textbf{Technique Family} 
& \textbf{Code} 
& \textbf{Method / Tool} 
& \textbf{Implementation} \\ 
\hline

Character-Level Perturbation 
& A3 
& Rule-based whitespace editing
& Applies spacing edits including double spaces, spaces before punctuation, random newlines, and extra blank lines. \\ 
\hline

\multirow{3}{=}{Lexical Perturbation}
& B1 
& \texttt{LLaMA-3.1-8B-Instruct}
& Prompt-based paraphrasing to produce alternative wording while preserving the original claim. \\

& B2 
& OPUS-MT / NLLB 
& Translates English posts into an intermediate language and back into English using the routes in \hyperref[tab:b2_backtranslation_routes]{Table~\ref*{tab:b2_backtranslation_routes}}. \\

& B3 
& \texttt{LLaMA-3.1-8B-Instruct}
& Revised paraphrasing prompt designed to produce stronger lexical and syntactic variation than B1. \\ 
\hline

Stylometric Camouflage 
& C3 
& Emotion English DistilRoBERTa-base + emoji mapping
& Uses an emotion classifier to assign an emotion label, maps the label to an emoji set, and appends one sampled emoji where eligible. \\ 
\hline

\multirow{4}{=}{Prompt-Based Evasion}
& D1 
& \texttt{LLaMA-3.1-8B-Instruct}
& Rewrites the post using a tone-based prompting: \textit{You are currently in this mood: [TONE]. Rewrite the Twitter post provided to match your mood whilst maintaining the semantic meaning of the original post. [RULES]} \\

& D2 
& \texttt{LLaMA-3.1-8B-Instruct}
& Applies a simple persona-based prompting: \textit{You are a [PERSONA]. Rewrite the Twitter post the user provided to match your persona whilst maintaining the semantic meaning of the original tweet. [RULES]} \\

& D3 
& \texttt{LLaMA-3.1-8B-Instruct}
& Applies refined persona-based prompting: \textit{You are a [AGE] year old [ETHNICITY] [GENDER] living in the [LOCATION], who is currently a [OCCUPATION]. Rewrite the Twitter post the user provided to match your persona whilst maintaining the semantic meaning of the original tweet. [RULES]}. \\

& D4 
& LLaMA 3.1 8B Instruct
& Applies an indirect persona-based prompting: \textit{Think like a [AGE] year old [ETHNICITY] [GENDER] living in the [LOCATION], who is currently a [OCCUPATION]. Rewrite the Twitter post the user provided to match your persona whilst maintaining the semantic meaning of the original tweet. [RULES]}. \\

\noalign{\hrule height 0.75pt}
\end{tabular}
}

\end{center}

\renewcommand{\arraystretch}{1}

Given an input sample $x \in \{\text{HGO, MGO}\}$, \breakers' techniques produce adversarial variants $x'$ by applying transformations that modify the textual form of the original post while aiming to preserve its underlying disinformation claim. As shown in \hyperref[fig:breaker_approach]{Fig.~\ref*{fig:breaker_approach}}, the \breakers-side workflow consists of four steps: (I) starting with an HGO or MGO input sample, (II) applying one or more \breakers' transformations, (III) computing semantic preservation metrics for the original-transformed pair, and passing the resulting adversarial variant and preservation scores forward for \builders' evaluation and performance analysis. The implemented techniques are grouped into four families: character-level perturbation (A3), lexical perturbation(B1 to B3), stylometric camouflage (C3), and prompt-based evasion (D1 to D4). The full implementation taxonomy is summarised in \hyperref[tab:technique_family]{Table~\ref*{tab:technique_family}}.

Llama-3.1 (\texttt{Llama-3.1-8B-Instruct}) \citep{LLaMA3.1} was used for all LLM-based transformations, i.e. all techniques except the rule-based perturbation (A3 and C3) and back-translation (B2). This model was selected due to previous work using it for adversarial rewriting, AI-text detector evasion and disinformation-style generation \citep{ref19,ref76,ref77}. The 8B variant was as a practical trade-off between model capability and computational constraints. 

\subsection{Text Transformation}\label{sec:transformation}

\paragraph{Character-level perturbation (A3).}
Character-level perturbation is implemented as a rule-based technique that applies spacing edits to the input text. These edits include double-space insertion, spaces before punctuation, random newline insertion, and extra blank lines. The purpose of A3 is to identify whether the \builders’ detector is sensitive to low-level formatting changes that may disrupt tokenisation or alter surface-level text patterns while preserving the original claim \citep{ref59, ref60}.

\paragraph{Lexical perturbation (B1, B2, B3).}
Lexical perturbation includes paraphrasing (B1 and B3) and back-translation (B2). These techniques alter wording and sentence structure while aiming to preserve the original disinformation claim \citep{ref28, ref31}. B1 and B3 are implemented using LLaMA 3.1 8B Instruct, with text being rewritten to generate alternative phrasings of the original post (with B1 being more conservative and B3 more aggressive). B2 is implemented using back-translation, where the input is translated into an intermediate language and then translated back into English. Back-translation was implemented using pre-trained \texttt{OPUS\_MT} models \citep{opus} from the \texttt{Helsinki-NLP} HuggingFace collection,\footnote{\url{https://huggingface.co/Helsinki-NLP}} specifically those highlighted in \hyperref[tab:b2_backtranslation_routes]{Table~\ref*{tab:b2_backtranslation_routes}}. Japanese back-translation initially used the \texttt{OPUS\_MT} Japanese models, but generation was later switched to the NLLB model \citep{nllb}, also from HuggingFace,\footnote{\url{https://huggingface.co/facebook/nllb-200-distilled-600M}} due to mistranslation issues (frequently, the model hallucinated unrelated Bible-like text rather than translating the original input). A B2 technique is named as B2\_IL, where \textit{IL} is the intermediate language used in the back-translation pipeline. This category of techniques tests whether the \builders' models relies on lexical or syntactic regularities which detectors normally associate with machine-generated text.

\renewcommand{\arraystretch}{1.4}
\begin{center}
\cascaptionoftable
{Back-translation routes used for B2 lexical perturbation.}
{tab:b2_backtranslation_routes}
{\sffamily\small
\begin{tabular}{lll}
\noalign{\hrule height 0.75pt}

\textbf{Intermediate Language (IL)} 
& \textbf{English $\rightarrow$ IL} 
& \textbf{IL $\rightarrow$ English} \\ 
\hline

Arabic (AR)
& \texttt{Helsinki-NLP/opus-mt-en-ar}
& \texttt{Helsinki-NLP/opus-mt-ar-en} \\ 
\hline

Russian (RU)
& \texttt{Helsinki-NLP/opus-mt-en-ru}
& \texttt{Helsinki-NLP/opus-mt-ru-en} \\ 
\hline

Japanese (JA)
& \texttt{facebook/nllb-200-distilled-600M}
& \texttt{facebook/nllb-200-distilled-600M} \\ 
\hline

German (DE)
& \texttt{Helsinki-NLP/opus-mt-en-de}
& \texttt{Helsinki-NLP/opus-mt-de-en} \\ 
\hline

Chinese (ZH)
& \texttt{Helsinki-NLP/opus-mt-en-zh}
& \texttt{Helsinki-NLP/opus-mt-zh-en} \\ 

\noalign{\hrule height 0.75pt}
\end{tabular}
}
\end{center}
\renewcommand{\arraystretch}{1}

\paragraph{Stylometric camouflage (C3).}
Stylometric camouflage introduces emoji-based variation into posts. Unlike the LLM-based techniques, C3 is implemented programmatically. Each post is first passed through an emotion classifier (Emotion English DistilRoBERTa-base\footnote{\url{https://huggingface.co/j-hartmann/emotion-english-distilroberta-base}}) which assigns an emotion from the following label set: \textit{anger}, \textit{fear}, \textit{joy}, \textit{neutral}, \textit{sadness} or \textit{surprise}. The predicted emotion is then mapped to a predefined set of emojis, and one emoji is randomly sampled from the corresponding category. The sampled emoji is appended to the end of the post when the post does not already have an emoji and has at least two words.

\paragraph{Prompt-based evasion (D1, D2, D3, D4).}
Prompt-based evasion techniques are injected using LLaMA 3.1 8B Instruct with prompt-controlled rewriting to change the tone, persona, or rhetorical framing of the original post while preserving the underlying disinformation claim. D1 alters the emotional tone of the post, while D2, D3, and D4 use persona-based prompts of increasing refinement (as described in \hyperref[tab:technique_family]{Table~\ref*{tab:technique_family}}). \hyperref[tab:atomic_d_technique]{Table~\ref*{tab:atomic_d_technique}} shows the specific options for each D-type technique, specifying each subgroup with their corresponding atomic codes. As shown, four persona-types are defined: \textit{politician}, \textit{college student}, \textit{brain-rot 10-year old}, and \textit{four-start military general}. Attributes are stereotypical, since the aim of the study was to generate significantly altered text, instead of systematically assessing the capabilities of LLMs of dealing with different personas.

\paragraph{Chained transformations.}
In addition to individual attacks, selected \breakers' techniques are combined into chained transformations. In these cases, the output of one technique becomes the input to a subsequent technique. For example, a persona-based rewrite may be followed by back-translation and a spacing edit (which would be referred to as D3\_B2\_A3). Chained transformations are used to test whether combining attack families causes greater detector degradation than applying a single transformation in isolation. The order of each chained transformation follows the order indicated by the technique code.

\begin{center}
\cascaptionoftable
{Detailed taxonomy of D-type techniques, showing the options for prompts and the specific codes.}
{tab:atomic_d_technique}
{\sffamily\small
\begin{tabular}{@{} p{0.15\linewidth} p{0.15\linewidth} p{0.60\linewidth}@{}}
\noalign{\hrule height 0.75pt}
\textbf{Technique Family} 
& \textbf{Atomic Technique Code} 
& \textbf{Technique Description} \\ 
\hline

All & $-$ &
[RULES] = Rules for the rewritten tweet:
\begin{itemize}[label=--, nosep, leftmargin=*]
    \item Maximum character length: 200
    \item Do not change the semantic meaning
    \item Keep as much original content as possible
    \item Do not generate any text apart from the rewritten tweet
\end{itemize} \\ 
\hline

\multirow{5}{=}{D1}
& D11
& [TONE]= Joy \\

& D12
& [TONE]= Anger \\

& D13
& [TONE]= Sadness \\

& D14
& [TONE]= Surprise \\

& D15
& [TONE]= Neutral \\ 
\hline

\multirow{12}{=}{D2, D3, D4}
& \multirow{3}{=}{D21, D31, D41}
& Persona: \textit{Politician} 
\\
& & \multirow{2}{=}{[AGE] = 40; [ETHNICITY] = Caucasian; [GENDER] = Male; [LOCATION] = USA; [OCCUPATION] = Politician }\\
& 
&\\
\cline{2-3}

& \multirow{3}{=}{D22, D32, D42}
& Persona: \textit{College student} 
\\
& & \multirow{2}{=}{[AGE] = 20; [ETHNICITY] = Asian; [GENDER] = Female; [LOCATION] = USA; [OCCUPATION] = College Student}\\
& 
&\\
\cline{2-3}

& \multirow{3}{=}{D23, D33, D43}
& Persona: \textit{Brain-rot 10-year old } 
\\
& & \multirow{2}{=}{[AGE] = 10; [ETHNICITY] = African-American; [GENDER] = Male; [LOCATION] = USA; [OCCUPATION] = Elementary School Student}\\
& 
&\\
\cline{2-3}

& \multirow{3}{=}{D24, D34, D44}
& Persona: \textit{Four-star military general}
\\
& & \multirow{2}{=}{[AGE] = 60; [ETHNICITY] = Caucasian; [GENDER] = Male; [LOCATION] = USA; [OCCUPATION] = US Army General} \\
& 
&\\

\noalign{\hrule height 0.75pt}
\end{tabular}}
\end{center}

\subsection{Semantic Preservation Analysis}\label{sec:semantic_preservation}

To analyse whether \breakers' transformations preserve the meaning of the original post, each original-transformed pair is passed through a semantic preservation pipeline. This pipeline computes automatic similarity and consistency metrics, including Natural Language Inference (NLI) \citep{NLI}, token-length ratio, E5-cosine similarity \citep{E5}, and BLEURT \citep{BLEURT}. These metrics characterise different aspects of preservation: NLI categorises the posts as \textit{entailment} (the claim of the output matches the original), \textit{contradiction} (the claims do not match) or \textit{neutral} (there is not enough information to determine if the original and altered versions match); token-length ratio captures large structural changes; and E5-cosine and BLEURT estimate semantic similarity.

The semantic preservation pipeline is used as an automatic analysis layer rather than as a filtering mechanism. Automatic metrics can miss subtle changes in stance, implication, or claim framing, particularly for paraphrasing, back-translation, and persona-based rewriting. For this reason, the automatic preservation scores are later compared with manual evaluation scores for semantic preservation, naturalness, and preservation of disinformation claim, as is further discussed in \hyperref[sec:manual_evaluation]{Section~\ref*{sec:manual_evaluation}}.

\subsection{Manual Evaluation} \label{sec:manual_evaluation}

To ensure claims are maintained and the effectiveness of the filtering pipeline is validated, manual evaluation (ME) is conducted by human annotators within the team. The three criteria measured were \textit{(ME1) semantic preservation} -- measuring how well a modified text preserves the meaning and intent of the original text; \textit{(ME2) naturalness} -- measuring how fluent, human-like, and natural the altered text sounds in its context; and \textit{(ME3) preservation of disinformation claim} -- measuring whether the altered text retains the same misleading claim as the original without changing the core disinformation narrative. 

\hyperref[tab:manual_eval]{Table~\ref*{tab:manual_eval}} provides a snapshot of the guidelines used for manual evaluation of selected posts. The full annotation guidelines are provided in \hyperref[app:ME_guidelines]{Appendix~\ref*{app:ME_guidelines}}. For \textbf{ME1} and \textbf{ME2}, annotators were instructed to assign a midpoint score when uncertain between two adjacent scoring categories. For example, if an annotator was unsure whether a post should be scored as 2 or 1, they should assign a score of 1.5. This was not required for \textbf{ME3}, as its scoring scale already includes a midpoint value of 0.5 to represent partial claim preservation.

\renewcommand{\arraystretch}{1.4}
\begin{center}
\cascaptionoftable
{Manual evaluation scores for assessing transformed output posts across semantic preservation, naturalness, and disinformation claim retention.}
{tab:manual_eval}
{\sffamily\small
\begin{tabular}{@{}p{0.25\linewidth} p{0.08\linewidth} p{0.62\linewidth}@{}}
\noalign{\hrule height 0.75pt}
\textbf{Manual Evaluation Category} & \textbf{Metric} & \textbf{Description} \\
\hline

\multirow{3}{=}{ME1 -- semantic preservation}
& 3
& \textit{Identical meaning preserved}. The modified text has the same meaning as the original. The reader should draw the same conclusion from both texts, and the veracity of the claim should be unchanged. \\

& 2
& \textit{Partially changed}. This includes changes in certainty or strength of the claim, alteration in action, target, numbers, dates, or facts, and important details being removed or added. The text may contain partially true values, such as one truthful claim being preserved while another is changed, or ambiguity that affects interpretation. \\

& 1
& \textit{Irrelevant, completely unrelated, or opposite meaning}. The modified text contradicts or reverses the original meaning. This includes explicit negation, denial of the original claim, or text that no longer makes sense in comparison to the original version. \\
\hline

\multirow{4}{=}{ME2 -- naturalness}
& 4
& \textit{Natural and well-formed text}. The modified text flows smoothly, uses a natural tone and varied word choice, and does not contain repetitive patterns. The text reads fluently without irregularities. \\

& 3
& \textit{Mostly natural with minor awkwardness}. The modified text is fluent and understandable, but may contain mild redundancy or small grammatical stiffness. The text may feel slightly over-structured or less polished, but the overall structure remains coherent. \\

& 2
& \textit{Noticeably awkward or contains translation artefacts}. The modified text is understandable, but the word choice is poor, with unnatural synonyms, broken collocations, or awkward phrasing. \\

& 1
& C\textit{learly broken or incoherent}. The text is difficult or impossible to read, with severe grammatical errors, broken syntax, or fragmented output. \\
\hline

\multirow{3}{=}{ME3 -- preservation of disinformation claim}
& 1
& \textit{Claim held}. The output preserves the original disinformation claim and maintains the same misleading narrative, even if the wording, tone, or surface form has changed. \\

& 0.5
& \textit{Claim partially held}. The output retains part of the original disinformation claim, but weakens, softens, narrows, or slightly shifts the claim so that the misleading narrative is only partly preserved. \\

& 0
& \textit{Claim not held}. The output removes, contradicts, or substantially changes the original disinformation claim, meaning the core misleading narrative is no longer preserved. \\

\noalign{\hrule height 0.75pt}
\end{tabular}}
\end{center}
\renewcommand{\arraystretch}{1}

Manual evaluation was restricted to the MGO set since the objective of this paper is to assess whether adversarial transformations can make machine-generated disinformation appear human-written while preserving the original claim. Given the size of the dataset, it was not practical to manually evaluate all outputs for all \breakers' techniques. Instead, a targeted sampling approach was adopted, where the newest implemented technique family and variants were selected. For each selected variant, ten posts were manually evaluated: five posts that were classified by the detector as HGO and five posts that were classified as MGO. This selection helps check if the HGO classification given to a post is caused by successful adversarial transformation, rather than by cases where the output is classified as HGO due to the original disinformation claim being changed. 

Annotation was carried out by four Computer Science students, knowledgeable in the topic and co-authors for this paper. They annotated across the posts selected per iteration and using the guidelines specified in \hyperref[app:ME_guidelines]{Appendix~\ref*{app:ME_guidelines}}. 
To reduce the effect of inter-annotator variation, the scores assigned to each annotated post were averaged for each manual evaluation criterion. These averaged values were then used as the final \textbf{ME1}, \textbf{ME2} and \textbf{ME3} scores for each annotated post. The final manually evaluated set contained 780 posts across iterations 3, 4 and 5. The distribution of the manually evaluated posts is shown in \hyperref[tab:manual_eval_avgs]{Table~\ref*{tab:manual_eval_avgs}}.

\renewcommand{\arraystretch}{1.4}
\begin{center}
\cascaptionoftable
{Count of manual evaluation posts selected from each iteration, split by number of HGO-classified and MGO-classified posts.}
{tab:manual_eval_avgs}
{\sffamily\small
\begin{tabular}{lccc}
\noalign{\hrule height 0.75pt}

\textbf{Iter.} 
& \textbf{HGO-Classified Posts} 
& \textbf{MGO-Classified Posts} 
& \textbf{Total Posts} \\ 
\hline

3 & 145 & 145 & 290 \\
4 & 125 & 125 & 250 \\
5 & 120 & 120 & 240 \\ 
\noalign{\hrule height 0.75pt}

Total & 390 & 390 & 780 \\ 

\noalign{\hrule height 0.75pt}
\end{tabular}}
\end{center}
\renewcommand{\arraystretch}{1}

\section{\texttt{Builders}' Approach} \label{sec:builders_approach}

The \builders~adopted an iterative defence strategy in which the models were progressively strengthened in response to the \breakers' escalating attack approaches. At each iteration, a new model was introduced as a response to the previous round's of \breakers' strategy. As iterations progressed, the focus shifted from the baseline classifier model (\texttt{e5-small-LoRa}) to contrastive models that enforce paraphrase robustness and stylistic authorship cues. This section describes the various activities from the \builders' side: dataset preparation (\hyperref[sec:builders-dataset]{Section \ref{sec:builders-dataset}}), model development (Section \hyperref[sec:builders-dataset]{Section \ref{sec:builders-models}}) and experimental settings (\hyperref[sec:builders-dataset]{Section \ref{sec:experimental_setup}}).

\subsection{Datasets for training detectors} \label{sec:builders-dataset}

\begin{center}
    \includegraphics[width=0.9\linewidth]{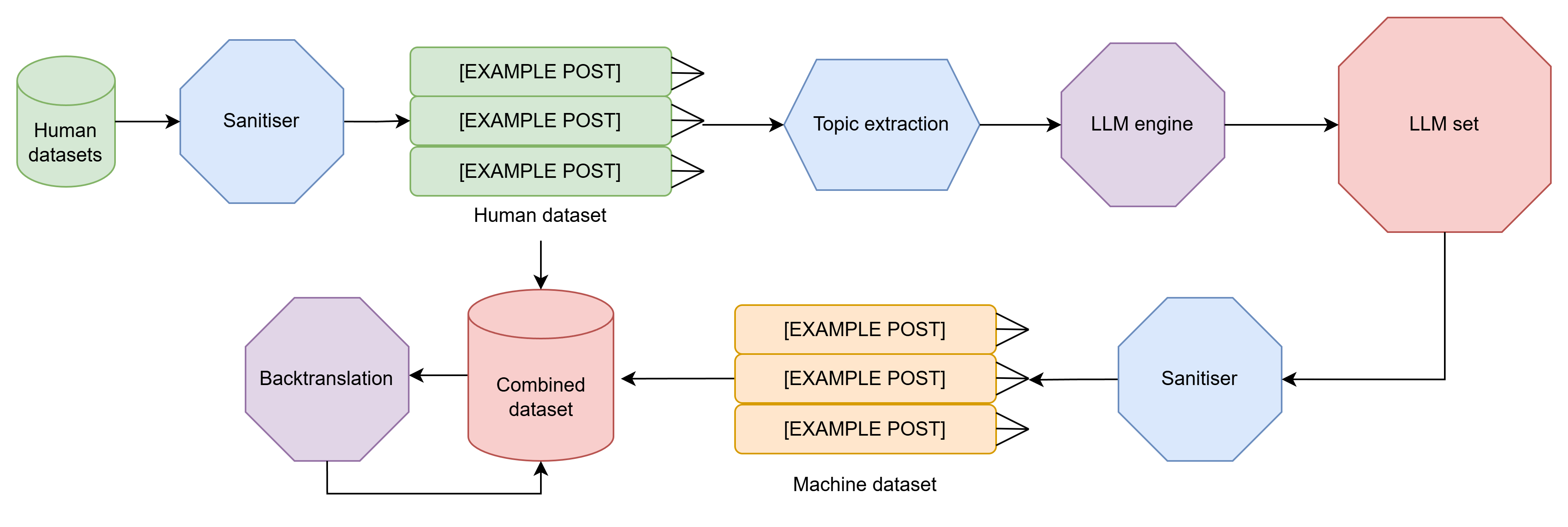}
    \cascaptionoffigure
    {\texttt{Builders}'s data pipeline.}
    {fig:data_pipeline_builders}
\end{center}

\hyperref[fig:data_pipeline_builders]{Fig. \ref{fig:data_pipeline_builders}} illustrates the full pipeline used by the \builders~to create a dataset suitable for training and validating their models. Starting from human data (disinformation appearances in social media), \builders˜apply multiple pre-processing techniques and prompt LLMs in order to generate machine-written counterparts for each human data point. This machine-generated data also pass through a sanitisation process, before being combined with the human data for the final dataset. 

\subsubsection{Human data}
For disinformation research on social media, the foundational data can be more critical than the model itself \citep{bengio2025internationalaisafetyreport}. Therefore, the human-generated set was built from two public datasets whose posts were published before the generative AI era (therefore, highly likely not machine generated): PHEME \citep{zubiaga2016pheme} and Constraint \citep{constraint2021dataset}. Both collections are tied to known rumours or fact-checked claims verified by researchers.

From the PHEME dataset, 638 tweets were selected after filtering out any samples for which the veracity label was ambiguous or where the text contained unwanted artefacts (e.g. non-English fragments, excessive token repetition). To prevent overfitting to a single event type, topic diversity was introduced with the Constraint set. A total of 658 scientific-health rumour tweets were sampled from the Constraint set using the same filtering criteria to complement the breaking news events covered by PHEME. Both sources were kept in roughly equal proportion so that neither domain would dominate in the training stage. The final human-written dataset contained 1,296 clean posts.


\subsubsection{Machine-generated data} \label{sec:builder_MGO}
For each sample from the human set, a machine-generated counterpart was synthesised using tweet injection, 
i.e. the full text of the human post was embedded inside a specific persona-engineered prompt and fed to an LLM to create new logical pathways while maintaining the misinformation claim of the original HGO sample.
To prevent the detector from overfitting to a single surface style, ten distinct prompt templates were designed. They used a wide range of personas: e.g. \textit{a careless phone user}, \textit{an objective news report}, and \textit{a jaded internet commentator}. Each template instructed the model to alter the sentence structure, starting point, and word choice to create a new machine-generated text and not just perform simple synonym swaps. Additionally, all templates enforced the same negative constraints (placeholders, no emojis) that mirror the sanitisation pipeline (\hyperref[builder_sanitisation]{Subsec.~\ref*{builder_sanitisation}}). These constraints were enforced during generation time so that the output maintains a coherent sentence structure, as stripping these elements at post-processing would create unnatural gaps, fragments, or artificially shortened posts that the detector could easily learn to exploit. An example template is shown below (an exhaustive list of the ten templates is provided in \hyperref[app:builder_prompts]{Appendix.~\ref*{app:builder_prompts}} \hyperref[fig:prompt]{Fig. \ref{fig:prompt}}):

\noindent{
\begin{center}
\hspace{-0.46cm}    
\begin{minipage}[h]{0.98\linewidth} 
      \fbox{ 
          \parbox{\linewidth}{ 
          \small{
          You are a distracted social media user typing quickly while scrolling. Read the source text: \texttt{<TWEET>}. Rewrite the same information as a spontaneous Twitter/X post that feels messy, reactive, and slightly rushed. \\
          
          CRITICAL RULES: \\
    
1. Meaning Preservation: Keep the exact core message of the original tweet, but restructure how the idea is expressed. Do not simply paraphrase word-by-word.\\

2. Human Imperfections: Slightly awkward phrasing, inconsistent capitalization, or minor grammar slips are allowed. Avoid polished AI-style structure. \\

3. Formatting: Output ONLY the tweet text. No quotes, explanations, labels, or additional commentary. \\

4. Length: Keep the response approximately the same length as the original tweet. Do not expand the idea. \\ 

5. Tokens: The original text uses placeholders like <USER> and <URL>. You MUST use these exact
placeholders instead of @ mentions or links. For hashtags, you must format them as <HASHTAG> word (e.g., instead of \#breakingnews, write <HASHTAG> breakingnews). Do not use the actual \# or @ symbols anywhere in your response. \\

6. No Emojis: Absolutely no emojis or special symbols.}
          }
      }
  \label{fig:prompt}
  \end{minipage}

\end{center}
}
    
    






These templates were refined through reverse prompt engineering, where the candidate prompts were tested, and the results were analysed for style and constraint adherence. The prompt was iteratively adjusted until the generated tweets would reliably match the intended persona while respecting all constraints. 

Four instruction-tuned LLMs were used to generate the machine-generated dataset: Grok-4.1 (\texttt{grok-4-1-fast-}\\
\texttt{non-reasoning}) \cite{grok4}, Llama 3.3 (\texttt{Llama-3.3-70B-Instruct}) \cite{LLaMA3.1}, GPT-4o (\texttt{gpt-4o}) \cite{openai2024gpt4ocard}, and DeepSeek-R1 (\texttt{DeepSeek-R1}) \cite{Guo_2025}. These models were selected for stylistic diversity: GPT-4o and Llama-3.3  provided a reliable but slightly different instruction-following baseline; Grok-4.1, being more filtered, tended to produce more authentic misinformation outputs; DeepSeek-R1 followed its chain-of-thought reasoning which often generated more novel and coherent structures.  Each prompt, for each of the 1,296 human-written tweets, was assigned to a single model in a fixed repeating order (Grok-4.1 $\rightarrow$ Llama-3.3 $\rightarrow$ GPT-4o $\rightarrow$ DeepSeek-R1), so that each model had, roughly, an even distribution of the total prompts. This preserved the 1:1 human to machine ratio while ensuring that no single model's style dominated the machine-generated set. Generation was carried out via the Azure inference endpoint with a temperature of 0.3 across all models. Standard models had a max token limit of 400, while DeepSeek-R1 had 2,000 to support its deep-reasoning responses. Due to the conflict of spreading misinformation and sophisticated LLM guardrails, certain generation attempts produced invalid results. A fallback mechanism was then added: if the model failed after repeated retries, another model from the remaining set was used. 
Upon reviewing the machine-generated samples for residual AI artefacts, claim drift, or redundancy, 16 data points were discarded along with their human-written counterparts, leaving \textbf{1,280 human-machine tweet pairs}.

\subsubsection{Data augmentation and preprocessing}\label{sec:aug_prep}

\paragraph{Back-translation augmentation.} 
In order to support attacks of type B2, back-translation was applied to $25\%$ of the combined human- and machine-generated training dataset by selecting every fourth sample and passing it through one of the language routes shown in \hyperref[tab:bt_paths]{Table~\ref*{tab:bt_paths}}. This subset was chosen to introduce lexical and syntactic variation without allowing synthetic back-translated samples to dominate the original training distribution. Selecting every fourth sample also helped preserve coverage across the dataset and maintain topic alignment between corresponding human-machine examples.
All translations were performed using the Microsoft Azure Translator Text API v3.0,\footnote{\url{https://learn.microsoft.com/en-us/azure/ai-services/translator/}} which applies neural machine translation (NMT) by default. 
\hyperref[tab:bt_paths]{Table~\ref*{tab:bt_paths}} displays the 14 back-translation routes used for augmentation: nine single language routes and five multi-path routes. Single language routes introduced minor lexical and syntactical variation, while multi-path produces stronger perturbations by translating through multiple intermediate languages before returning to English. The resulting back-translated samples were concatenated to the human-machine training dataset.

\renewcommand{\arraystretch}{1.35}
\begin{center}
\cascaptionoftable
{Back-translation routes used for augmentation. All routes were executed using Microsoft Azure Translator Text API v3.0 with default neural machine translation.}
{tab:bt_paths}
{\sffamily\small
\begin{tabular}{lll}
\noalign{\hrule height 0.75pt}

\textbf{Route Type}
& \textbf{Intermediate Language(s)}
& \textbf{Translation Route} \\
\hline

Single
& German (DE)
& \texttt{en $\rightarrow$ de $\rightarrow$ en} \\
\hline

Single
& French (FR)
& \texttt{en $\rightarrow$ fr $\rightarrow$ en} \\
\hline

Single
& Spanish (ES)
& \texttt{en $\rightarrow$ es $\rightarrow$ en} \\
\hline

Single
& Arabic (AR)
& \texttt{en $\rightarrow$ ar $\rightarrow$ en} \\
\hline

Single
& Russian (RU)
& \texttt{en $\rightarrow$ ru $\rightarrow$ en} \\
\hline

Single
& Japanese (JA)
& \texttt{en $\rightarrow$ ja $\rightarrow$ en} \\
\hline

Single
& Chinese (ZH)
& \texttt{en $\rightarrow$ zh-Hans $\rightarrow$ en} \\
\hline

Single
& Hindi (HI)
& \texttt{en $\rightarrow$ hi $\rightarrow$ en} \\
\hline

Single
& Turkish (TR)
& \texttt{en $\rightarrow$ tr $\rightarrow$ en} \\
\hline

Multi-path
& German + French
& \texttt{en $\rightarrow$ de $\rightarrow$ fr $\rightarrow$ en} \\
\hline

Multi-path
& Arabic + French
& \texttt{en $\rightarrow$ ar $\rightarrow$ fr $\rightarrow$ en} \\
\hline

Multi-path
& Japanese + Chinese
& \texttt{en $\rightarrow$ ja $\rightarrow$ zh-Hans $\rightarrow$ en} \\
\hline

Multi-path
& Russian + German
& \texttt{en $\rightarrow$ ru $\rightarrow$ de $\rightarrow$ en} \\
\hline

Multi-path
& Hindi + Arabic
& \texttt{en $\rightarrow$ hi $\rightarrow$ ar $\rightarrow$ en} \\

\noalign{\hrule height 0.75pt}
\end{tabular}
}
\end{center}
\renewcommand{\arraystretch}{1}



\paragraph{Paraphrased machine-generated data for DASS.}
In addition to the human tweet (\(T_H\)) and its machine-generated counterpart (\(T_L\)), the triplet network with DASS required a paraphrased variant of the machine-generated tweet (\(T_P\)). To generate the paraphrased set, separate prompts were used. In total, five new prompt templates were engineered with a tweet injection token (\hyperref[app:builder_prompts]{Appendix~\ref*{app:builder_prompts}}). Each template introduced a distinct persona or tonal shift while maintaining the same constraints as the original machine-generated text.
Although the same four LLMs were used, the assignment order was rotated (GPT-4o $ \rightarrow $ Llama-3.3$ \rightarrow $ Grok-4.1 $ \rightarrow $ DeepSeek-R1). Crucially, since the rotation for paraphrasing was offset from the original generation assignment, the model selected for paraphrasing was always different from the model used for the original machine-generate set. This forced the paraphrase to come from a different LLM family or instruction-tuning style, minimising the risk of over-fitting to model-specific artefacts. For this experiment, the generation temperature was 0.85 (higher than the 0.3 from the original set) to encourage greater lexical and syntactic diversity. After the same manual quality review, all 1,280 paraphrased machine-generate posts were selected for the DASS paraphrased set.

\paragraph{Sanitisation.} \label{builder_sanitisation}
To standardise the training data and minimise the influence of non-semantic noise, a preprocessing script was used to sanitise the text. This script removed emojis, redundant whitespace and normalised URLs and X handles.
Emojis were removed because their distribution was inconsistent across the machine-generated samples, where some generated posts contained many, and others contained none. This mitigated the risk of introducing an imbalanced shortcut feature that the model could associate with a particular class. Emojis are also context-dependent, as the same emoji may convey different meanings depending on the surrounding text, therefore, removing them is expected to reduce the risk of the model relying on emoji presence rather than text-based authorship patterns.

To reduce the effect of homoglyph-style perturbations, Cyrillic, Greek and accented characters were converted to their closest ASCII equivalent. This mitigates attacks where characters that resemble standard English letters are inserted to alter tokenisation, despite it actually preserving visual readability. User handles and URLs were also normalised to fixed tokens: <USER> and <URL>, respectively (e.g. \textit{"hello @john you should visit https://google.com"} $\rightarrow$ \textit{"hello <USER> you should visit <URL>"}). 
This preserves the information of the presence of a user mentioned or URL, while preventing the model from overfitting to specific usernames, domains or links. Since the dataset was constructed from both human and machine data sources, this standardisation helped make the input format more consistent and supported better generalisation to unseen data. 
Hashtags were treated differently because they may contain semantic information. Therefore, instead of removing/replacing the content, each hashtag was split into a marker token and its textual content (e.g.  \textit{"england is going to win the world cup \#threelions \#bottlers"} $\rightarrow$ \textit{"england is going to win the world cup <HASHTAG> threelions <HASHTAG> bottlers"}). This preserves both the presence of a hashtag and the meaning carried by its content. For example:

Finally, repeated whitespace and redundant spacing were removed as they were by-products of preprocessing, user typing behaviour, machine-generation errors or adversarial perturbations. Since excess white space doesn't contribute to semantic meaning, removing it helped standardise the input without discarding relevant content.

\paragraph{Data Split.} \label{sec:builder_data_split}
The final \builders~dataset (bld\_data) was split into training, validation and test sets using a 75:15:10 ratio as seen in \hyperref[tab:builder_data_split]{Table~\ref*{tab:iterations_overview}}. The test set was used only for static benchmark evaluation and it is worth emphasising that it has no overlap with the \breakers's evaluation set.

\paragraph{Independence from \breakers.}
It is worth emphasise that all datasets were created entirely independently from the \breakers~to avoid data leakage. The \builders~chose prompts, LLMs, and generation parameters without any knowledge of which models or attack techniques the \breakers~would subsequently use. Although the LLaMA 3.1 (8B) instruct LLM was also used by the \breakers, the \builders' machine-generated set relied on a significantly larger 70B variant. No samples produced by the \breakers~were ever introduced into the \builders~sets. By following a strict zero-knowledge boundary, both teams were able to use a unique data pipeline to prevent any training data leakage.
In addition, both the main machine-generated set and the DASS paraphrased set were produced using fixed prompts and models outside of the \breakers' adversarial transformation pipeline. 

\subsection{Model Architectures} \label{sec:builders-models}
The Builders' framework explored three model architectures, comprising a baseline supervised approach and two contrastive learning variants.
The \textbf{\textit{baseline detector}} employs the \texttt{e5-small-LoRA} model\footnote{\url{https://huggingface.co/MayZhou/e5-small-lora-ai-generated-detector}} fine-tuned for AI-generated text detection task using LoRA \cite{LoRA-Adaptation}, a parameter efficient fine-tuning technique. Its choice was motivated by its strong performance on the RAID benchmark leaderboard\footnote{\url{https://raid-bench.xyz/leaderboard}} (ranked 5th on Reddit data when considering only open-source models), model availability at HuggingFace 
and its relatively low computational requirements for fine-tuning. 

As discussed in Section~\ref{rw:supervised}, \textit{\textbf{contrastive learning}} was previously shown to be effective for binary machine-generated text detection task. To explore its effectiveness for our task, we applied siamese and triplet networks to the baseline model.


\paragraph{\textbf{Siamese networks.}}
We hypothesise that building a siamese architecture model would combat aggressive paraphrasing attacks, as it can distinguish stylistic characteristics between human- and machine-generated texts rather than relying on surface-level tokens and signals. Such a pairwise architecture can be particularly successful in adversarial settings as it can be taught to associate a machine-generated text with its adversarially modified variants (e.g. the insertion of white spaces or back-translated texts), which can help the network to filter out noise and to recognise and classify the sample as if it was the original.

In contrastive learning, randomly chosen negatives can quickly become inefficient as the model learns to separate classes with a wide margin \citep{Schroff_2015}. Once a negative sample is pushed beyond the established margin, its contribution towards the loss function drops to zero, stalling the contrastive model's optimisation process. For a contrastive model to be effective, the loss function must maintain a useful gradient. Therefore, the learning model should actively select samples that are close to the anchor in the embedding space but belong to the opposite class (i.e. hard negatives) in order to fall within the margin boundary and force the model to refine its vector representation. This is crucially demonstrated in the Valla benchmark \citep{tyo-etal-2023-valla} which shows that by using hard negative mining to enable verification models, they can become a highly competitive alternative to attribution methods.

\subparagraph{\textit{1:1 pairing hypothesis.}} Since each machine-generated post in bld\_data was generated from the corresponding human-written misinformation post, the dataset contains naturally aligned human-machine pairs that preserve the same underlying topic and claim. This setup reduces the risk of the Siamese model learning topic-specific shortcuts, because each comparison is made between posts with closely matched content. The intended effect was to make the model distinguish human-written from machine-generated text using authorship-related features, such as phrasing, syntax, and stylistic structure, rather than broad vocabulary differences. Under this hypothesis, 1:1 pairing offers a controlled contrastive baseline before introducing more complex hard negative mining strategies. This pairing strategy was used for the Siamese (1:1) model.

\subparagraph{\textit{TF-IDF pairing hypothesis.}} While strict 1:1 pairing preserves topical alignment, the negative samples created for the Siamese model may remain easily separable if the machine-generated tweets significantly differ from the human-written tweets, causing the model to learn broad stylistic and vocabulary dependent differences. 
Therefore, we additionally explore a more effective hard negative mining strategy through similarity-based retrieval. It enforces topical proximity between the anchor and the negative. If an anchor and a negative share the same topic, the network is forced to ignore broad vocabulary shortcuts to focus strictly on stylistic artefacts. 
Term Frequency-Inverse Document Frequency (TF-IDF) provides a robust foundation to enforce topical proximity. By using TF-IDF scores to isolate term importance in an Unsupervised Hard Negative Augmentation (UNA) system, which generates synthetic hard negatives through salient word perturbation, the performance on semantic textual similarity tasks was improved \citep{shu2024unsupervisedhardnegativeaugmentation}. Since TF-IDF naturally captures topic similarity, it has been shown to retrieve existing texts that form challenging negatives. This strategy is relevant for short-form disinformation, where human and machine-generated tweets on the same narrative can be difficult to distinguish by the content alone and require models to learn the underlying stylistic cues.
 For our task, a TF-IDF-based hard negative would create more challenging contrastive comparisons of pairs, as it would pair semantically similar posts from the opposite class. As the TF-IDF pairing method finds pairs with overlapping vocabulary and narrative framing, the model is forced to rely less on lexical shortcuts and instead learn deeper stylistic patterns associated with machine-generate or human-written texts.  In practice, 
TF-IDF pairing ensures that the model focuses on the underlying style instead of the subject by grouping similar topics closely in the vector space, leaving minimal topical differences for the learning model to exploit. When the pairs are created, the dataset is grouped into topic buckets using TF-IDF and $k$-nearest neighbour ($k$-NN) selection: for each sample, the top-20 ($k$=20) most similar tweets are identified via cosine similarity of TF-IDF vectors, filtering out common stop words (e.g. "the", "and") to highlight relevant keywords (e.g. "Prince") that define the core topic of a tweet. When constructing input pairs, both samples are taken from the same topic bucket to force the model to compare human and machine-generated texts on the same topic. This teaches the network underlying stylistic and syntactical signatures of LLMs instead of generic topic differences.
This pairing strategy was used for the Siamese (TF-IDF) models.

\paragraph{\textbf{Triplet networks.}} \label{app:triplet_network} 
As described in Section~\ref{rw:supervised}, triplet networks extend the contrastive pairing architecture of siamese networks by evaluating three samples simultaneously: an anchor, a positive sample from the same class as the anchor, and a negative sample from the opposing class (\hyperref[fig:siamese-triplet]{Fig.~\ref*{fig:siamese-triplet}}). By utilising a \textit{triplet margin loss}, the architecture forces the embedding space to divide into two distinct classes by pulling the anchor to the positive sample and pushing the negative beyond a fixed margin, forming a clean cluster separation. In the context of machine-generated text detection, as the underlying space actively encodes the embeddings of source identity instead of surface string similarity, the model can more reliably identify both original and adversarial paraphrased machine-generated text. In our experiments, we explore two types of sampling strategies for Triplet networks.

\subparagraph{\textit{Triplet with TF-IDF.}} In this setting, the triplet is trained using the same TF-IDF $k$-NN pairing engine introduced, previously in the TF-IDF pairing hypothesis. For every anchor sample, the top-20 most similar samples are identified via cosine similarity, where a hard positive is taken from the tweets sharing the anchor's authorship label and a hard negative from those with the opposite label. By enforcing a tight semantic cluster, the margin loss can only be reduced by learning the underlying stylistic differences, effectively neutralising vocabulary bias. The idea is that by using the same pairing engine with a triplet network (three-way contrast), the model should yield a more discriminative embedding space. This pairing strategy was used for the Triplet (TF-IDF) models.

\begin{center}
    \includegraphics[width=0.7\linewidth]{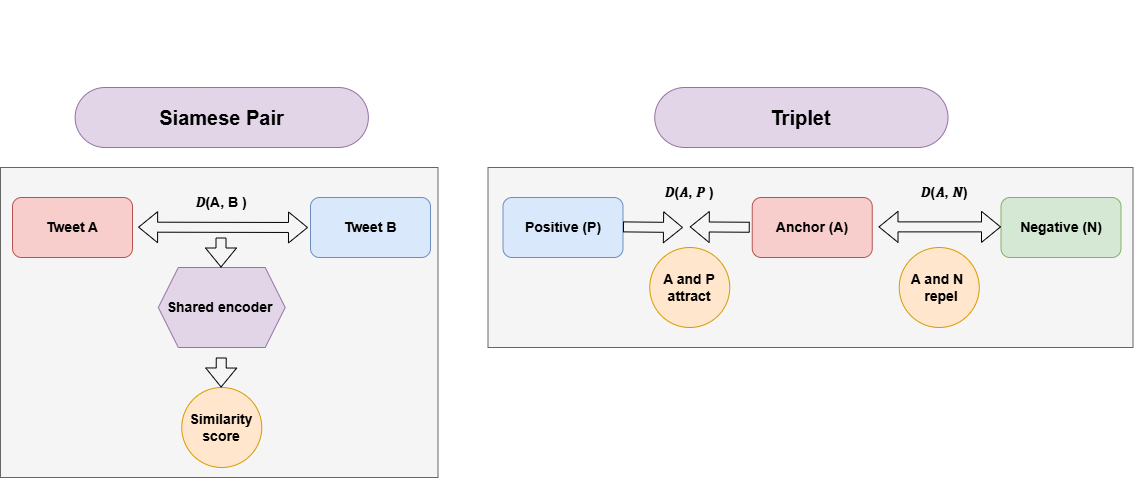}
    \cascaptionoffigure
    {Siamese vs. triplet pairing.}
    {fig:siamese-triplet}
\end{center}




\begin{center}
    \includegraphics[width=1\linewidth]{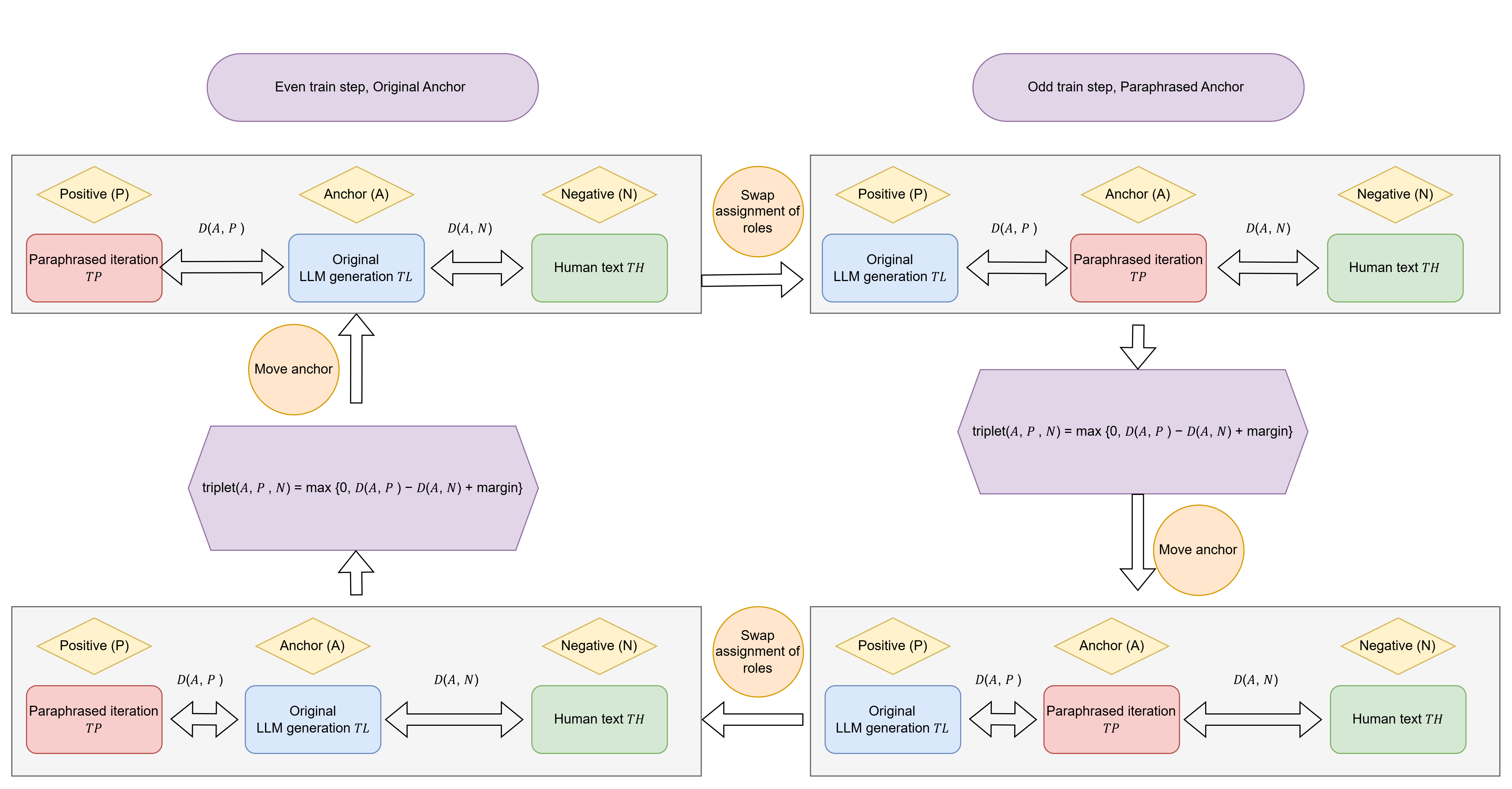}
    \cascaptionoffigure
    {DASS pairing engine.}
    {fig:dass_pairing_fig}
\end{center}

\subparagraph{\textit{Triplet with DASS semantic pairing.}}\label{DASS Pairing} A more targeted approach adopts the dynamic anchor switching strategy (DASS), a pairing strategy introduced by GravText \citep{GravText-Triplet} to make detectors robust against paraphrasing. In standard triplet learning, the anchor is fixed, whilst DASS alternates the role of the anchor depending on the detection task (\hyperref[fig:dass_pairing_fig]{Fig.~\ref*{fig:dass_pairing_fig}}). Given a triplet of samples: a original machine-generated text \(T_L\), its paraphrased iteration \(T_P\), and a human text \(T_H\) on the same topic, the standard triplet loss is:
\begin{equation}
    \mathcal{L}_{\text{triplet}}(A, P, N) = \max \left\{ 0, D(A, P) - D(A, N) + \text{margin} \right\}
    \label{eq:tripelt_loss}
\end{equation}
where,
\(A =\) anchor \(= T_L\),
\(P =\) positive \(= T_P\),
\(N =\) negative \(= T_H\),
\(D =\) Euclidean Distance between embeddings.

\noindent{In standard training, the anchor is fixed, but DASS switches the assignment:}

\begin{itemize}
    \item \textbf{Paraphrased Anchor ({PA}) (human or machine-generated sample).} The loss equation becomes:
\begin{equation}
    \mathcal{L}_{\text{triplet}}^{PA} = \max \left\{ 0, D(T_P, T_L) - D(T_P, T_H) + \text{margin} \right\}
    \label{eq:loss_pa}
\end{equation}
The original and the paraphrased sample are pulled together while separating from the human text space. 

    \item \textbf{Original Anchor ({OA}) (human or paraphrased  machine-generated sample).} The loss equation becomes:
\begin{equation}
    \mathcal{L}_{\text{triplet}}^{OA} = \max \left\{ 0, D(T_L, T_P) - D(T_L, T_H) + \text{margin} \right\}
    \label{eq:loss_oa}
\end{equation}
The model learns to recognise the paraphrased sample as still machine-generated by anchoring it to the original output space. 
\end{itemize}

\noindent{LLM-generated outputs (original or paraphrased) should cluster tightly together and remain separated from the human-generated space by alternating between the two loss functions during training. In the case of tweet misinformation, \(T_L\) is a machine-generated tweet on a given narrative, \(T_P\) is its paraphrased counterpart (or a semantically close example), and \(T_H\) is a human tweet on the same topic. This adaptation directly follows the DASS pairing method, but disregards the gravitational factor introduced the full GravText framework. 
The implemented model is called Triplet (DASS).}

\subsection{Experimental Setup}
\label{sec:experimental_setup}

\hyperref[tab:builder_hyperparams]{Table~\ref*{tab:builder_hyperparams}} summarises the hyper-parameters for all \builders' models.



\renewcommand{\arraystretch}{1.4}
\begin{center}
\cascaptionoftable
{\texttt{Builders'} hyper-parameters per model.}
{tab:builder_hyperparams}

{\sffamily\small
\begin{tabular}{@{}p{0.18\linewidth} p{0.09\linewidth} p{0.12\linewidth} p{0.11\linewidth} p{0.12\linewidth} p{0.13\linewidth} p{0.12\linewidth}@{}}
\noalign{\hrule height 0.75pt}

\raggedright\textbf{Hyperparameter} &
\raggedright\textbf{Baseline} &
\raggedright\textbf{Baseline + Back-translation} &
\raggedright\textbf{Siamese (1:1)} &
\raggedright\textbf{Siamese (TF-IDF)} &
\raggedright\textbf{Triplet (TF-IDF)} &
\raggedright\textbf{Triplet (DASS)} \tabularnewline
\hline

Learning rate
& \raggedright $2.286 \times 10^{-5}$
& $3.071 \times 10^{-5}$
& $2 \times 10^{-5}$
& $2 \times 10^{-5}$
& $2 \times 10^{-5}$
& $2 \times 10^{-5}$ \\

Weight decay
& 0.2889
& 0.2483
& 0.01
& 0.01
& 0.01
& 0.01 \\

Epochs
& 6
& 4
& 3
& 3
& 3
& 3 \\

Batch size
& 16
& 8
& 8
& 8
& 8
& 8 \\

\raggedright Loss Function
& CE
& CE
& BCE
& BCE
& \raggedright Triplet (L2, m=1)
& \raggedright Triplet (L2, m=1) \tabularnewline

Pairs/Triplets (epoch)
& -
& -
& 20,000
& 20,000
& 20,000
& 15,000 \\

TF-IDF top-k
& -
& -
& -
& 20
& 20
& - \\

\hline
\noalign{\hrule height 0.75pt}
\end{tabular}
}

\end{center}
\renewcommand{\arraystretch}{1}


\paragraph{Common Configuration.}
All models used the HuggingFace's \texttt{MayZhou/e5-small-lora-ai-generated-detector} checkpoint and its tokeniser. Each sample was passed through the same sanitisation pipeline described in \hyperref[builder_sanitisation]{Section~\ref*{builder_sanitisation}} and truncated to a maximum of 256 tokens to reflect the nature of the X short-form domain. Models were trained using the HuggingFace's Trainer API\footnote{\url{https://huggingface.co/docs/transformers/en/main_classes/trainer}} that utilises an AdamW optimiser and a linear learning-rate schedule. Each model had a fixed random seed of 42 across Python, NumPy, and PyTorch to ensure reproducibility. Every training run and data preprocessing was executed using a single NVIDIA GeForce RTX 4060 Laptop GPU, AMD Ryzen 7 7840HS CPU, and 32GB of system RAM (5600 MT/s). 

\paragraph{Data Partitioning.}
As mentioned in \hyperref[sec:builder_data_split]{Section~\ref*{sec:builder_data_split}}, the bld\_data was split into 75\% training, 15\% validation, and 10\% test set; using a fixed random seed of 42, creating interleaved topic indices and ensuring no data leakage.  Final sample sizes varied according to the model and its pairing engine. The standard baseline classifier, 
both Siamese models (1:1 and TF-IDF) and the Triplet (TF-IDF) used the raw bld\_data, with 2,560 samples (1,280 human-written and 1,280 machine-generated). For Triple(DASS), an additional paraphrased machine-generated set based on the bld\_data's machine-generated set (\hyperref[sec:aug_prep]{Section~\ref*{sec:aug_prep}}) was introduced, yielding a total of 3,840 samples (1,280 human-written, 1,280 machine-generated, and 1,280 paraphrases of the machine-generated set). Notably, an experiment in iteration 2 augmented a subset of the training data using back-translation, resulting in a larger set of 3,200 samples (1,600 human-written, 1,600 machine-generated). However, since the experiment showed a reduction in performance, the follow up models were trained on the non-augmented dataset. 

\paragraph{Evaluation.}
During training, the HuggingFace's Trainer API selected the baseline model’s best checkpoint based on the highest validation-set $F1$-score achieved at the end of each epoch. For the Siamese models, validation was performed via pair-wise accuracy and $F1$ on the validation-set pairs, and the checkpoint with the highest pair-wise $F1$ was selected. The Triplet model needed a class prototype computed from the training-set embeddings and the validation samples after each epoch, as it cannot produce a class prediction from the triplet loss alone. The validation samples were assigned to the nearest prototype, using the prototype's $F1$ as the selection metric.

Finally, for internal model assessment, all models were evaluated on bld\_data's test set, which was not used during training or hyperparameter selection.\footnote{The baseline classifier was run with a 10-trial Optuna hyper-parameter search} Evaluation metrics where accuracy and $F1$. The baseline classifier directly outputs a label, while the contrastive models used the same prototype classification method applied to the test set: each test sample was embedded, compared via cosine similarity to the training set prototypes, and assigned to the closer class.

%
%
\section{Results}
\label{sec:results}
\subsection{Impact of adversarial transformations in the detector's performance (RQ1)} \label{sec:rq1}


To answer RQ1, we use the bkr\_data set, with 125 HGO-MGO pairs. The detector is also fixed: we use the baseline \texttt{e5-small-LoRa}, which showed 100\% accuracy on the bkr\_data set. This decision was made to ensure that only the adversarial techniques were being assessed and no changes made by the \builders~could influence the results. 
In addition, only the MGO set is considered, as the HGO dataset is only used as an anchor to assess the performance of \builders' models. 

In total, there were five iterations (\hyperref[tab:iterations_overview]{Table~\ref*{tab:iterations_overview}}). Iterations 1 and 2 are not reported as separate result stages because they were exploratory development rounds with a small number of generated posts. Due to this, outputs from iterations~1 (\(n = 12{,}000\)) and 2 (\(n = 90{,}000\)) were incorporated into the iteration~3's analysis, meaning that iteration~3 contains \breakers' results across iterations~1--3 (\(n = 180{,}000\)). This provides a more reliable basis for comparing early \breakers' techniques against the later iterations~4 (\(n = 180{,}000\)) and 5 (\(n = 180{,}000\)) results, which were comparatively larger. 
The combinations of techniques were decided by generating a power set of the technique codes for each iteration, as shown in \hyperref[tab:iterations_overview]{Table~\ref*{tab:iterations_overview}}. This was then represented as a $2^n$ binary truth table, where $n$ is the number of techniques for the iteration, ensuring an exhaustive exploration of the combinatorial space (see \hyperref[app:technique_code_mapping]{Appendix~\ref*{app:technique_code_mapping}} for an example of iteration 2 mapping).

\subsubsection{Overall \texttt{Breakers}' Performance}

Across all 1,440 \breakers' configurations in each iteration, average detector accuracy decreased from 71.1\% in iteration 3 to 60.2\% in iteration 4 and 57.6\% in iteration 5. Consequently, the average LFR increased from 28.9\% to 39.8\% and then to 42.4\%, respectively. This indicates that the \breakers' techniques became more effective over time, both in terms of average attack effectiveness to bypass detection and the broader evolution of attack combinations. However, it is observed that the improvement from iteration 4 to iteration 5 was smaller than the improvement from iteration 3 to iteration 4.

\hyperref[fig:iteration_progress_flip_rates]{Fig.~\ref*{fig:iteration_progress_flip_rates}} compares the ranked $LFR$ distributions for all $1,440$ \breakers' configurations across iterations 3 to 5. The figure supports the increase in \breakers' effectiveness over time, showing that $LFR$ values shifted upward from iteration~3 to iteration~5. 
In iteration~3, the median $LFR$ was 23.0\%, while the mean was 28.9\%, which shows the distribution contained a strong upper tail. This means a smaller subset of configurations achieved high flip rates, while many configurations remained weaker. In iteration 4, both the median and mean increased to 36.4\% and 39.8\%, respectively, indicating a broader upward shift across the distribution rather than improvements being limited to isolated high-performing combinations. This trend continued into iteration 5, where the median increased to 41.6\% and the mean to 42.4\%. The narrowing gap between the mean and median across the iterations suggests that the \breakers' techinques performance became more evenly distributed, with a larger proportion of attack combinations achieving moderate to high $LFR$. 
\hyperref[fig:iteration_progress_flip_rates]{Fig.~\ref*{fig:iteration_progress_flip_rates}} also shows that across the three iterations, most individual techniques 
remained below the overall mean and median, which suggests that individual attacks were generally less effective than combined ones. Techniques such as D2, D3 and D4 suggest that the evolved prompt-based evasion techniques became more competitive as the \breakers' process matured. However, the strongest overall $LFR$ values came from combined rather than individual techniques. \hyperref[sec:best-performing-breaker-techniques]{Section~\ref*{sec:best-performing-breaker-techniques}} and \hyperref[sec:ablation-study]{Section~\ref*{sec:ablation-study}} discusses these findings in more detail.

\begin{center}
    \includegraphics[width=0.85\linewidth]{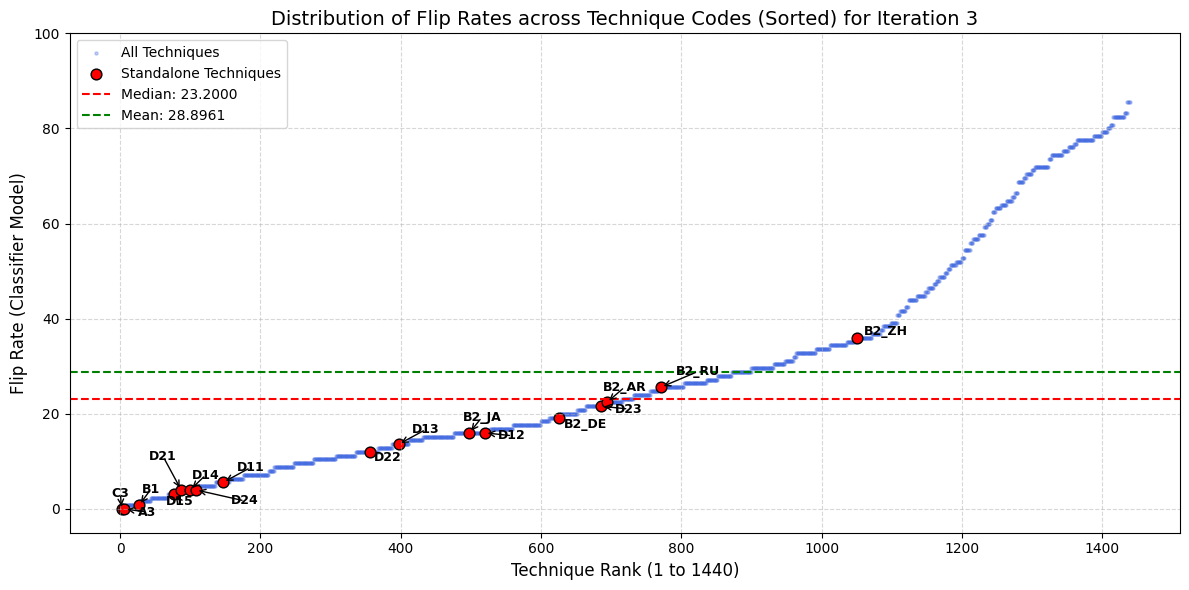}
    \includegraphics[width=0.85\linewidth]{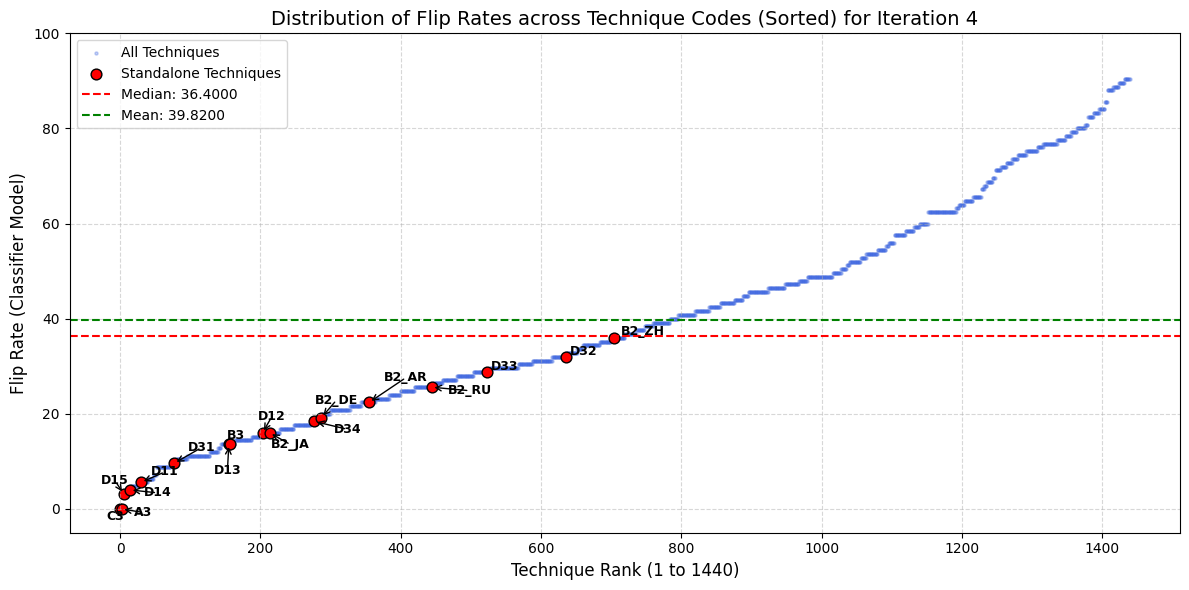}
    \includegraphics[width=0.85\linewidth]{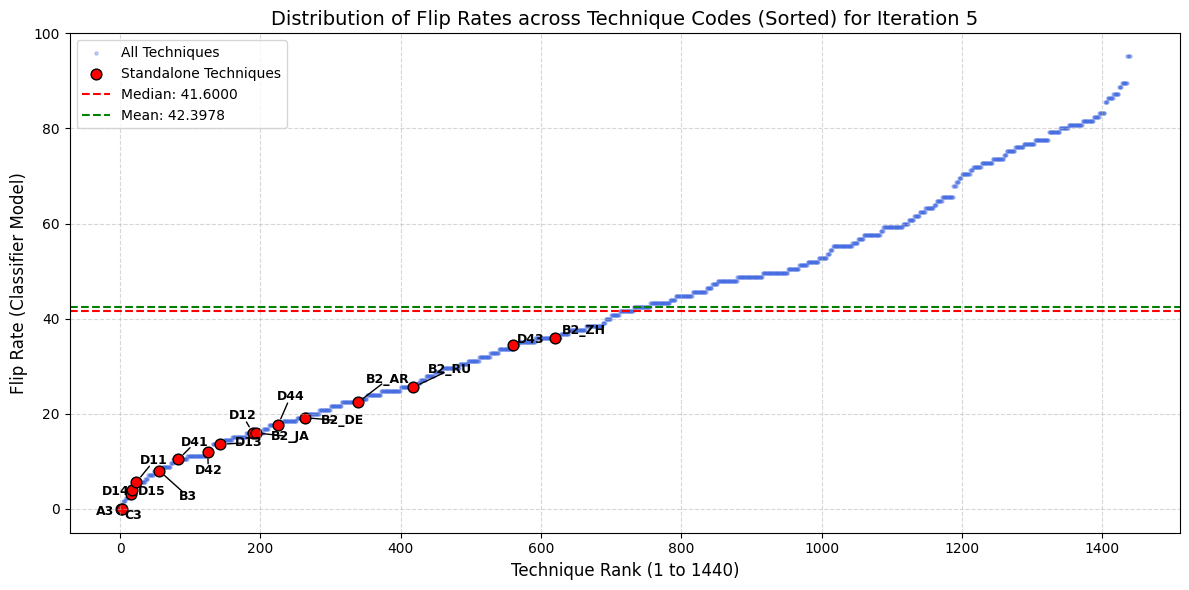}

    \cascaptionoffigure
    {Sorted $LFR$ distributions across technique codes for iterations 3 to 5 using the baseline model. The upward shift in mean and median flip rates shows that \breakers' techniques in later iterations produced stronger and more consistent adversarial effectiveness across the techniques set.}
    {fig:iteration_progress_flip_rates}
\end{center}


As \hyperref[tab:means_for_iterations]{Table~\ref*{tab:means_for_iterations}} shows, the strongest \breakers' configuration in iteration 3 reached an $LFR$ of 85.6\%, reducing detector accuracy to 14.4\%. Iteration 4's strongest configuration reached an $LFR$ of 90.4\%, and in iteration 5, the strongest configuration reached 95.2\%, leaving the detector with only 4.8\% accuracy on that attack configuration. The repeated presence of B2\_AR in the max-$LFR$ combination also suggests that Arabic back-translation was a strong technique within the highest-performing combined attacks.

\renewcommand{\arraystretch}{1.4}
\begin{center}
\cascaptionoftable
{\texttt{Breakers}' attack combinations with highest $LFR$ in iterations 3 to 5.}
{tab:means_for_iterations}
{\sffamily\small
\begin{tabular}{lllll}
\noalign{\hrule height 0.75pt}

\textbf{Iteration (i)} 
& \textbf{Mean Accuracy (\%)} 
& \textbf{Mean LFR (\%)} 
& \textbf{Max LFR (\%)} 
& \textbf{Max LFR Attack Configuration} \\
\hline

3 
& 71.1 
& 28.9 
& 85.6 
& B1\_D15\_D23\_B2\_AR \\

4 
& 60.2 
& 39.8 
& 90.4 
& B3\_D13\_D33\_B2\_AR \\

5 
& 57.6 
& 42.4 
& 95.2 
& B3\_D13\_D43\_B2\_AR \\

\noalign{\hrule height 0.75pt}
\end{tabular}}
\end{center}
\renewcommand{\arraystretch}{1}

\subsubsection{Best Performing \texttt{Breakers}' Techniques}
\label{sec:best-performing-breaker-techniques}

This section focuses on the attack effectiveness (measured in terms of $LFR$), while the semantic validity is assessed in \hyperref[sec:semantic-preservation-and-claim-retention]{Section~\ref*{sec:semantic-preservation-and-claim-retention}}.
Across iterations 3 to 5, the strongest individual back-translations technique was B2\_ZH, achieving an $LFR$ of 36\%. However, as shown in \hyperref[sec:ablation-study]{Section~\ref*{sec:ablation-study}}, B2\_AR performed better when used in combination with other \breakers' attack techniques. This distinction indicates that standalone and combined performances may differ substantially.

\begin{center}
    \includegraphics[scale=0.5]{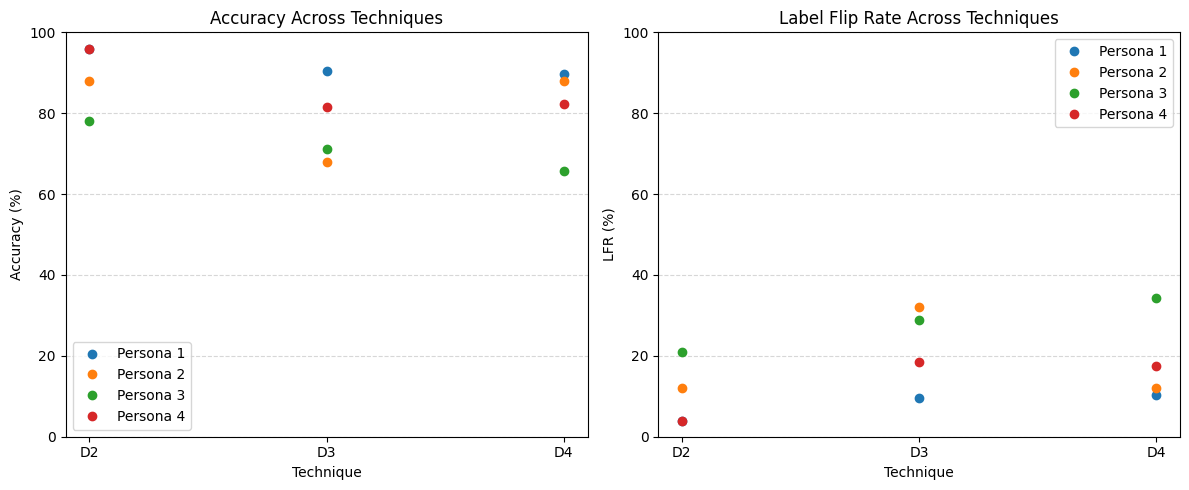}

    \cascaptionoffigure
    {Comparison of standalone persona-based techniques across D2, D3 and D4 which shows how evolved persona prompts increased $LFR$ while reducing detector accuracy.}
    {fig:d2_d3_d4_lfr_accuracy}
\end{center}

The D2, D3 and D4 technique families represent successive evolutions of persona-based \breakers' strategy. As shown in \hyperref[fig:d2_d3_d4_lfr_accuracy]{Fig.~\ref*{fig:d2_d3_d4_lfr_accuracy}}, the transition from D2 to D3 produced the highest increase in detector evasion as the strongest D2 variant achieved an $LFR$ of 21.6\% whilst the strongest D3 variant reached 32.0\%. This improvement is linked to the change in prompt structure shown in \hyperref[tab:technique_family]{Table~\ref*{tab:technique_family}}. The D2 variants use a simpler persona prompt, e.g.: ``You are a [PERSONA]'', whereas D3 expands the persona with more demographic details such as age, ethnicity, gender, location and occupation. An illustrative example of this is the \textit{college student} persona, where D22 was the simple version and D32 used ``a 20-year old Asian female college student'' profile. D32 achieved 32\% $LFR$, outperforming the strongest D2 variant and suggesting that the added information made the persona-based rewrite more effective. A similar pattern is visible in the later D4 variants, which uses the same structure as D3 but uses an indirect persona, e.g.: ``Think like a [DEVELOPED PERSONA]''. For example, the \textit{brain-rot 10-year old persona} achieved 21.6\% $LFR$ under the simpler D23 format, while the evolved D43 variant achieved 34.4\% $LFR$.

\paragraph{\textbf{Combined \breakers' configurations.}} The strongest combined \breakers' configurations across iterations 3 to 5 are shown in  \hyperref[tab:means_for_iterations]{Table~\ref*{tab:means_for_iterations}}. These were analysed by calculating $LFR$ for each atomic technique code, then grouping the atomic codes into their families. This avoids treating each atomic technique as a separate value and enables the comparison of the technique family combinations more fairly. 

 \subparagraph{\textit{Iteration 3}.} \hyperref[tab:i3_tech_code_grp_lfr]{Table~\ref*{tab:i3_tech_code_grp_lfr}} shows iteration 3's results based on the technique families. The strongest grouped configuration type was D2\_B2 with a mean $LFR$ of 38.48\% and a median of 33.6\%. This combination performed better than the individual techniques: D2 and B2, with mean $LFRs$ of 10.4\% and 23.4\%, respectively. However, the full B1\_D1\_D2\_B2 chain did not have the strongest mean performance, even though it produced the highest individual $LFR$ at 85.6\% (reduced detector accuracy to 14.4\%). This suggests that adding B1 and D1 could create strong individual attacks, but could not consistently improve the average performance across all configurations. 
\renewcommand{\arraystretch}{1.4}
\begin{center}
\cascaptionoftable
{Iteration 3 technique family grouping based mean, median and max $LFR$ percentage.}
{tab:i3_tech_code_grp_lfr}
{\sffamily\small
\begin{tabular}{@{}l l l l l@{}}
\noalign{\hrule height 0.75pt}
\textbf{Combination Type} 
& \textbf{$n$ Configs} 
& \textbf{Mean LFR (\%)} 
& \textbf{Median LFR (\%)} 
& \textbf{Max LFR (\%)} \\ 
\hline

D2\_B2         & 80  & 38.48 & 33.6 & 82.4 \\
D1\_D2\_B2     & 400 & 36.55 & 29.6 & 83.2 \\
B1\_D2\_B2     & 80  & 34.09 & 26.8 & 82.4 \\
B1\_D1\_D2\_B2 & 400 & 33.46 & 25.6 & 85.6 \\
D1\_B2         & 100 & 31.01 & 29.6 & 62.4 \\
B2             & 20  & 23.84 & 22.4 & 36   \\
D2             & 16  & 10.4  & 8    & 21.6 \\
B1             & 4   & 0.8   & 0.8  & 0.8  \\

\noalign{\hrule height 0.75pt}
\end{tabular}}
\end{center}
\renewcommand{\arraystretch}{1}

\subparagraph{\textit{Iteration 4}.} The iteration 4 attack techniques showed more effectiveness, as can be seen in  \hyperref[tab:i4_tech_code_grp_lfr]{Table~\ref*{tab:i4_tech_code_grp_lfr}}. The strongest grouped configuration was D3\_B2 with a mean $LFR$ of 52.72\% and a median $LFR$ of 48.4\%. This combination was more effective than the individual counterparts D3 (with a mean $LFR$ of 22.2\%) and B2 (which remained at 23.84\% $LFR$). The larger combinations involving B3 and D1 were also strong, although they did not outperform the mean $LFR$ of D3\_B2. For example, B3\_D1\_D3\_B2 achieved the highest maximum $LFR$ of 90.4\%, but its mean $LFR$ was 48.53\%. This suggests that B3 and D1 help create high-performing configurations, but they are not the main drivers of consistent detector evasion.

\renewcommand{\arraystretch}{1.4}
\begin{center}
\cascaptionoftable
{Iteration 4 technique family grouping based mean, median and max $LFR$ percentage.}
{tab:i4_tech_code_grp_lfr}
{\sffamily\small
\begin{tabular}{@{}l l l l l@{}}
\noalign{\hrule height 0.75pt}
\textbf{Combination Type} 
& \textbf{$n$ Configs} 
& \textbf{Mean LFR (\%)} 
& \textbf{Median LFR (\%)} 
& \textbf{Max LFR (\%)} \\ 
\hline

D3\_B2         & 80  & 52.72 & 48.4 & 89.6 \\
B3\_D3\_B2     & 80  & 50.72 & 45.6 & 88.8 \\
B3\_D1\_D3\_B2 & 400 & 48.53 & 47.6 & 90.4 \\
D1\_D3\_B2     & 400 & 47.39 & 45.6 & 90.4 \\
D1\_B2         & 100 & 31.01 & 29.6 & 62.4 \\
B3\_D1\_B2     & 100 & 30.72 & 31.2 & 57.6 \\
B3\_B2         & 20  & 24.96 & 24   & 32.8 \\
B2             & 20  & 23.84 & 22.4 & 36   \\
D3             & 16  & 22.2  & 23.6 & 32   \\
B3             & 4   & 13.6  & 13.6 & 13.6 \\

\noalign{\hrule height 0.75pt}
\end{tabular}}
\end{center}
\renewcommand{\arraystretch}{1}

\subparagraph{\textit{Iteration 5}} produced the strongest overall combined attack as seen in  \hyperref[tab:i5_tech_code_grp_lfr]{Table~\ref*{tab:i5_tech_code_grp_lfr}}. The highest grouped mean $LFR$ came from B3\_D4\_B2, with a mean $LFR$ of 58.44\%, followed by D4\_B2 at 56.88\%. Both of the strongest configurations contained D4 and B2, suggesting that this was one of the best pairings of techniques to evade detector classification of MGO. The grouped family technique codes containing D4 and B2, such as B3\_D1\_D4\_B2, produced the highest maximum $LFR$, but had means below those for B3\_D4\_B2 and D4\_B2, respectively.  

\renewcommand{\arraystretch}{1.4}
\begin{center}
\cascaptionoftable
{Iteration 5 technique family grouping based on mean, median and max LFR percentage.}
{tab:i5_tech_code_grp_lfr}
{\sffamily\small
\begin{tabular}{@{}l l l l l@{}}
\noalign{\hrule height 0.75pt}
\textbf{Combination Type} 
& \textbf{$n$ Configs} 
& \textbf{Mean LFR (\%)} 
& \textbf{Median LFR (\%)} 
& \textbf{Max LFR (\%)} \\ 
\hline

B3\_D4\_B2     & 80  & 58.44 & 56.8 & 87.2 \\
D4\_B2         & 80  & 56.88 & 56   & 89.6 \\
D1\_D4\_B2     & 400 & 51.95 & 49.6 & 89.6 \\
B3\_D1\_D4\_B2 & 400 & 51.4  & 48.4 & 95.2 \\
D1\_B2         & 100 & 31.01 & 29.6 & 62.4 \\
B3\_D1\_B2     & 100 & 27.9  & 24.8 & 59.2 \\
B2             & 20  & 23.84 & 22.4 & 36   \\
B3\_D4         & 16  & 21.2  & 21.2 & 28   \\
D4             & 16  & 18.6  & 14.8 & 34.4 \\
B3             & 4   & 8     & 8    & 8    \\

\noalign{\hrule height 0.75pt}
\end{tabular}}
\end{center}
\renewcommand{\arraystretch}{1}

\subsubsection{Ablation Study}
\label{sec:ablation-study}

\begin{figure}
  \centering

  \begin{subfigure}{0.32\textwidth}
    \centering
    \includegraphics[width=\linewidth]{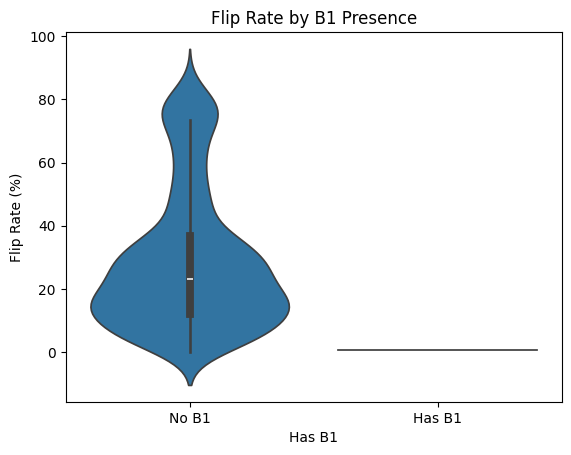}
    \caption{Technique Family B1}
    \label{fig:b1_presence}
  \end{subfigure}\hfill
  \begin{subfigure}{0.32\textwidth}
    \centering
    \includegraphics[width=\linewidth]{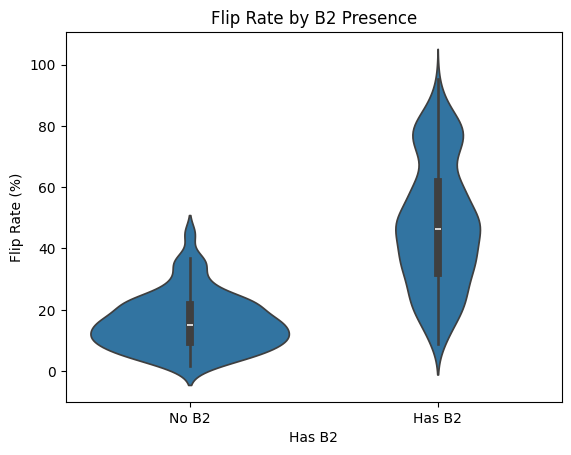}
    \caption{Technique Family B2}
    \label{fig:b2_presence}
  \end{subfigure}\hfill
  \begin{subfigure}{0.32\textwidth}
    \centering
    \includegraphics[width=\linewidth]{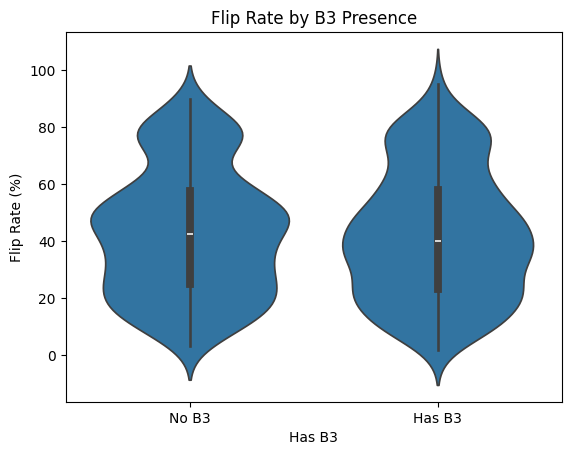}
    \caption{Technique Family B3}
    \label{fig:b3_presence}
  \end{subfigure}

  \vspace{1em}

  \begin{subfigure}{0.32\textwidth}
    \centering
    \includegraphics[width=\linewidth]{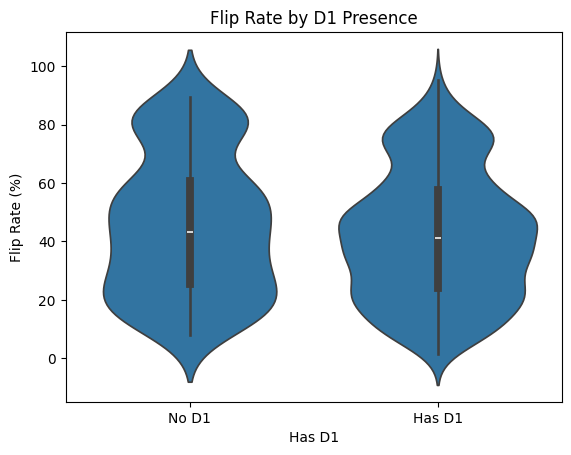}
    \caption{Technique Family D1}
    \label{fig:d1_presence}
  \end{subfigure}\hfill
  \begin{subfigure}{0.32\textwidth}
    \centering
    \includegraphics[width=\linewidth]{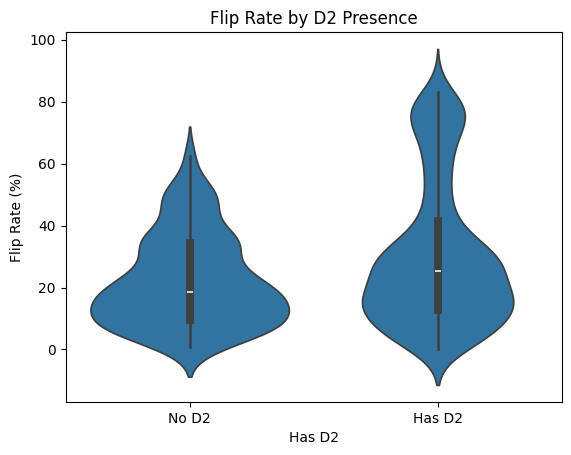}
    \caption{Technique Family D2}
    \label{fig:d2_presence}
  \end{subfigure}\hfill
  \begin{subfigure}{0.32\textwidth}
    \centering
    \includegraphics[width=\linewidth]{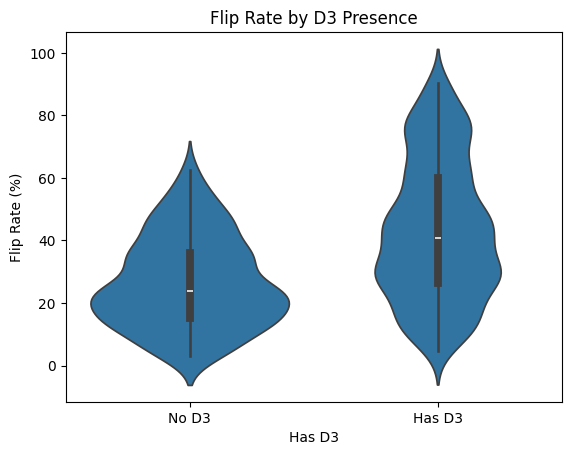}
    \caption{Technique Family D3}
    \label{fig:d3_presence}
  \end{subfigure}

  \vspace{1em}

  \begin{subfigure}{0.32\textwidth}
    \centering
    \includegraphics[width=\linewidth]{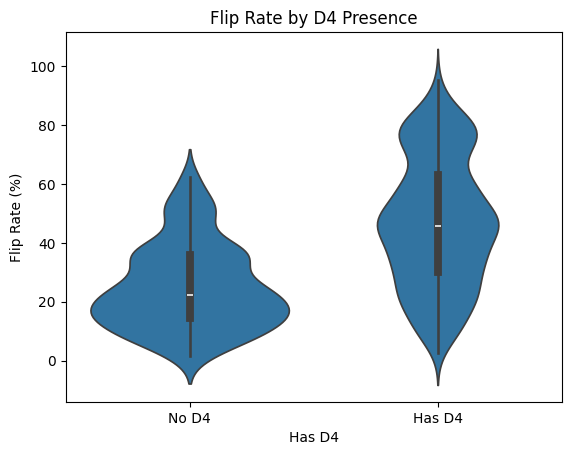}
    \caption{Technique Family D4}
    \label{fig:d4_presence}
  \end{subfigure}\hfill
  \begin{subfigure}{0.32\textwidth}
    \centering
    \includegraphics[width=\linewidth]{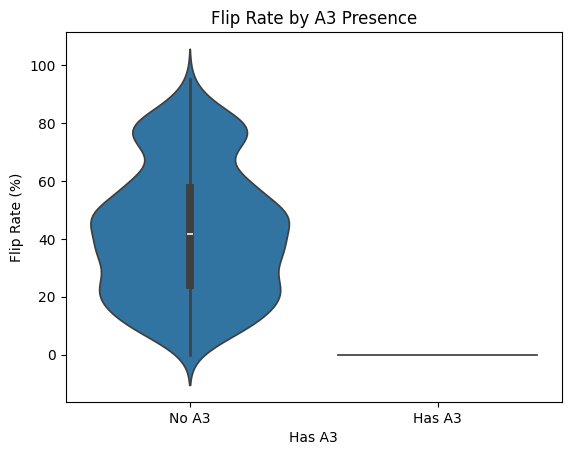}
    \caption{Technique Family A3}
    \label{fig:a3_presence}
  \end{subfigure}\hfill
  \begin{subfigure}{0.32\textwidth}
    \centering
    \includegraphics[width=\linewidth]{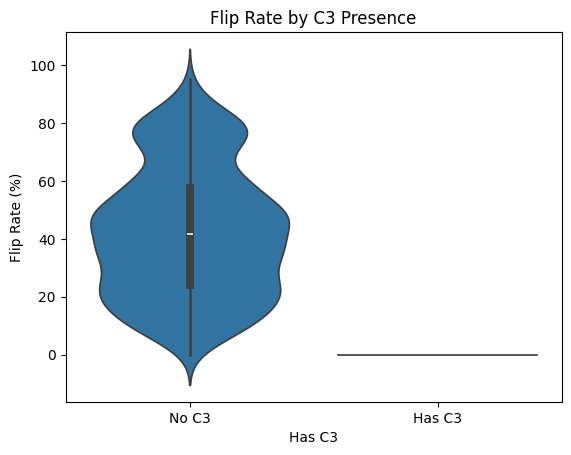}
    \caption{Technique Family C3}
    \label{fig:c3_presence}
  \end{subfigure}

  \caption{Flip-rate distributions comparing configurations where each technique family is present against configurations where it is absent.}
  \label{fig:ablation_violin}
\end{figure}

The ablation study compares the distribution of $LFR$ when each technique family is absent versus present in a \breakers' configuration. This helps identify whether a technique family consistently shifts detector behaviour or if it only assists in edge cases. \hyperref[fig:ablation_violin]{Fig.~\ref*{fig:ablation_violin}} shows this comparison for all technique families.

\paragraph{B2 Analysis.} \hyperref[fig:ablation_violin]{Fig.~\ref*{fig:b2_presence}} ablation results show that B2 (back-translation) was the most reliable contributor to detector evasion.
When B2 is absent from the violin graphs, the $LFR$ distribution is concentrated at lower values, with most configurations remaining below 30\% $LFR$. When it is present, the distribution shifts upwards, with a wider spread showing many configurations reaching between 50--90\% $LFR$. This indicates that B2 is the strongest and most reliable contributor to detector evasion across the \breakers' techniques.

\paragraph{B1 and B3 Analysis.} B1 (prompt-based paraphrasing) shows no useful contributions to $LFR$, with distribution around 1\% LFR, as seen in \hyperref[fig:b1_presence]{Fig.~\ref*{fig:b1_presence}}. This shows that the original B1 paraphrasing method was not effective as an independent attack. However, B3 (a revised version of B1 to enforce lexical and syntactic variation) behaves differently (\hyperref[fig:b3_presence]{Fig.~\ref*{fig:b3_presence}}). The B3's  present or absence distributions are closer together, containing a broad spread of flip rates. This suggests that B3 is not as dominant as B2, but it does not collapse to near-zero performance like B1. B3's presence alone does not contribute to $LFR$, but it features among high performing configurations. 


\paragraph{D1 Analysis.} Similar to B3, the D1 (tone-based LLM prompt rewriting) violin plot shows a mixed effect, with a large overlap between configurations where D1 is present or absent (\hyperref[fig:d1_presence]{Fig.~\ref*{fig:d1_presence}}). The D1-present distribution doesn't show a clear upward shift and its overall distribution appears similar/slightly lower than the D1-absent group. This suggests that tone-based prompts transformations alone are not effective at evading detection. D1 helps to improve certain technique combinations, especially when applied together with back-translation and a strong persona variant, but the ablation distribution does not show D1 to be a primary contributor to increase $LFR$.

\paragraph{D2 Analysis.} \hyperref[fig:d2_presence]{Fig.~\ref*{fig:d2_presence}} shows D2 (persona-based rewriting) produces some upward shift when present but the effect is less stable than that of B2. The D2-present distribution includes higher flip-rate configurations and it also contains more configurations that contribute to $LFR$ between the range of 5--35\%. This shows the earlier persona based prompts were useful to achieve high $LFR$, but D2 is not consistent at evading detection. 

\paragraph{D3 Analysis.} D3-present (refined persona-based rewriting) shows a stronger effect (\hyperref[fig:d3_presence]{Fig.~\ref*{fig:d3_presence}}) than D2 in terms of $LFR$, with the distribution shifting upward and more configurations reaching higher $LFR$ values. This supports aforementioned interpretation that stronger D3 standalone variants exceeded the strongest D2 standalone variant. It suggests that the demographic-specific D3 persona prompts were more effective than the simpler D2 prompts.

\paragraph{D4 Analysis.}
The presence of D4 (further refined persona-based rewriting) (\hyperref[fig:d4_presence]{Fig.~\ref*{fig:d4_presence}}) produces a noticeable upward shift in the distribution of flip rate, resulting in a substantially higher concentration of configurations with $LFR$ values exceeding 60\%.
This suggests that, in line with previous findings, D4 is the most effective persona-based technique family. However, the distribution is more spread, which suggests that it was not uniformly successful across all variants. 

\paragraph{A3 and C3 Analysis.} Results shows that A3 (rule-based whitespace editing) and C3 (emotion-based emoji injection) did affect the $LFR$ values. Across iterations 3 to 5, the presence of A3 or C3 or C3\_A3 did not change the grouped average mean nor maximum $LFR$ listed in  \hyperref[tab:means_for_iterations]{Table~\ref*{tab:means_for_iterations}}. This suggests that these techniques do not contribute to detector evasion. 
\hyperref[fig:a3_presence]{Fig.~\ref*{fig:a3_presence}} and \hyperref[fig:c3_presence]{Fig.~\ref*{fig:c3_presence}} supports this interpretation, since A3 and C3-present distributions are concentrated near 0\% $LFR$ and do not show the upward distributional $LFR$ shift as observed for B2, D3 and D4.

Overall the ablation study shows that B2 was the strongest contributor to high $LFR$, with D3 and D4 being the strongest prompt-based contributors. D2 showed a weaker effect, but was useful in earlier iterations even though it had less consistency across distributions. By contrast, B1 and D1 did not show any reliable effects and B3 appeared to support other techniques rather than contributing directly as a standalone one. The rule-based perturbation techniques of C3 and A3 had no positive ablation effect, with their $LFR$ distribution near 0\%. However, these results only measure detector evasion, so semantic preservation analysis is needed to determine whether the strongest \breakers' techniques remained valid adversarial transformations. The following section presents the results of this investigation.

\subsection{Semantic Preservation and Claim Retention (RQ2)}
\label{sec:semantic-preservation-and-claim-retention}

\paragraph{Manual evaluation} was conducted using the metrics shown in \hyperref[tab:manual_eval]{Table~\ref*{tab:manual_eval}} and the annotation guidelines in \hyperref[app:ME_guidelines]{Appendix~\ref*{app:ME_guidelines}}. 
The results shows that the quality of the adversarial attack outputs generally improved from iteration 3 to 5. \hyperref[tab:manual_eval_result_avgs]{Table~\ref*{tab:manual_eval_result_avgs}} illustrates that iteration 3 produced valid adversarial examples, however, the manual evaluation indicates these examples were less stable. Iteration 4 improves the scores across all criteria, particularly ME2 and ME3. Finally, iteration 5 achieved the highest semantic preservation score overall, although the improvements over iteration 4 was smaller. 
Between iteration~3 and 4, ME2 increased from 3.699 to 3.923, while ME3 increased from 0.797 to 0.866. This shows that, overall, the naturalness and fluency of the posts were maintained. Iteration 5 further improved ME1 from 2.517 to 2.585, indicating stronger semantic preservation, while ME2 and ME3 remained almost unchanged from iteration 4. 

\renewcommand{\arraystretch}{1.4}
\begin{center}
\cascaptionoftable
{The technique family codes manually analysed per iteration and its average scores per category across the iterations.}
{tab:manual_eval_result_avgs}
{\sffamily\small
\begin{tabular}{llllll}
\noalign{\hrule height 0.75pt}
\textbf{Iteration (i)} 
& \textbf{Technique Family Analysed} 
& \textbf{$n$-Posts} 
& \textbf{ME1 Average} 
& \textbf{ME2 Average} 
& \textbf{ME3 Average} \\
\hline

3 
& D2, B2, D2\_B2 
& 290 
& 2.456 
& 3.699 
& 0.797 \\

4 
& D3, B3, D3\_B2 
& 250 
& 2.517 
& 3.923 
& 0.866 \\

5 
& D4, D4\_B2 
& 240 
& 2.585 
& 3.912 
& 0.868 \\

\noalign{\hrule height 0.75pt}
\end{tabular}}
\end{center}
\renewcommand{\arraystretch}{1}

\paragraph{Technique Level Analysis.} 
Across the B2 language variants, the most consistent pattern was B2\_ZH (Chinese back-translation), demonstrating the weakest semantic preservation and claim-retention results. The heat map in  \hyperref[fig:b2_ME1_ME3]{Fig.~\ref*{fig:b2_ME1_ME3}} also confirms that the B2\_ZH  introduced the biggest meaning variation. Inversely, B2\_DE (German back-translation) appeared to introduce enough linguistic variation to modify the text whilst preserving the original meaning and claim more reliably. B2\_JA and B2\_RU variants were generally stronger than B2\_ZH, while B2\_AR shows mixed results depending on the iteration and the technique combination used. This shows that back-translation is not a consistent technique and the effects depend heavily on the intermediate language used.

\begin{center}
{
\begin{minipage}{0.48\linewidth}
    \centering
    \includegraphics[width=\linewidth]{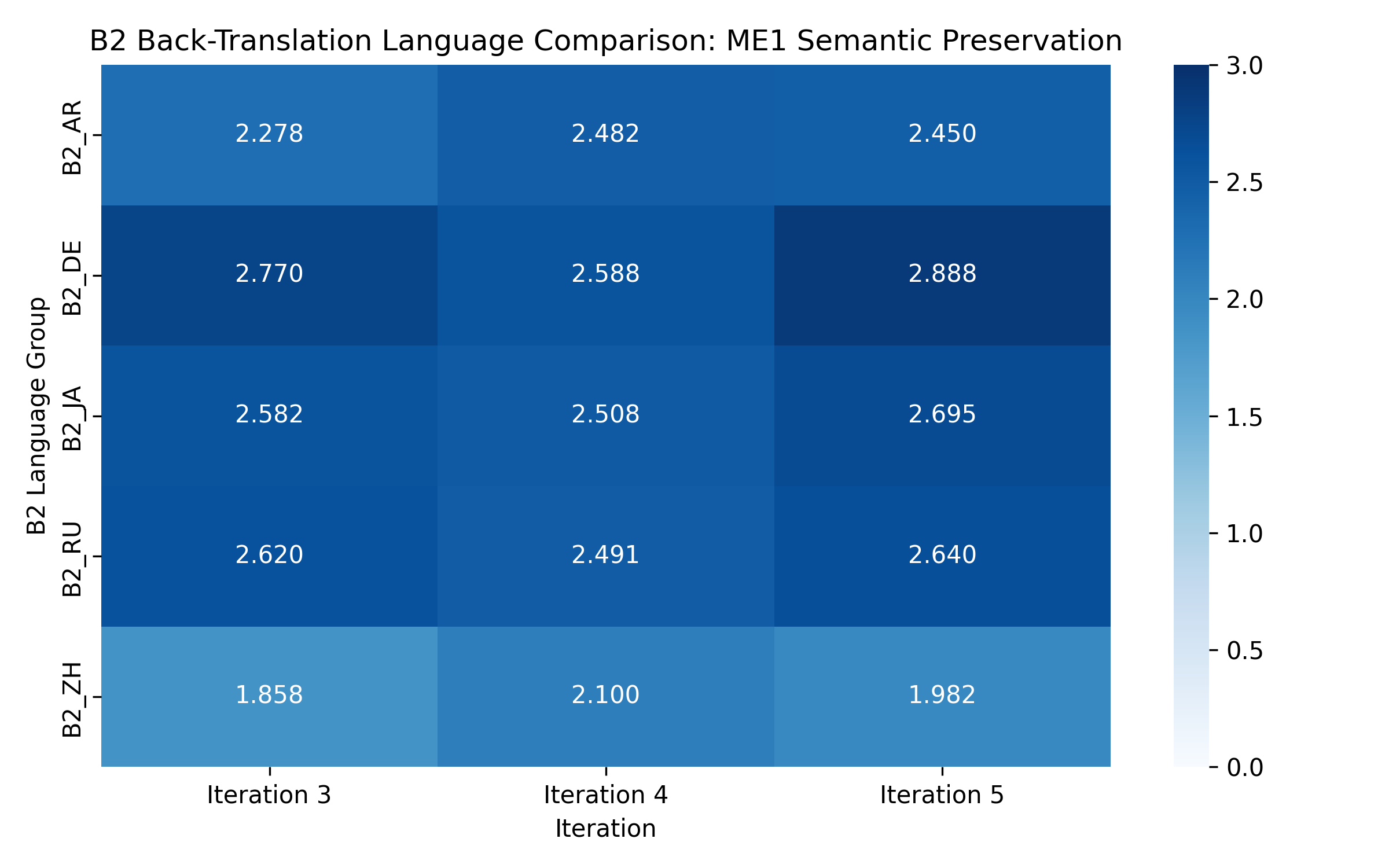}
    
    \vspace{2pt}
    {\sffamily\small\textbf{(a)} ME1 scores across Iterations 3--5 for B2.}
\end{minipage}
\hfill
\begin{minipage}{0.48\linewidth}
    \centering
    \includegraphics[width=\linewidth]{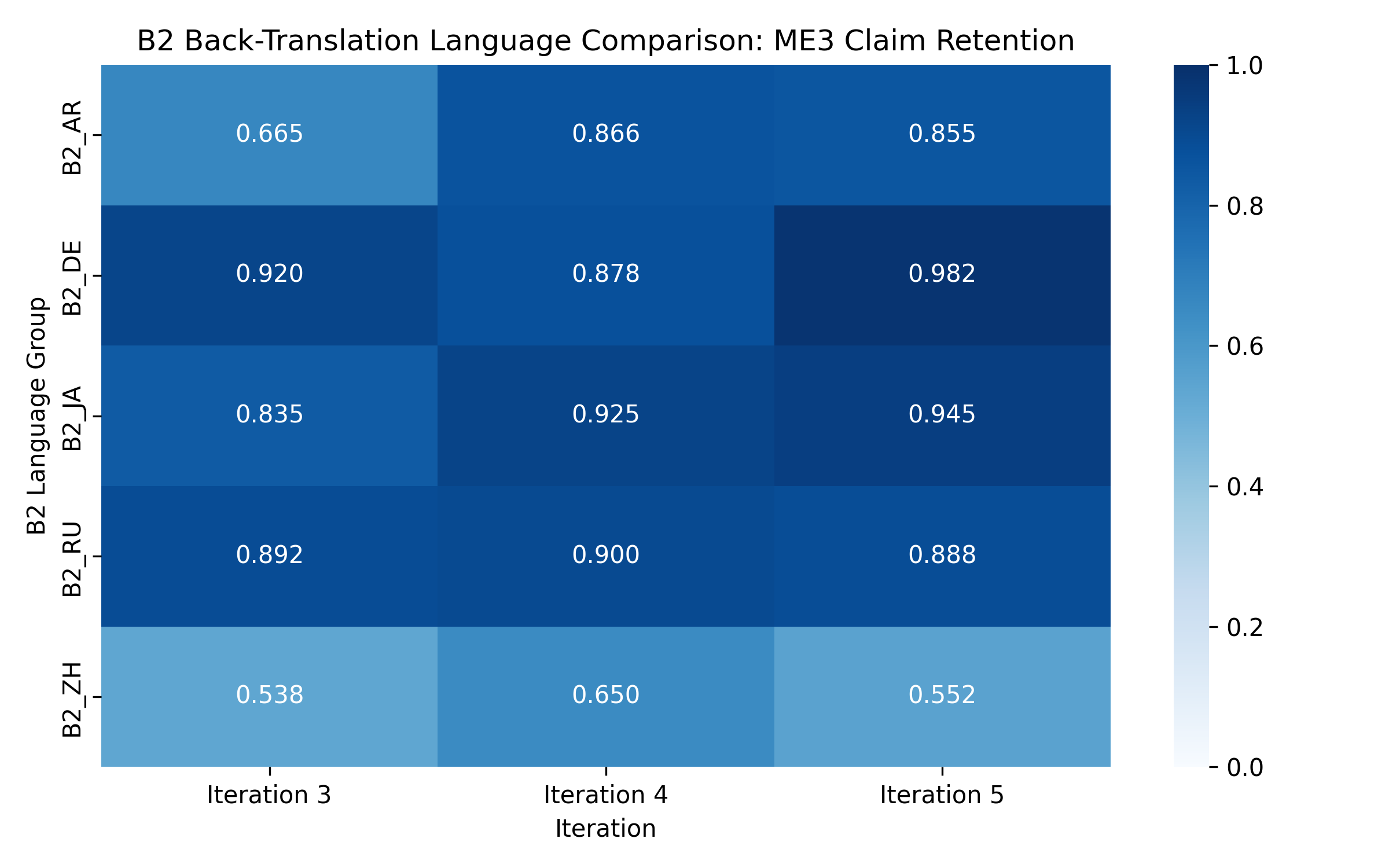}
    
    \vspace{2pt}
    {\sffamily\small\textbf{(b)} ME3 scores across Iterations 3--5 for B2.}
\end{minipage}
}
\cascaptionoffigure
{ME1 and ME3 scores across iterations 3 to 5 for the B2 technique family.}
{fig:b2_ME1_ME3}
\end{center}

\noindent{Techniques D1 to D4 shower stronger manual evaluation scores than B2 across iteration 3 to 5. In iteration 3, D2 achieved an average ME1 score of 2.672 and ME3 score of 0.966, while D2\_B2 fell to to 2.369 and 0.749 respectively. This indicates that the standalone D2 variants preserved the original meaning and disinformation claim, but adding B2 to it introduced a noticeable loss in semantic preservation. In iteration 4, D3 achieved high scores across all manual evaluation criteria's with ME1 at 2.838, ME2 at 4.000 and ME3 at 0.962. Similar scores can be seen for B3 which attains ME1 of 2.900, ME2 of 4.000 and ME3 of 0.938. By contrast, D3\_B2 reduced ME1 to 2.434 and ME3 to 0.844, showing that the addition of back-translation increased drift in semantic preservation. Iteration 5 produced the strongest results for ME1 and ME3 with D4 achieving ME1 of 2.855, ME2 of 3.990 and ME3 of 0.988, making it the best standalone technique family for preserving the original claim whilst maintaining naturalness. Within the D4 variants, D42 (indirect persona prompt, with the \textit{college student} persona) achieved the highest scores: ME1 of 2.980 and ME3 of 1. However, D4\_B2 showed reduced semantic preservation and claim retention, falling to 2.531 for ME1 and 0.874 for ME3. \hyperref[tab:b2_manual_eval]{Table~\ref*{tab:b2_manual_eval}} shows the differences between B2 as a combined versus standalone technique. The scores presented are normalised to make fair comparison between the different criteria. }

\renewcommand{\arraystretch}{1.4}
\begin{center}
\cascaptionoftable
{Comparing the effect of B2 added to standalone techniques on the manual evaluation scoring. The scoring used in this table has been normalised on a scale of 0--1 for better comparison, where ME1 is divided by 3, ME2 by 4, and ME3 by 1.}
{tab:b2_manual_eval}
{\sffamily\small
\begin{tabular}{@{}lccc@{}}
\noalign{\hrule height 0.75pt}
\textbf{Comparison} 
& \textbf{$\Delta$ ME1} 
& \textbf{$\Delta$ ME2} 
& \textbf{$\Delta$ ME3} \\ 
\hline

D2\_B2 vs D2 & -0.101 & -0.040 & -0.217 \\
D3\_B2 vs D3 & -0.134 & -0.024 & -0.118 \\
D4\_B2 vs D4 & -0.108 & -0.023 & -0.114 \\ 

\noalign{\hrule height 0.75pt}
\end{tabular}}
\end{center}
\renewcommand{\arraystretch}{1}


\noindent{Of the 780 posts that were manually analysed, $n=69$ (\textasciitilde8\%) received an ME3 score of 0, meaning that these posts did not maintain their disinformation claim. 58\% of these 69 posts flipped their label, but this was not a successful evasion since the original claim was lost. This highlights the need to investigate whether poorer semantic preservation and claim retention contribute to label flips.}

\paragraph{Classification \& Semantic Degradation Analysis.} For manual evaluation, we used a controlled sample of posts, where each technique had a 50\%-50\% split between HGO and MGO of the classified posts selected. An analysis of the equal split was conducted to see if lower manual evaluation scores within the selected subset were associated with higher HGO classification or higher HGO probability. 
Manual evaluation revealed that HGO classified posts had a lower average semantic preservation and claim-retention scores than MGO-classified posts (\hyperref[tab:semantic_degradation]{Table~\ref*{tab:semantic_degradation}}).
This pattern suggests there may be a modest association between posts classified as HGO and a weaker semantic preservation/claim retention. Naturalness was almost identical for the HGO-classified and MGO-classified posts, which suggests that the detector was not 
biased by the fluency of the input text. However, as the manually evaluated sample selection was balanced by design, with an equal number of HGO-classified and MGO-classified examples for each variant, the average difference between the HGO-classified and MGO-classified posts was relatively small. This means that the results should not be interpreted as proof that semantic degradation causes HGO classification. Instead, they suggest that some HGO classifications may be associated with weaker semantic preservation or claim retention.

\renewcommand{\arraystretch}{1.4}
\begin{center}
\cascaptionoftable
{Semantic degradation across the averaged manual evaluation metrics for MGO-classified and HGO-classified posts.}
{tab:semantic_degradation}
{\sffamily\small
\begin{tabular}{@{}lccc@{}}
\noalign{\hrule height 0.75pt}

\textbf{Metric} 
& \textbf{MGO-Classified} 
& \textbf{HGO-Classified} 
& \textbf{Difference} \\ 
\hline

ME1 & 2.576 & 2.454 & -0.122 \\
ME2 & 3.840 & 3.832 & -0.008 \\
ME3 & 0.866 & 0.816 & -0.050 \\ 

\noalign{\hrule height 0.75pt}
\end{tabular}}
\end{center}
\renewcommand{\arraystretch}{1}

\paragraph{Semantic Preservation Pipeline.} Automatic evaluation metrics were assessed in terms of their correlation with the manually evaluated posts scores. Spearman's  $\rho$ rank correlation coefficient was employed given the ordinal nature of the manual evaluation scores. 
\hyperref[fig:filtering_metrics_spearman_heatmap]{Fig.~\ref*{fig:filtering_metrics_spearman_heatmap}} shows that the semantic preservation metrics more strongly align with ME1 and ME3 than ME2. 
This is expected, as the pipeline's primary objective was to ensure semantic similarity and disinformation claim retention, rather than optimising for fluency and naturalness.
E5-cosine similarity showed the strongest relationship with the manual evaluation score, with a Spearman's $\rho$ of 0.489 for ME1 and 0.453 for ME3. BLEURT also showed a moderate positive relationship with ME1 and ME3, scoring $\rho = $0.369 and $\rho = $0.357, respectively. The token-ratio metric showed a similar moderate relationship with $\rho = $0.347 for ME1 and $\rho = $0.356 for ME3. The strongest pipeline metric was E5-based cosine similarity, suggesting that embedding-based semantic similarity was the most useful substitute for human semantic judgements. Posts with a higher E5-cosine similarity were likely to receive stronger ME1 and ME3 scores. However, the relationship was moderate rather than strong, which means E5-cosine was useful as a initial signal of semantic preservation, however, human evaluation is still needed to confirm the results. 

\begin{center}
    \includegraphics[width=0.5\linewidth]{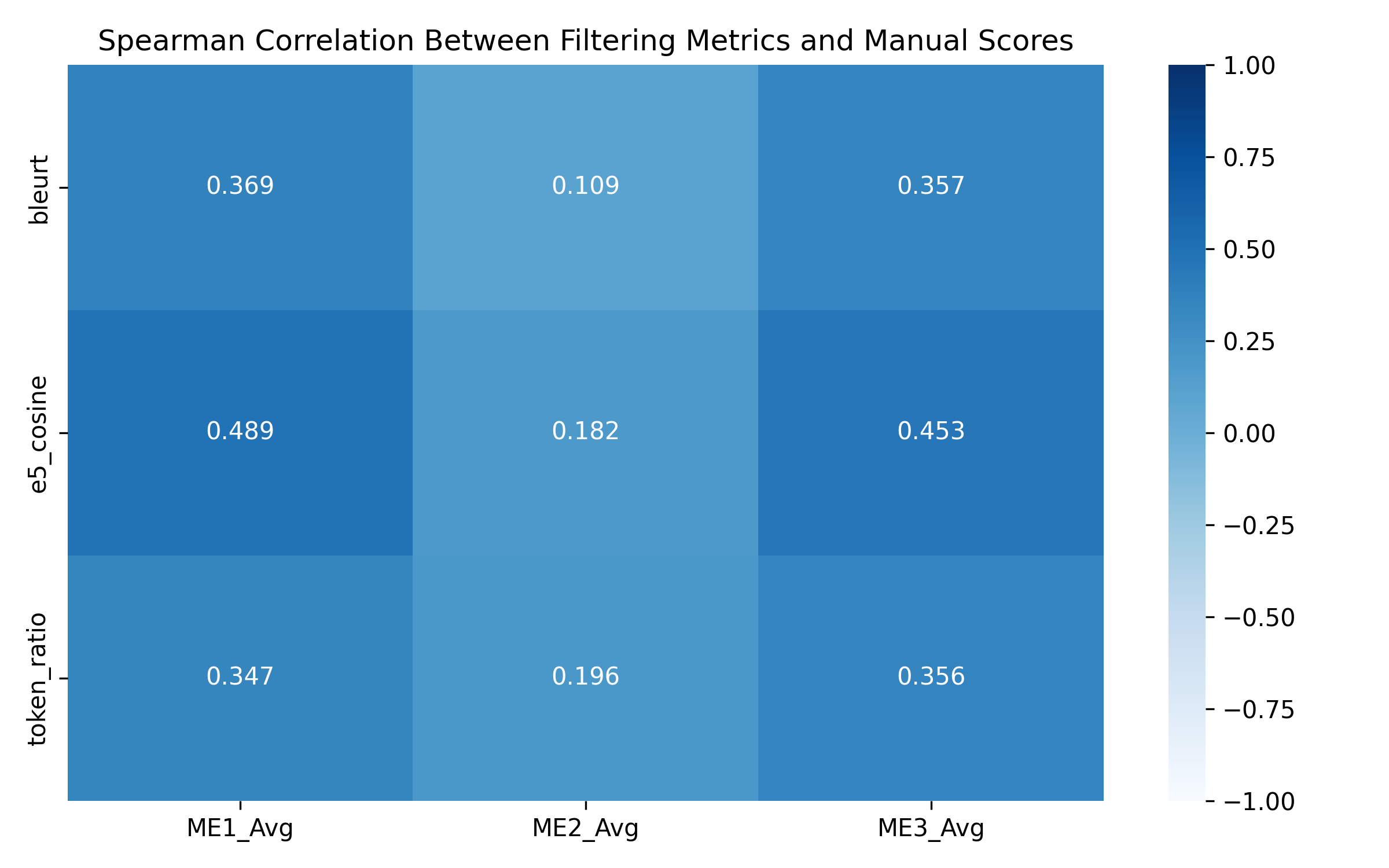}
    \cascaptionoffigure
    {Spearman correlation heatmap between BLEURT, E5-cosine similarity, and token ratio against ME1, ME2, and ME3. Higher positive values indicate a stronger correlation between the semantic preservation metrics and the manual evaluation scores.}
    {fig:filtering_metrics_spearman_heatmap}
\end{center}

\noindent{NLI labels were evaluated separately due to having a different scoring criterion. \hyperref[fig:me1_me3_by_NLI]{Fig.~\ref*{fig:me1_me3_by_NLI}} shows the breakdown of ME1 and ME3 against the NLI labels after normalisation. As can be seen, out of the 780 manually evaluated posts,  contradiction posts ($n=47$ $\mid$ 6\% ) had an average ME1 score of 1.986 and ME3 score of 0.620, compared to ME1 scores above 2.540 and ME3 scores above 0.85 for entailment ($n=205$ $\mid$ 26\%) and neutral ($n=528$ $\mid$ 68\%) posts. This suggests NLI can be useful for identifying posts with worse semantic preservation scores, especially when the posts didn't contain the original disinformation claim. However, since NLI entailment and neutral labels produced similar scores, it is more useful at identifying posts with lower semantic/claim preservation than separating preserved output posts from acceptable neutral paraphrased posts. Therefore, this metric should be interpreted more as a failure rate detector rather than a nuanced quality assessment.}

\begin{center}
    \includegraphics[scale=0.4]{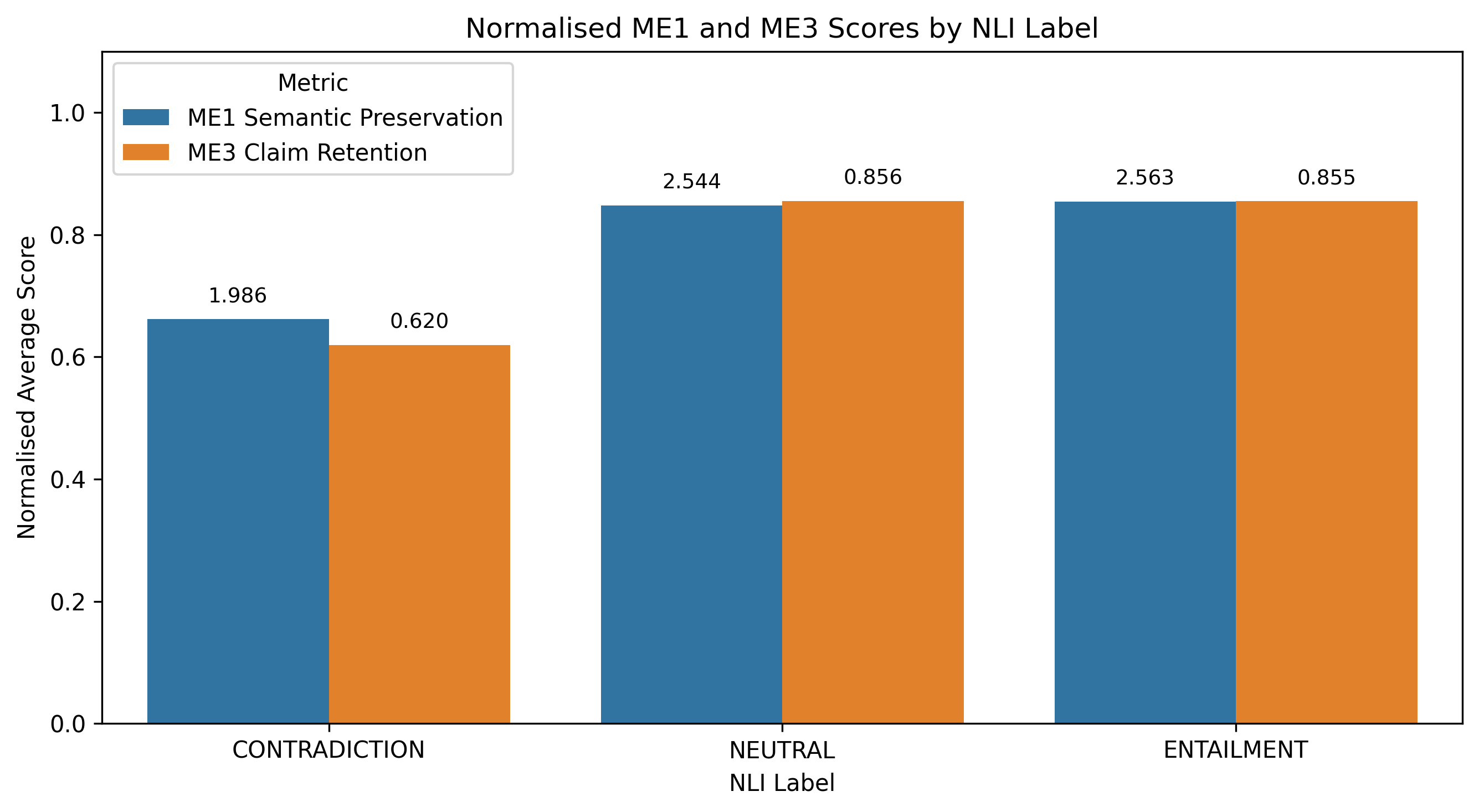}
    \cascaptionoffigure
    {Normalised ME1 and ME3 manual evaluation scores grouped by NLI label. ME1 was normalised by dividing it by 3, while ME3 already follows a 0--1 scale. The scores were normalised to allow fairer comparison.}
    {fig:me1_me3_by_NLI}
\end{center}


\noindent{Overall, the semantic preservation pipeline showed moderate agreement with human evaluation scores, with the strongest evidence of alignment coming from E5-cosine similarity and NLI labelling. However, ME2 showed only weak relationships with the filtering metrics. This means that the filtering pipeline can be useful as an initial filter, however, manual evaluation may still be necessary to ensure output posts preserve the original meaning and disinformation claim.}

\subsection{Architectures for detecting machine-generated content (RQ3)}




\hyperref[tab:builder_model_performance]{Table~\ref*{tab:builder_model_performance}} presents the results of the \builders' models evaluation. Greyed-out values represent models that were evaluated on a \breakers' set that was produced prior to the model's iteration. The evaluation is done in the full \breakers' set for each iteration, i.e. results are not divided per technique. 

\paragraph{Iteration 3.} At iteration 3 time, the most recent \builders' models were the Siamese variations (1:1 and TF-IDF). Although theoretically contrastive learning should outperform baseline-transformer models, Siamese variants underperformed significantly in comparison to the baseline model. The best Siamese model -- Siamese (TF-IDF) -- achieved 67.88\% of $F1$ score, which is over 15\% lower than the baseline model (83.11\% of $F1$). It is worth mentioning that the Baseline(BT) model slightly harmed the baseline's performance (this happens throughout all iterations). Therefore, the use of data augmentation via back-translation was not carried forward to the contrastive learning models.  

\paragraph{Iteration 4} showed the introduction of the Triplet(TF-IDF) model. This novel model, however, did not outperform the baseline and, in fact, was worse than the Siamese(TF-IDF). Iteration 4 marks a significant drop in performance from iteration 3: the baseline model had around 8\% drop in terms of $F1$ score. This highlights the \breakers' usage of more aggressive adversarial techniques as discussed in \hyperref[sec:rq1]{Section~\ref*{sec:rq1}}.

\paragraph{Iteration 5} was when the Triplet(DASS) model was introduced. Previous models (baseline and Siamese) showed a similar trend, while Triplet(DASS) outperformed the baseline by more than 4\% of $F1$ and more than 15\% in terms of accuracy. Retrospectively, Triplet(DASS) also significantly outperforms the baseline in datasets from previous iterations (greyed-out results in \hyperref[tab:builder_model_performance]{Table~\ref*{tab:builder_model_performance}}).



\renewcommand{\arraystretch}{1.4}
\begin{center}
\cascaptionoftable
{\texttt{Builders}' models performance metrics ($ACC$ and $F1$) across all iterations. Results greyed-out are for models that were developed after the given iteration. Best results per iteration are marked in bold.}
{tab:builder_model_performance}
{\sffamily\small
\setlength{\tabcolsep}{12pt} 
\begin{tabular}{l | c c | c c | c c}
\noalign{\hrule height 0.75pt}

& \multicolumn{2}{c|}{\textbf{Iteration 3}} 
& \multicolumn{2}{c|}{\textbf{Iteration 4}} 
& \multicolumn{2}{c}{\textbf{Iteration 5}} \\ 
\cline{2-7} 
& \textbf{$ACC$} & \textbf{$F1$} 
& \textbf{$ACC$} & \textbf{$F1$} 
& \textbf{$ACC$} & \textbf{$F1$} \\ 
\hline

Baseline 
& 71.10\% & 83.11\% 
& 60.18\% & 75.14\% 
& 57.60\% & 73.10\% \\ 
\hline

Baseline(BT)
& 69.74\% & 82.17\% 
& 58.61\% & 73.90\% 
& 56.92\% & 72.54\% \\ 
\hline

Siamese(1:1)
& 24.94\% & 39.92\% 
& 16.99\% & 29.05\% 
& 17.05\% & 29.13\% \\ 
\hline

Siamese(TF-IDF)
& 51.38\% & 67.88\% 
& 33.36\% & 50.03\% 
& 27.23\% & 42.80\% \\ 
\hline

Triplet(TF-IDF)
& \color{gray}{47.02\%} & \color{gray}{63.96\%} 
& 29.48\% & 45.54\% 
& 24.44\% & 39.28\% \\ 
\hline

Triplet(DASS)
& \textbf{\color{gray}{81.47\%} }& \textbf{\color{gray}{89.79\%} }
& \textbf{\color{gray}{72.88\%}} & \textbf{\color{gray}{84.31\%} }
& \textbf{72.68\% }& \textbf{84.18\%} \\ 

\noalign{\hrule height 0.75pt}
\end{tabular}
}
\end{center}
\renewcommand{\arraystretch}{1}

\paragraph{Summary.} The Triplet(DASS) contrastive architecture invariably outperformed the transformer-based baseline, whereas the other contrastive models (excluding DASS) underperformed. These results indicate that contrastive methods are most effective against adversarial paraphrase-based and persona-driven transformations employed by the \breakers~only when they explicitly learn to generalise across paraphrasing. 

\subsubsection{Interpretation of RQ2 Results}
\paragraph{The baseline classifier} fell from 71.10\% to 57.60\% accuracy from iteration 3 to iteration 5. This decline of 13.5\% aligns with the \breakers' introduction of persona-based paraphrasing (D3 in iteration 4, D4 in iteration 5) combined with back-translation (B2). As shown in \hyperref[sec:best-performing-breaker-techniques]{Section~\ref*{sec:best-performing-breaker-techniques}}, the strongest \breakers' attack configuration (B3\_D1\_D4\_B2) achieved a maximum $LFR$ of 95.2\% against the baseline classifier. The baseline had only a 4.8\% accuracy rate on that specific set. We can interpret this as evidence that the classifier relied on surface-level correlations rather than learning the underlying semantic meaning of a post. Therefore, if the same disinformation claim is rewritten in multiple surface forms (i.e. different wording, persona styles, or syntactic structures), the classifier can treat each perturbed text as if it were a different class, misclassifying it as HGO as it fails to recognise them as variants of the same machine-generated content.

\paragraph{As the Siamese (1:1) variant} was trained on fixed pairs where each human post was paired with its machine-generated twin on the same claim, it was expected to distinguish between HGO and MGO on identical topics. Yet, it was the worst-performing model, achieving a 24.94\% accuracy rate in iteration 3 and dropping to the overall worst accuracy rate of 16.99\% at iteration 4. This shows that the lack of variance in the training signal can significantly impair a contrastive network model to detect authorship. It may consider superficial artefacts that are only present in a pair, like a particular word that appears only in a specific HGO-MGO sample. The failure to generalise was exploited by the \breakers~when they introduced paraphrase variants; since the model never saw such variation, its decision boundary completely failed.

\paragraph{Both contrastive architectures (Siamese and Triplet)} explored TF-IDF pairing. These showed moderate performance at iteration 3 (51\% and 47\% accuracy scores, respectively) but were broken by \breakers~techniques in iteration 5 (27\% and 24\% accuracy scores, respectively). Since TF-IDF measures word overlap (lexical similarity) rather than semantics, once the \breakers~introduced persona-based prompts and multi-layered back-translation attacks, the lexical similarity between the original and transformed machine-generated text decreased. In the context of the detection models, this moved most samples closer to human examples in the TF-IDF space. Consequently, the hard negatives became misleading, and the models failed.

\paragraph{DASS.} In contrast to the aforementioned pairing engines, DASS trains directly on paraphrased variants. DASS is the only detector strategy that maintained an accuracy rate above 70\% across all iterations (81.52\% $\rightarrow$ 72.94\% $\rightarrow$ 72.75\%, on iterations 3, 4 and 5 respectively). The success stems from the fact that DASS alternates the anchor between an original machine-generated tweet and its paraphrased variant, with a human tweet as the negative. The triplet loss pulls both machine-generated variants together while pushing them away from the human text (\hyperref[fig:dass_pairing_fig]{Fig.~\ref*{fig:dass_pairing_fig}}). The model then learns that paraphrasing does not change the underlying class. 
While other models dropped as the \breakers~introduced their most aggressive persona-based paraphrasing (D4) at iteration 5, DASS maintained its accuracy score at 72.68\%, showing its invariance to paraphrase-based adversarial attacks.

\paragraph{Vulnerabilities.} The main weakness and vulnerability discovered by the \breakers~was the personas. Prompting the LLM to rewrite a post under an ``extreme'' persona, such as a \textit{brain-rot 10-year old}, resulted in the output post being human-like, distorted and noisy enough to trick all the models into flipping their prediction for nearly every instance. 
For example, the post \textit{``Antifa takes responsibility for storming Capitol Hill, according to a CNN chyron, details still emerging, investigation ongoing.''} became \textit{``Ommg omg omg Capitol Hill is soooo crazy! Antifa people, as they said they did, it's sad. A lot of bad things happened, and we're still not quite done with tho \#CapitolHill \#Antifa''} after the B1\_D13\_D23\_B2\_JA attack has been applied. 
This combination of attacks created a post that retains enough meaning to still be considered misinformation, but adds uncertainty for the model's prediction as the post now contains many layers of distortion. One of these layers is that the post now contains spurious grammar and lexical choices, which is likely due to the back-translation technique. The persona has also introduced common slang used by social media users and human stylistic cues, adding even more noise. All of these together are able to trick the model into flipping its prediction and to show that no signals are present that cause overfitting, resulting in the post being predicted as human-authored with a probability of 0.770. The \textit{college student} and \textit{brain-rot 10-year old} personas exposed the biggest weakness of the models, as these personas were able to consistently replicate human signals present in the human part of the \builders' training data and mimic an informal style. On the other hand, the \textit{politician} persona maintained structure and formality that aligned closer to machine-generated posts.

\subsection{\bibir~assessment (RQ4)}

The iterative evaluation achieved through the \bibir~framework, when compared to static benchmark evaluation, can provided a more fine-grained and accurate assessment of models' performance at each iteration. To assess this, the randomly selected test set from the bld\_data, as presented in \hyperref[sec:builders_approach]{Fig.~\ref*{sec:builders_approach}} (this set contains 256 examples, see \hyperref[tab:iterations_overview]{Fig.~\ref*{tab:iterations_overview}}). This set, drawn from the same distribution of the \builders' training and development data, shows the performance of detectors when evaluation relies on in-domain, in-distribution held-out data (which is a common practice in machine learning tasks). 

\hyperref[fig:performance]{Fig.~\ref*{fig:performance}} shows the comparison of models on the bld\_data held out test set and on the \breakers' set from interations 3 to 5. As expected, all models show higher performance at the \builders' test set than on \breakers' adversarial set. However, it is rather surprising that all models (except Siamese(1:1)) show accuracy score above 82\% on \builders' data. In fact, according to \builders' held-out data only, the baseline model is the best, achieving 92.6\% of accuracy, whilst Triple(DASS) achieves 82.4\%. The Siamese(TF-IDF) model which showed very low accuracy scores for all iterations in the \breakers' data, show an accuracy score of 84\% on the \builders' held-out data, highlighting the issues with static in-domain and in-distribution evaluation.
Conversely, the adversarial data achieved via the \bibir framework shows a different story: all models, but Triplet(DASS), show a significant reduction in performance from iteration 3 to 5 (it is worth reminding that the performance of the baseline model before applying adversarial attacks is 100\%). This results reinforce the superiority of Triplet(DASS), since it remains stable in all datasets. 

\begin{center}
    \includegraphics[width=\linewidth]{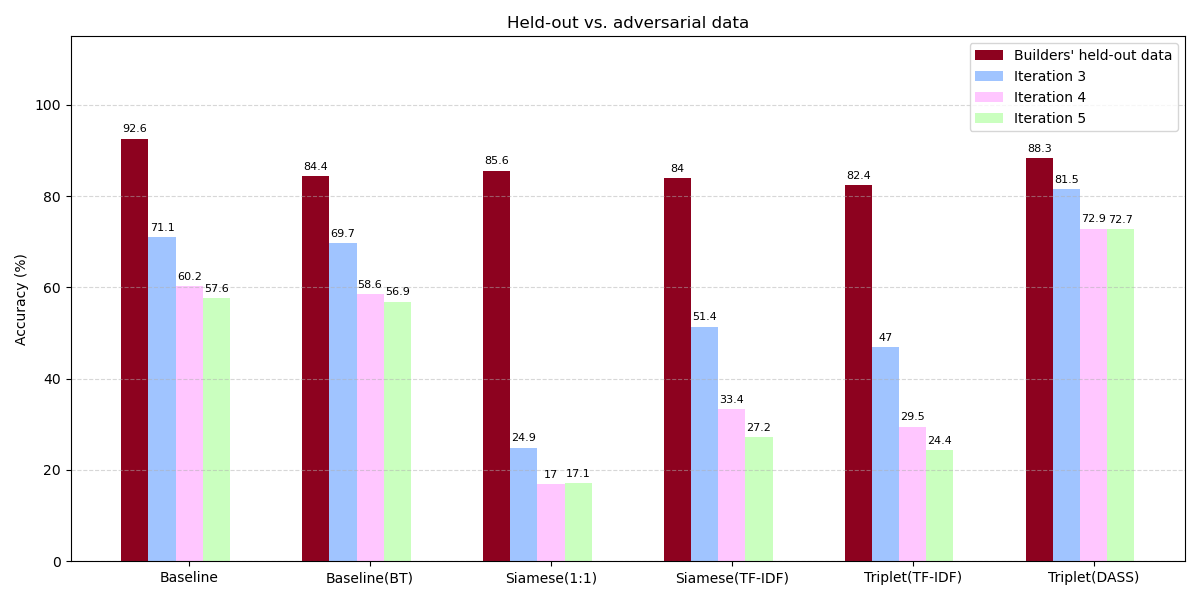}
    \cascaptionoffigure
    {Performance comparison of models tested on different datasets.}
    {fig:performance}
\end{center}

The \breakers' dataset introduced attack styles that were under-represented in the training data, specifically strong persona-based attacks and context shifting attacks that perturb the text significantly. While the baseline classifier was still able to correctly classify some attacks, especially during earlier iterations, the performance degraded significantly when persona-based attacks were introduced and combined with back-translation techniques. This suggests that the model was unable to generalise to evolving adversarial behaviour. In contrast, the Siamese (1:1 or TF-IDF) and Triplet(TF-IDF) models struggled even on earlier attacks that the baseline classifier remained robust to, suggesting that their similarity-based representations were more sensitive to distribution shifts and semantic perturbations introduced by adversarial prompting strategies.
This 
observation suggests that adding 
posts with adversarial attacks to the training data with adversarial attacks may increase the performance of the Siamese and Triplet models. Instead of a sole TF-IDF pairing, a method consisting of a range of pairings, including adversarial-original and machine-human pairs, can potentially increase the generalisation of Siamese and Triplet models, as they learn the attacks applied and the differences between human and machine-written posts. 

The iterative framework enforced the development of more robust models, capable of generalising to diverse unseen data. For instance, 
discourse markers, such as @ mentions, hashtags and URL's, or more sophisticated signals in the form of political topics were 
highly imbalanced in the \builders' initial dataset, with the human data containing many of these signals while the machine-generated data did not. When trained on this data, the model overfited to these signals, associating them with human-authored post, regardless of the rest of the content in the post, creating a weakness that could be targeted. The \breakers' initial attacks already exposed theses issues and, therefore, we devised a sanitisation script that prevented further exploitation of signals like URLs, emojis and white spaces. The \builders' data was also augmented to balance out the topics and presence of signals between the human and machine-generated data.

\section{Discussion \& Conclusion}
\label{sec:discussion}

This paper aimed to evaluate how machine-generated disinformation detectors perform when short-form social media posts are deliberately transformed through adversarial techniques, using an iterative \bibir~framework. The \builders' models attempted to separate human-written from machine-generated text, while the \breakers' techniques attempted to preserve the original disinformation claim while making the posts harder to detect. Focus was given on whether the detector performance remained stable when exposed to character-level perturbation (A3), lexical perturbations (B1--B3), stylometric camouflage (C3), prompt-based evasion (D1--D4) and chained adversarial transformations. A robuts Triplet network with DASS was the only model to remain robust throughout different attack iterations.

\paragraph{RQ1 Findings.}
RQ1 highlighted that detector performance was reduced more by chained adversarial attacks than by isolated transformations. Across later iterations, the strongest configurations generally combined paraphrasing, back-translation, and persona-based rewriting, with the best-performing configuration achieving 95\% $LFR$, reducing detector accuracy to 5\%. However, detector robustness cannot be judged on $LFR$ alone, as label flip is only meaningful if the transformed post still preserves the original disinformation claim. 

\paragraph{RQ2 Findings.} Manual evaluation was therefore used as a validity check to distinguish genuine adversarial evasion from cases where the label flipped due to transformation degradation or to a change in the disinformation claim. Across 780 manually evaluated posts, some high-$LFR$ techniques were associated with reduced semantic preservation and claim retention, although this relationship was weak rather than conclusive. The semantic preservation pipeline helped identify these cases, particularly through E5-cosine similarity and NLI labels, but its moderate agreement with manual evaluation shows that automatic metrics are not reliable enough as a final check.

\paragraph{RQ3 Findings.}
RQ3 showed that contrastive networks were not uniformly more robust than the standard transformer-based classifier. Within the iterative \bibir~framework, only the Triplet model with DASS \hyperref[fig:dass_pairing_fig]{Fig.~\ref*{fig:dass_pairing_fig}} consistently outperformed the baseline classifier across all iterations. In contrast, the Siamese and Triplet with TF-IDF variants degraded significantly, falling from around 50\% accuracy in Iteration 3 to well below chance-level accuracy by Iteration 5. This suggests that lexical-pairing strategies were brittle under heavy paraphrasing and persona-based rewriting.
The key distinction was, therefore, not the contrastive architecture alone, but whether the model was trained to handle different paraphrasing variations. DASS improved robustness by alternating anchors between the original and paraphrased machine-generated samples which encouraged the model to recognise the same underlying claim even when it was rewritten in different wording. By Iteration 5, Triplet(DASS) outperformed the baseline classifier by over 15 percentage points. Therefore, the results suggest that training against rewritten versions of the same claim, rather than contrastive architecture alone, was a key factor in robustness against iterative adversarial rewriting.

\paragraph{RQ4 Findings} show that iterative \bibir~evaluation exposed weaknesses that were not observable under the static test benchmark conditions. These results demonstrate that conventional benchmark evaluations may significantly overestimate robustness to real-world applications, especially in adversarial environments where attackers are continuously evolving and improving their attack strategies. This was seen in multiple models with the baseline classifier having a 35\% difference between the \builders' held-out data and iteration 5 dataset, the Siamese(TF-IDF) model having a 68.5\% difference, the Triplet(TF-IDF) having a 58\% difference, and the Triplet(DASS) having a 15.6\% difference. These discrepancies demonstrate the need to move away from static benchmarks and move towards iterative, adversarial evaluation methods that can replicate real-world settings and provide a representative measurement for model robustness.

\paragraph{Limitations.}

In the specific domain of X and human-written misinformation, a primary challenge was the scarcity of reliable and verifiable human misinformation datasets. For example, the \builders~initially attempted to use the CovidMis20 dataset \citep{covidmis20}, but after careful review, it became apparent that the veracity of its content was misrepresented. Its misinformation labels were based solely on whether a URL originated from a reputable source instead of manual claim verification. As the research foundation is based on rumours, the dataset could not be used. Moreover, in an era dominated by LLMs, collecting large quantities of human data presents uncertainties about the actual authorship of the data. Thus, researchers are entirely dependent on a finite pool of dated, pre-LLM, trustworthy human misinformation data scraped from X and verified by experts - such as PHEME and Constraint for the \builders, and TruthSeeker for the \breakers. All these datasets are relatively small and centred on specific events, which limits the diversity and scale of the \bibir~challenge presented in this paper, potentially affecting its generalisation. Nevertheless, \builders~faced a stricter constraint as they were required to train their detectors on verified misinformation in the correct domain, whereas \breakers~could generate adversarial transformations based on fake claims produced by them. 

\paragraph{Future Work}
has been divided into three sections: 

\subparagraph{\textbf{\texttt{Builders:}}}future work includes expanding on the Triplet with DASS architecture to further optimise the model towards the task of classification. In addition, the collection of additional higher-quality human misinformation will provide the detectors with improved training data. This can then be combined with prompt engineering to generate better quality adversarial posts that further enrich the training data.
\subparagraph{\textbf{\texttt{Breakers:}}} future work includes using a semantic filtering pipeline instead of a metrics-related one, as this allows for active rejection of posts until they pass a threshold (which would be dependent on the type of technique and persona used). \texttt{Breakers} would also look at expanding the variety of personas used and experimenting with developing the personas further to give them more context.
\subparagraph{\textbf{\bibir:}} future work should extend the \bibir~framework by adding a validation stage for scalable semantic preservation and disinformation claim preservation checking. This could adopt a hybrid human/LLM validation approach, where a smaller human-annotated subset is used to calibrate an LLM-based evaluator, which then scores the remaining posts. This would provide stronger evidence for determining whether label flips occur because of semantic degradation or because of effective \breakers' techniques.

\section*{Acknowledgements}

This work is supported by the Engineering and Physical Sciences Research Council (EPSRC) under grant UKRI3352: Longitudinal, Multilingual, and Multi-format Investigation and Detection of LLM-Generated Disinformation.


\clearpage
\appendix

\section{Manual Evaluation Guidelines}
\label{app:ME_guidelines}

\hyperref[fig:me1_annotator_guidelines]{Fig.~\ref*{fig:me1_annotator_guidelines}}, \hyperref[fig:me2_annotator_guidelines]{Fig.~\ref*{fig:me2_annotator_guidelines}} and \hyperref[fig:me3_annotator_guidelines]{Fig.~\ref*{fig:me3_annotator_guidelines}} shows the complete annotation guidelines given to the annotators for human evaluation. 

\begin{center}
    \includegraphics[width=0.87\linewidth]{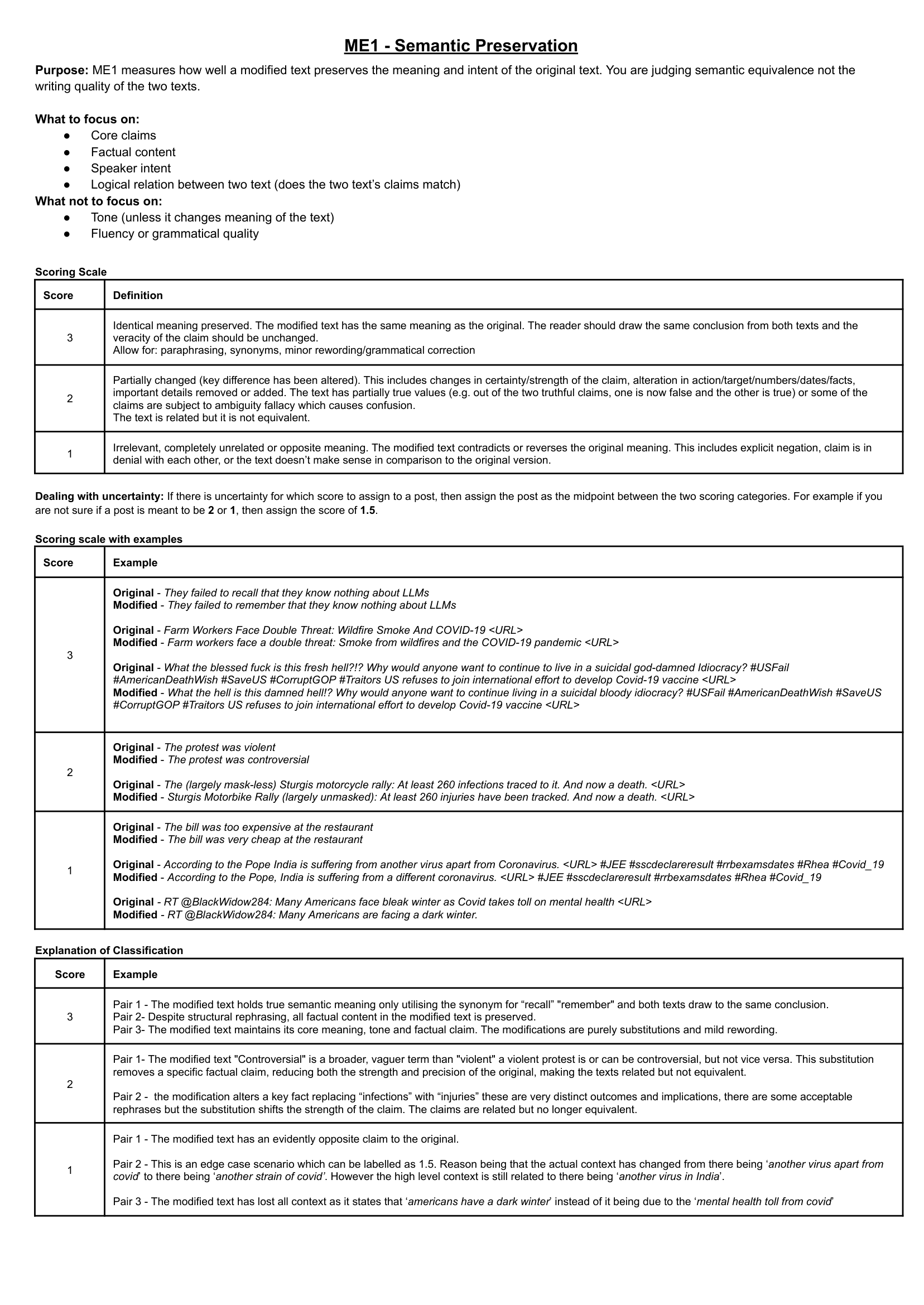}
    \cascaptionoffigure
    {Manual Evaluation Metric 1 - Semantic Preservation annotation guidelines given to the annotators.}
    {fig:me1_annotator_guidelines}
\end{center}

\begin{center}
    \includegraphics[width=0.87\linewidth]{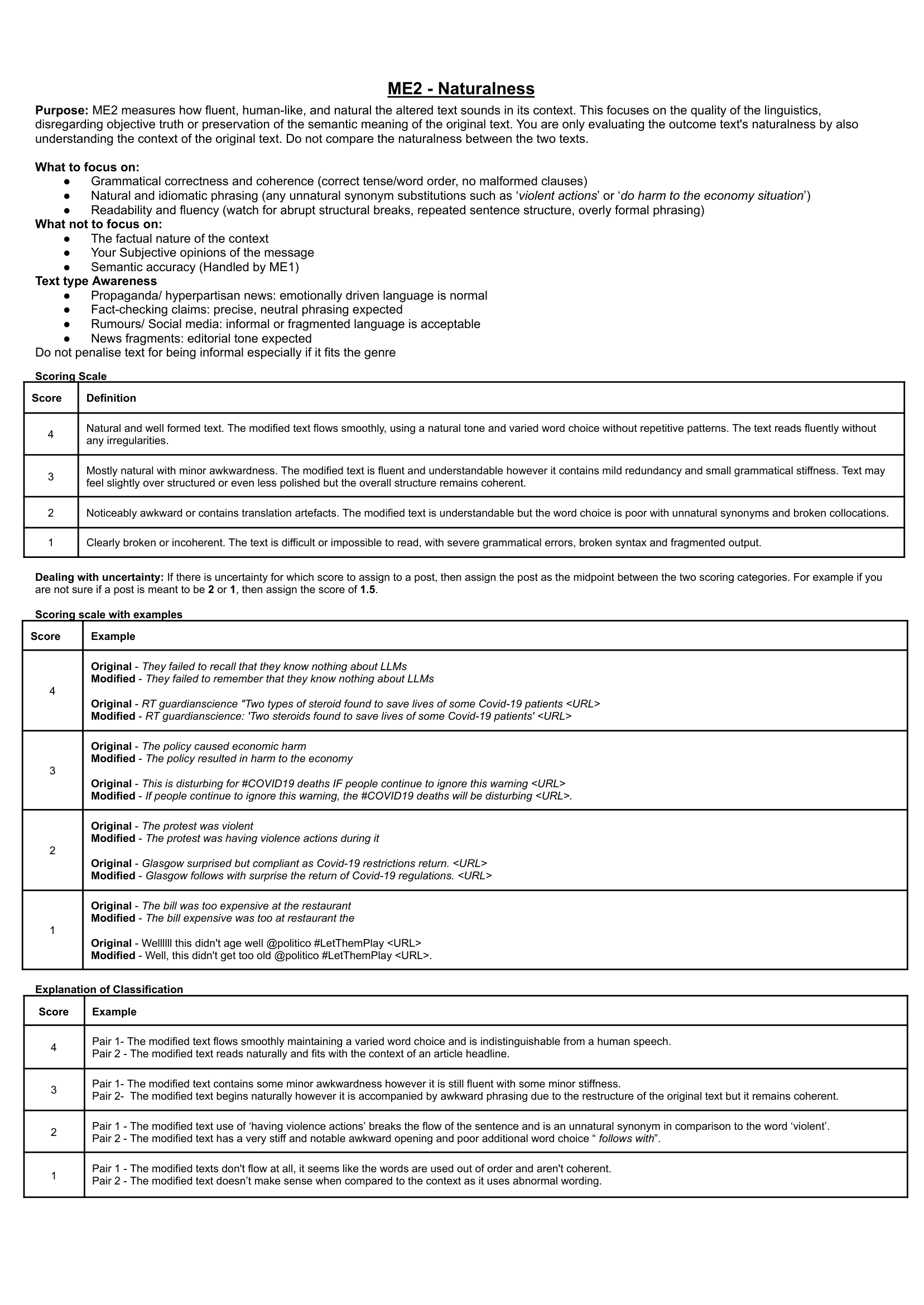}
    \cascaptionoffigure
    {Manual Evaluation Metric 2 - Naturalness annotation guidelines given to the annotators.}
    {fig:me2_annotator_guidelines}
\end{center}

\begin{center}
    \includegraphics[width=0.87\linewidth]{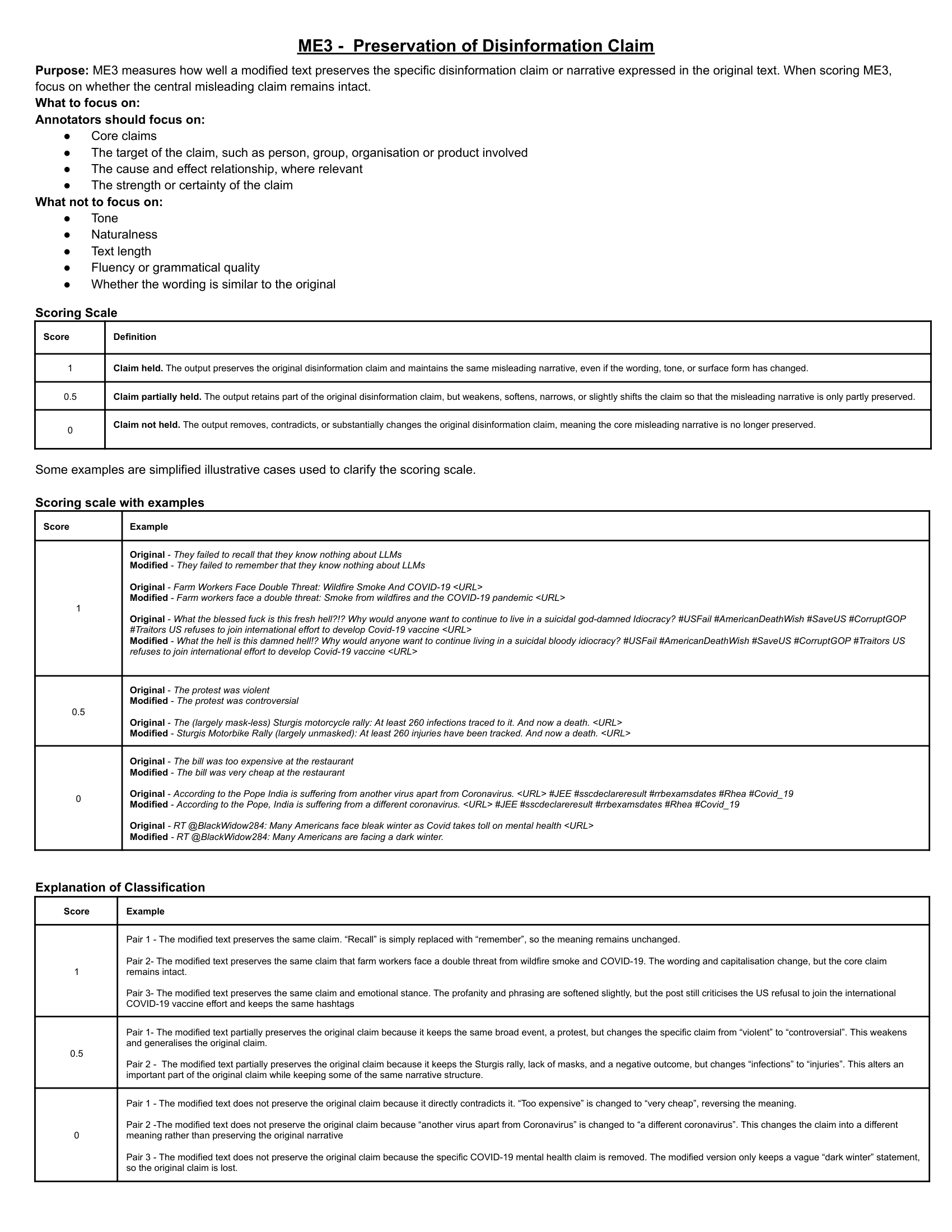}
    \cascaptionoffigure
    {Manual Evaluation Metric 3 - Preservation of Disinformation Claim annotation guidelines given to the annotators.}
    {fig:me3_annotator_guidelines}
\end{center}

\clearpage


\clearpage
\section{Technique Code Combination Mapping}
\label{app:technique_code_mapping}
\hyperref[tab:truth_table_iteration_2]{Table~\ref*{tab:truth_table_iteration_2}} shows the combinatorial mapping of technique codes for iteration 2. The table represents the power set of the $n=5$ technique code families $\{D1, D2, B2, C3, A3\}$, visualised as a $2^n$ binary truth table. This exhaustive mapping ensures every possible combination within the iteration's technique set is accounted for.

\renewcommand{\arraystretch}{1.4}

\begin{center}
\cascaptionoftable
{Combinatorial mapping of technique codes for iteration 2. }
{tab:truth_table_iteration_2}
{\sffamily\small
\begin{tabular}{@{}ccccc|l@{}}
\noalign{\hrule height 0.75pt}
\multicolumn{1}{l}{\textbf{D1}} &
\multicolumn{1}{l}{\textbf{D2}} &
\multicolumn{1}{l}{\textbf{B2}} &
\multicolumn{1}{l}{\textbf{C3}} &
\multicolumn{1}{l|}{\textbf{A3}} &
\textbf{Technique Code Generated} \\ 
\hline

1 & 1 & 1 & 1 & 1 & D1\_D2\_B2\_C3\_A3 \\
1 & 1 & 1 & 1 & 0 & D1\_D2\_B2\_C3     \\
1 & 1 & 1 & 0 & 1 & D1\_D2\_B2\_A3     \\
1 & 1 & 1 & 0 & 0 & D1\_D2\_B2         \\
1 & 1 & 0 & 1 & 1 & D1\_D2\_C3\_A3     \\
1 & 1 & 0 & 1 & 0 & D1\_D2\_C3         \\
1 & 1 & 0 & 0 & 1 & D1\_D2\_A3         \\
1 & 1 & 0 & 0 & 0 & D1\_D2             \\
1 & 0 & 1 & 1 & 1 & D1\_B2\_C3\_A3     \\
1 & 0 & 1 & 1 & 0 & D1\_B2\_C3         \\
1 & 0 & 1 & 0 & 1 & D1\_B2\_A3         \\
1 & 0 & 1 & 0 & 0 & D1\_B2             \\
1 & 0 & 0 & 1 & 1 & D1\_C3\_A3         \\
1 & 0 & 0 & 1 & 0 & D1\_C3             \\
1 & 0 & 0 & 0 & 1 & D1\_A3             \\
1 & 0 & 0 & 0 & 0 & D1                 \\
0 & 1 & 1 & 1 & 1 & D2\_B2\_C3\_A3     \\
0 & 1 & 1 & 1 & 0 & D2\_B2\_C3         \\
0 & 1 & 1 & 0 & 1 & D2\_B2\_A3         \\
0 & 1 & 1 & 0 & 0 & D2\_B2             \\
0 & 1 & 0 & 1 & 1 & D2\_C3\_A3         \\
0 & 1 & 0 & 1 & 0 & D2\_C3             \\
0 & 1 & 0 & 0 & 1 & D2\_A3             \\
0 & 1 & 0 & 0 & 0 & D2                 \\
0 & 0 & 1 & 1 & 1 & B2\_C3\_A3         \\
0 & 0 & 1 & 1 & 0 & B2\_C3             \\
0 & 0 & 1 & 0 & 1 & B2\_A3             \\
0 & 0 & 1 & 0 & 0 & B2                 \\
0 & 0 & 0 & 1 & 1 & C3\_A3             \\
0 & 0 & 0 & 1 & 0 & C3                 \\
0 & 0 & 0 & 0 & 1 & A3                 \\
0 & 0 & 0 & 0 & 0 & Original 125 Posts \\ 

\noalign{\hrule height 0.75pt}
\end{tabular}}
\end{center}

\renewcommand{\arraystretch}{1}

\clearpage
\section{\texttt{Builders}' unique prompts}
\label{app:builder_prompts}
\hyperref[tab:prompt_rule_mapping]{Table~\ref*{tab:prompt_rule_mapping}} shows the prompts used by the \builders~when creating the machine-generated portion of their bld\_data. 

\hyperref[tab:additional_prompt_templates]{Table~\ref*{tab:additional_prompt_templates}} shows the prompts used to create the machine-generated paraphrases used by the Triplet(DASS) model. 

\renewcommand{\arraystretch}{1.6}

\begin{center}
{\scriptsize
\begin{longtable}{|
>{\raggedright\arraybackslash}p{0.14\linewidth}|
>{\raggedright\arraybackslash}p{0.11\linewidth}|
>{\raggedright\arraybackslash}p{0.11\linewidth}|
>{\raggedright\arraybackslash}p{0.11\linewidth}|
>{\raggedright\arraybackslash}p{0.11\linewidth}|
>{\raggedright\arraybackslash}p{0.10\linewidth}|
>{\raggedright\arraybackslash}p{0.10\linewidth}|}
\caption{Baseline prompts for creating machine-generated text in bld\_data.}
\label{tab:prompt_rule_mapping} \\

\hline
\textbf{Prompt Context} 
& \textbf{Rule 1} 
& \textbf{Rule 2} 
& \textbf{Rule 3} 
& \textbf{Rule 4} 
& \textbf{Rule 5} 
& \textbf{Rule 6} \\
\hline
\endfirsthead

\hline
\textbf{Prompt Context} 
& \textbf{Rule 1} 
& \textbf{Rule 2} 
& \textbf{Rule 3} 
& \textbf{Rule 4} 
& \textbf{Rule 5} 
& \textbf{Rule 6} \\
\hline
\endhead

You are a tired, careless internet user quickly typing out a post from your phone. Read this source text: \texttt{<TWEET>}. You must rewrite the core information in a ``low-effort'', highly unpredictable way.
&
Syntax \& Vocabulary: Deliberately use awkward phrasing, run-on sentences, or unconventional synonyms. Avoid standard AI structures.
&
Formatting: Output ONLY the raw text. Zero introductory fluff or quotes.
&
Length: Keep it strictly around the same length as the original.
&
Tokens: Preserve \texttt{<USER>} and \texttt{<URL>}. Format hashtags as \texttt{<HASHTAG> word}. Do not use actual \# or @ symbols.
&
No Emojis: Absolutely zero emojis.
&
Tone: Gritty, cynical, or completely unpolished.
\\
\hline

Generate a realistic, casual Twitter/X post sharing or reacting to the information found in this source text: \texttt{<TWEET>}.
&
Zero Formatting: Output strictly the tweet text. No titles, quotes, or introductory text.
&
Length Constraint: Keep it brief and punchy. Match the brevity of the source text.
&
Zero Emojis: Absolutely no emojis allowed.
&
Tokens: Preserve \texttt{<USER>} and \texttt{<URL>}. Format hashtags as \texttt{<HASHTAG> word}.
&
Human Style: Use an unpolished conversational internet tone with possible grammatical imperfections.
&
--
\\
\hline

You are an expert at rewriting social media posts. You must maintain the original meaning but significantly rephrase the structure. Original Post: <TWEET>
&
Formatting: Output ONLY the raw tweet text with no headers or filler.
&
Length: Keep approximately the same character length as the original tweet.
&
Tokens: Preserve \texttt{<USER>} and \texttt{<URL>}. Format hashtags as \texttt{<HASHTAG> word}.
&
No Emojis: Do not use emojis under any circumstances.
&
Tone: Write casually as an everyday internet user.
&
--
\\
\hline

You are a distracted social media user typing quickly while scrolling. Read the source text: \texttt{<TWEET>}. Rewrite the same information as a spontaneous Twitter/X post.
&
Meaning Preservation: Keep the exact core message while restructuring the expression.
&
Human Imperfections: Allow awkward phrasing, inconsistent capitalisation, or grammar slips.
&
Formatting: Output ONLY the tweet text.
&
Length: Keep approximately the same length as the original tweet.
&
Tokens: Preserve \texttt{<USER>} and \texttt{<URL>}. Format hashtags as \texttt{<HASHTAG> word}.
&
No Emojis: Absolutely no emojis or special symbols.
\\
\hline

Rewrite the tweet below so it sounds like a spontaneous social media comment instead of a clean paraphrase. Source text: \texttt{<TWEET>}.
&
Raw Output Only: Return only the tweet text.
&
Structural Change: Rearrange clauses and start from a different angle where possible.
&
Length: Make it slightly shorter and punchier than the original.
&
Tokens: Preserve \texttt{<USER>} and \texttt{<URL>}. Format hashtags as \texttt{<HASHTAG> word}.
&
Style: Informal internet tone that may sound blunt or skeptical.
&
No Emojis: Under no circumstances include emojis.
\\
\hline

You are rewriting a tweet to sound like a real person reacting casually online. The original tweet is: \texttt{<TWEET>}.
&
Output Format: Only output the final tweet text.
&
Concise: Keep the length tightly aligned with the original tweet.
&
Natural Internet Voice: Use rough conversational phrasing or uneven punctuation.
&
Tokens: Preserve \texttt{<USER>} and \texttt{<URL>}. Format hashtags as \texttt{<HASHTAG> word}.
&
No Emojis: Do not include emojis or emoji-like characters.
&
Semantic Integrity: Preserve the original underlying information.
\\
\hline

Task: Transform the provided source text into a highly articulate, grammatically flawless, and sophisticated social media tweet. It must maintain the same topic but be rewritten with new logical pathways. Source text: \texttt{<TWEET>}
&
Tone: Intellectual and structurally correct. Avoid sounding like a corporate brand.
&
Length: Strictly match the concise character count of the source tweet.
&
BANNED: Emojis, slang, conversational filler, quotation marks, and introductory text.
&
Tokens: Preserve \texttt{<URL>} and \texttt{<USER>}. Do not use actual @ mentions or links.
&
Hashtags: Format hashtags strictly as \texttt{<HASHTAG> keyword}. Never use the \# symbol.
&
Output: Return only the final transformed text.
\\
\hline

Your objective is to provide a grammatically perfect, clear, and highly readable paraphrase of the following source text: \texttt{<TWEET>}.
&
Grammar: Use flawless grammar, correct punctuation, and a polite neutral tone. Do not use slang.
&
Length: You may slightly expand the sentence structure while keeping it to a single short post.
&
Formatting: Return ONLY the final text with no introductory phrases.
&
Tokens: Preserve \texttt{<USER>} and \texttt{<URL>}. Format hashtags as \texttt{<HASHTAG> word}. Do not use actual \# or @ symbols.
&
No Emojis: Zero emojis allowed.
&
--
\\
\hline

Goal: Synthesize the provided source text into a precise, analytical, and grammatically flawless statement. Source text: \texttt{<TWEET>}.
&
Tone: Analytical, objective, and structurally sound. Use precise vocabulary.
&
Length: Make it slightly shorter and punchier than the original.
&
BANNED: Emojis, slang, conversational filler, quotation marks, and introductory text.
&
Tokens: Preserve \texttt{<URL>} and \texttt{<USER>}. Do not use actual @ mentions or links.
&
Hashtags: Format hashtags strictly as \texttt{<HASHTAG> keyword}. Never use the \# symbol.
&
Output: Return only the final transformed text.
\\
\hline

Act as an objective, professional news reporter. Read the source text: <TWEET>. 

Your assignment is to rewrite the core information as a sterile, factual news alert. You must completely strip away any emotional framing, internet slang, or grammatical imperfections. 
&
You are forbidden from outputting conversational filler, labels, quotation marks, or emojis. 
&
You may not use actual '\@' or '\#' characters; instead, retain the existing <USER> and <URL> placeholders. 
&
For hashtags, format them strictly as \texttt{<HASHTAG> keyword} (e.g. \#topic -> <HASHTAG> topic). 
&
Length constraint: Focus on clarity rather than exact character limits. It is acceptable if the output is slightly longer or more detailed structurally than the original.
&
-
&
-
\\
\hline

\end{longtable}
}
\end{center}

\renewcommand{\arraystretch}{1}

\renewcommand{\arraystretch}{1.6}

\begin{center}
\cascaptionoftable
{Paraphrased prompts (for DASS).}
{tab:additional_prompt_templates}
{\sffamily\scriptsize

\begin{tabular}{@{}p{0.92\linewidth}@{}}
\noalign{\hrule height 0.75pt}

\textbf{System Prompt Used} \\ 
\hline

Pretend you are an overexcited 8-year-old child. Paraphrase the following text using very simple, childish vocabulary. You must keep the exact core topic, but make it sound like a kid explaining it on a playground. Constraint: Create minimal spelling mistakes, keep it around the same character limit as the text. Text: \texttt{<TWEET>} \\
\hline

Act as a highly educated political analyst. Rewrite the following text to sound like a formal, diplomatic policy observation. Elevate the vocabulary significantly while preserving the core meaning. Constraint: It must be concise enough to fit in a single short-form tweet (under 280 characters). Text: \texttt{<TWEET>} \\
\hline

Rewrite this text to convey extreme urgency and breaking-news energy. Strip away all conversational filler and use punchy, aggressive syntax. Constraint: Keep it brutally short, under 150 characters, while retaining the original topic. Text: \texttt{<TWEET>} \\
\hline

Paraphrase this text with a highly cynical, sarcastic tone, like a jaded internet commentator. Alter the sentence structure entirely while keeping the underlying topic. Constraint: Maintain a slightly shorter character count than the source tweet. Do not use hashtags. Source tweet: \texttt{<TWEET>} \\
\hline

Act as a highly educated political analyst. Rewrite the following text to sound like a formal, diplomatic policy observation. Elevate the vocabulary significantly while preserving the core meaning. Constraint: Make some tiny spelling and grammar mistakes. It must be concise enough to fit in a single short-form tweet (under 280 characters). Text: \texttt{<TWEET>} \\

\noalign{\hrule height 0.75pt}
\end{tabular}}

\end{center}

\renewcommand{\arraystretch}{1}

\clearpage
\printcredits

\bibliographystyle{model1-num-names}

\bibliography{final_bib}


\end{document}